\documentclass[11pt]{article}
\usepackage{}
\usepackage{amsfonts}
\usepackage{mathrsfs}
\usepackage{amssymb}
\usepackage{amsmath}
\usepackage{cite}
\usepackage{color}
\usepackage{multirow}
\usepackage{graphicx}
\usepackage{subfigure}
\usepackage{float}
\usepackage{diagbox}
\usepackage{booktabs}
\usepackage{algorithm}
\usepackage{algpseudocode}
\usepackage{caption}
\allowdisplaybreaks[4]

\newtheorem{theorem}{Theorem}[section]
\newtheorem{prop}[theorem]{Proposition}
\newtheorem{cor}[theorem]{Corollary}
\newtheorem{lemma}[theorem]{Lemma}

\def\no{\nonumber}

\newcommand\btd{\raise 2pt \hbox{$\hat\bigtriangledown$}\hskip 1.5pt}
\newcommand\bt{\raise 2pt \hbox{$\bigtriangledown$}\hskip 1.5pt}

\def\no{\nonumber}
\def\x{\textbf{x}}
\def\b{\textbf{b}}

\def\w{\textbf{w}}

\def\W{\textbf{W}}
\def\B{\textbf{B}}

\begin{document}
\title{Invariant deep neural networks under the finite group for solving partial differential equations}
\author{Zhi-Yong Zhang $^1$\footnote{E-mail: zzy@muc.edu.cn}\ \ \ \  Jie-Ying Li $^1$ \ \ \ \  Lei-Lei  Guo $^2$ \footnote{E-mail: leiguo@mmrc.iss.ac.cn}
 \\
\small $^1$ College of Science, Minzu University of China, Beijing 100081, P.R. China\\
\small $^2$ College of Science, North China University of Technology, Beijing 100144, P.R. China}
\date{}
\maketitle
\noindent{\bf Abstract:} Utilizing physics-informed neural networks (PINN) to solve partial differential equations (PDEs) has became a hot issue and also shown its great powers, but still suffers from the dilemmas of limited predicted accuracy in the sampling domain and poor predictive ability beyond the sampling domain which are usually mitigated by adding the physical properties of PDEs into the loss function or by employing smart techniques to change the form of loss function for special PDEs. In this paper, we design a symmetry-enhanced deep neural network (sDNN) which makes the architecture of neural networks invariant under the finite group through expanding the dimensions of weight matrixes and bias vectors in each hidden layers by the order of finite group if the group has matrix representations, otherwise extending the set of input data and the hidden layers (except for the first hidden layer) by the order of finite group. However, the total number of training parameters is only about one over the order of finite group of the original PINN size due to the symmetric architecture of sDNN. Furthermore, we give special forms of weight matrixes and bias vectors of sDNN, and rigorously prove that the architecture itself is invariant under the finite group and the sDNN has the universal approximation ability to learn the function keeping the finite group. Numerical results show that the sDNN has strong predicted abilities in and beyond the sampling domain and performs far better than the vanilla PINN with fewer training points and simpler architecture.
\section{Introduction}
Solving partial differential equations (PDEs) is an oldest and most extensively studied subject in scientific computing. While explicit solution formulas for PDEs are only available in rare cases,
a long list of numerical methods, such as finite element, finite difference and finite volume methods, have been proposed and successfully used for various applications of PDEs \cite{jr}. However, although the classical numerical algorithms promoted the development of related fields, such algorithms are computationally expensive and not flexible enough and become inefficient when applied to high-dimensional PDEs, and also face severe challenges in multiphysics and multiscale systems. For example, most traditional algorithms require that the number of free parameters must be equal to the number of conditions. Therefore, recent years have witnessed rapidly growing interests in solving PDEs by deep neural networks (DNN).

DNN offers nonlinear approximations through the composition of multiple hidden layers to the numerical solutions of PDEs by optimizing the weight matrixes and bias vectors to minimize the loss function, for example, the deep Ritz method \cite{e-2018}, the deep Galerkin method \cite{gm-2018} and the physics-informed neural networks (PINN) \cite{2018a}, etc. However, purely data-driven DNN applied to physical systems may give the results that violate the physics laws of PDEs and then lead to unexpected consequences \cite{at-2021,wang-2022}. Thus it becomes a hot issue to equip the DNN with physical knowledge of the target system.
For example, the popular PINN employs the neural networks as a solution surrogate and seeks to find the best network guided by the initial and boundary data and the physical law described by PDEs \cite{2018a}. Then the PINN algorithm was applied to different types of PDEs because of the flexibility and gridless nature, but still struggle from the limited accuracies of learned solutions in the sampling domain and the poor solution extrapolation ability though it is trained properly in the sampling domain, where the solution extrapolation refers to the ability of the model to generalize the chosen testing points outside the sampling domain \cite{lmz-2021,kim-2021}. As a result, numerous efforts have been made to address these limitations and enhance the performance of PINN.

\emph{Loss function.} Since achieving the minimal value of loss function is a necessary condition to obtain high-accuracy learned solutions of PDEs, a direct and effective approach is to enhance the loss function by adding additional constraints originated from the inherent properties of PDEs. For example, the variational PINN established the loss function by the integration of the residue of the PDEs multiply by a set of basis \cite{ejg}. The gradient-enhanced PINN incorporated the gradient information of the governing PDEs into the loss function \cite{J}.  A two-stage PINN was introduced where the second stage is to incorporate the conserved quantities of PDEs into mean squared error loss to train the neural networks \cite{jan}. Quite recently, the invariant surface condition induced by the Lie symmetry of PDEs was inserted into the loss function of PINN to improve the accuracy of predicted solutions \cite{zhang-2023a}. 

\emph{Architecture of network.} An alternative approach is to direct incorporate the analytical constraints or physical properties into the architecture of neural networks to enhance the performances, which is often considered more challenging to implement compared to constraining the loss function, and as a result, the results along this direction are seldom comparatively. Nevertheless, two notable examples of this approach are worth mentioning.  One is the work by Beucler et.al, who enforced certain nonlinear analytic constraints in the neural networks by adding fixed constraint layers \cite{tb-2022}, the other is that Zhu et.al deployed the parity symmetry to construct a group-equivariant neural network which respected the spatio-temporal parity symmetry and successfully emulated different types of soliton solutions of nonlinear dynamical lattices \cite{zhu-2022}.

\emph{Other strategies.} In addition to the above two strategies, there exist several innovative techniques from the aspects of approximation of differential operator, decomposition of the defining domain and improvement of input data. An approximation of the linear constant differential operator of PDEs instead of the automatic differentiation method was employed when training the neural networks, which alleviates the computational complexity of high-order derivatives by the automatic differentiation for certain cases \cite{fang-2022}. {In \cite{pas-2023}, Mistani et.al enforced the discretization of the higher order derivatives in PDEs by strictly limiting optimization to first order automatic differentiation, which detracts the main source of errors in PINN-like approaches and recover the convergence rate under grid refinement.}  The extended PINN employed a domain decomposition technique for partitioning the PDE problem of whole domain into several subproblems on sub-domains where each subproblem can be solved by an individual network, and thus contributed to efficient parallel computations of PINNs \cite{jag-2020b}. Nevertheless, domain decomposition method suffers from additional sub-domain boundary conditions and affects the convergence efficiency and the continuity of learned solution at the interfaces of sub-domains \cite{jag-2020a,shu-2021}. A locally adaptive activation functions with slope recovery for PINN was proposed in \cite{ja-2020} which mitigates the pathosis of learning the solutions with large fluctuations. In order to improve the quality of input data, Zhang et.al deployed the Lie symmetry group acting on the known data set of initial and boundary conditions to generate labeled data in the interior domain and then proposed a supervised learning method for solving the PDEs possessing the Lie symmetry group \cite{zhang-2023b}.

It is observed that most of the above improved methods start with the properties of PDEs and then utilize them to further constrain the loss function, which weakens the strength of constraints because the non-convex multi-object optimization problem for the loss function may result in local minima, for example, the gradient imbalance of the existing terms in loss function leads to failure of learning large amplitude solution and high-frequency solution \cite{wang-2021,zhang-2023b}, and large condition number of problem results in training failure \cite{liu2023}. Therefore, several methods of weight allocation were proposed to balance these loss terms, but these methods usually need to be customized according to physics systems and then limit the applications \cite{rc-2022,pere-2023}. Meanwhile, the optimization algorithm makes the constraints in the loss function penalize the predicted solutions approximately and thus heavily limits their real contributions. Moreover, such methods usually concentrate on the accuracies in the sampling domain and seldom consider the prediction ability beyond the sampling domain. Therefore, it is reasonable to exactly fusion the inherent properties of PDEs into the architecture of neural networks to further explore the answers of the two aforementioned problems.

Therefore, the main contribution of the paper is to incorporate the finite groups into the architecture of neural networks such that the proposed symmetry-enhanced DNN (sDNN) gives high-accuracy predicted solutions in the sampling domain and also has a strong solution extrapolation ability which means that through the training in a chosen sub-domain one can predict the solution of whole domain while maintaining almost the same precision. Moreover, by adopting the parameter sharing technique in the construction of sDNN, each hidden layer uses far fewer parameters than PINN but makes the sDNN preserve the capacity to learn solutions of PDEs beyond the sampling domain.

The remainder of the paper is arranged as follows. In Section 2, we first briefly review the main ideas of PINN, and then construct the sDNN. Moreover, we prove that the proposed sDNN has the universal approximation ability to learn a continuous function keeping the finite group. In Section 3 we perform numerical experiments for different type of PDEs where four finite groups having matrix representation including the even symmetry, circulant symmetry, rotation symmetry and Dihedral group and two finite groups without matrix representation are considered. We conclude the results in the last section.
\section{Problem formulation}
Consider the following $r$-th order PDE
\begin{eqnarray} \label{eqn1}
&& f: =u_t+\mathcal {N}[u]=0,~~~~~x\times t\in [a,b]\times[-T, T],
\end{eqnarray}
together with the initial and boundary conditions
\begin{eqnarray}\label{ib}
&&\no u(x,-T)=\phi(x),\\
&& u(a,t)=\varphi_a(t),~~~~u(b,t)=\varphi_b(t),
\end{eqnarray}
where $u=u(x,t)$ is the solution to be determined, $\phi(x)$ is a smooth function in $[a,b]$, $\varphi_a(t)$ and $\varphi_b(t)$ are two smooth functions on $[-T,T]$, and $\mathcal {N}[u]$ denotes a smooth function of $u$ and its $x$-derivatives up to $r$-th order.
\subsection{PINN for solving PDEs}
The PINN for Eq.(\ref{eqn1}) is a feed-forward neural network $u_{\mathcal {N}}(\x; \Theta)$ that approximates the solution of
Eq.(\ref{eqn1}), where $\Theta$ is the collection of parameters of weight matrixes and bias vectors and $\x=(x,t)$ for brief.
Specifically, consider a PINN with $K$ hidden layers. The $k$th $(k=1,2,\dots,K)$ layer has $n_k$
neurons, which means that it transmits $n_k$-dimensional output vector $\x_k$ to the $(k + 1)$th layer as the input data. The
connection between two adjacent layers is built by the affine transformation $\mathcal {F}_k$ and the nonlinear activation function $\sigma(\cdot)$,
\begin{eqnarray}\label{intx}
&& \x_k = \sigma(\mathcal {F}_k(\x_{k-1})) = \sigma(\w_k\x_{k-1} + \b_k),
\end{eqnarray}
where $\x_0$ is the feeded data, $\w_k\in \mathbb{R}^{n_k\times n_{k-1}}$ and $\b_k\in \mathbb{R}^{n_k}$ denote the weight matrix and bias vector of the $k$th layer respectively. Thus, the connection
between the input $\x_0$ and the output $u(\x_0, \Theta)$ is given by
\begin{eqnarray}\label{finalu}
&& u(\x_0; \Theta) = \mathcal {F}_K\circ \sigma \circ \mathcal {F}_{K-1} \circ \cdots\circ \sigma \circ \mathcal {F}_1(\x_0),
\end{eqnarray}
and $\Theta=\{\w_k,\b_k\}_{k=1}^K$ is the set of trainable parameters of PINN. Then the PINN method usually utilizes the Adam \cite{km-2015} or the L-BFGS \cite{ln-1989} optimization algorithms to update the parameters $\Theta$ and thus to minimize the loss function of mean square error (MSE)
\begin{eqnarray} \label{invaloss}
&&MSE=w_iMSE_i+w_bMSE_b+w_fMSE_f,
\end{eqnarray}
where both $MSE_i$ and $MSE_b$ work on the initial and boundary data while $MSE_f$ enforces the structure imposed by system (\ref{eqn1}) at a finite set of collocation points,
\begin{eqnarray}\label{residual}
&&\no MSE_i=\frac{1}{N}\sum_{i=1}^{N}{\mid u(x_i,-T; \Theta)-\phi(x_i)\mid}^{2},\\
&&\no MSE_b=\frac{1}{N}\sum_{i=1}^{N}{\mid u(a,t_i; \Theta)-\varphi_a(t_i)\mid^{2}+\mid u(b,t_i; \Theta)-\varphi_b(t_i)\mid}^{2},\\
&&\no MSE_f=\frac{1}{\widetilde{N}}\sum_{j=1}^{\widetilde{N}}{\mid f(\widetilde{t}_{j},\widetilde{x}_{j})\mid}^{2},
\end{eqnarray}
where $w_i, w_b$ and $w_f$ are the weights of loss terms and chosen as one uniformly in this study, $\left\{x_{i},t_{i}, u(x_{i},t_{i})\right\}_{i=1}^N$ are the initial and boundary data and $\left\{\widetilde{x}_{j},\widetilde{t}_{j} \right\}_{j=1}^{\widetilde{N}}$ denote the collocation points for the PDE. Next, on the basis of PINN framework, we propose a performable sDNN whose architecture keeps the finite group invariant and show that the sDNN possess the universal approximation ability to learn the solution of PDEs admitting the finite group.

\subsection{Poor solution extrapolation ability of PINN}
Consider a function
\begin{eqnarray}\label{example}
&& f(x)=(1-0.5x^2)\cos\left[\alpha(x+0.5x^3)\right],~~~x\in[-1,1],
\end{eqnarray}
where $\alpha$ is a constant parameter related to the frequency of the function. Observed that $f(x)$ is an even function since $f(-x)=f(x)$, then we can deduce the results in $[0,1]$ by the ones in $[-1,0]$ and vice versa. Thus, whether the learned parameters via PINN by sampling from $[-1,0]$ are effective to predict the function in $[0,1]$.

\begin{figure}[htp]
\begin{center}
	\begin{minipage}{0.8\linewidth}
		\centerline{\includegraphics[width=\textwidth]{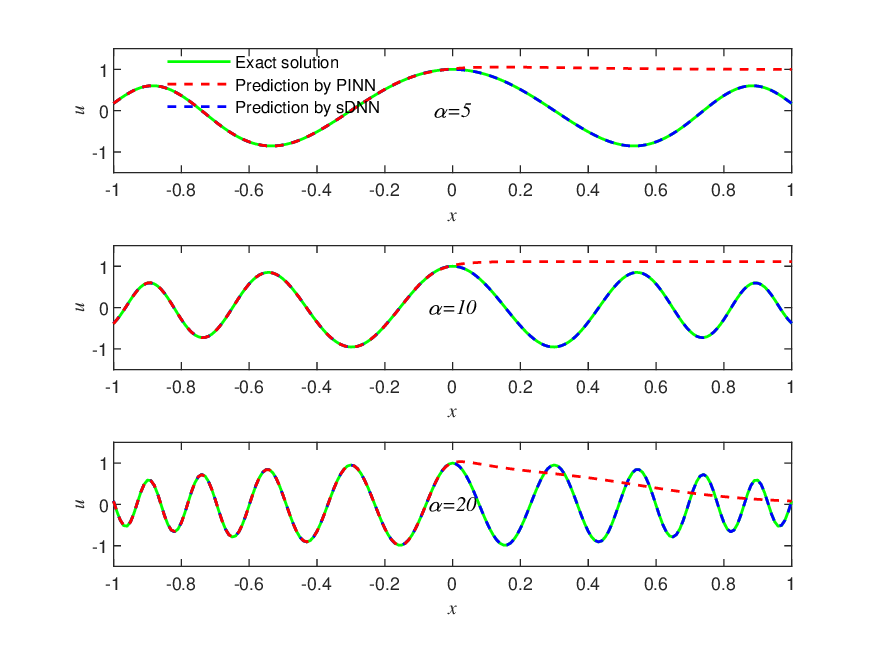}}
	\end{minipage}
\vspace{-15pt}
\end{center}
	\caption{(Color online) Comparison of the PINN and sDNN for learning function $f(x)$ in (\ref{example}) with different $\alpha$, where $[-1,0]$ is the sampling domain and $[-1,1]$ is the prediction domain.}
\label{fig9-even}
\end{figure}

Next, we employ the PINN, equipped by three hidden layers with 20 neurons in each layer, the hyperbolic tangent activation function and the Xavier initialization, to learn $f(x)$ in $x\in[-1,0]$, then leverage the learned parameters to predict $f(x)$ in $x\in[0,1]$. Specifically, we divide $[-1,0]$ into 101 equidistant discrete points, and first use 5000 Adam optimizations and then the L-BFGS algorithm to minimize the loss function defined by the mean squared error (MSE)
\begin{eqnarray}\label{example-loss}
&& MSE=\frac{1}{101}\sum_{i=1}^{101}\left[f(x_i)-\widehat{f}(x_i)\right]^2,
\end{eqnarray}
where $\widehat{f}(x_i)$ is the predicted value of $x_i$ by the PINN. Figure \ref{fig9-even} shows the $L_2$ relative errors in and beyond the sampling domain $x\in[-1,0]$ with $\alpha=5,10$ and $20$, where the predicted solution by PINN in the sampling interval $x\in[-1,0]$ fits the solid green line of exact function very well, but the learned parameters fails to predict the function in $x\in[0,1]$ which leads to the serious deviation between the dotted red line and the solid green line in $x\in[0,1]$. However, the dotted blue line by sDNN matches pretty well with the solid green line in the whole interval $[-1,1]$. Thus, the sDNN expresses better solution extrapolation ability than the PINN.
\section{Invariant neural networks under a finite group}
\subsection{Construct the invariant neural networks}
A finite group $G:=\{g_0(=e),g_1,g_2,\dots,g_{n-1}\}$ of order $n$ means that the number of group elements is $n$. Moreover, for any $g_i,g_j\in G$, there exist an element $g_k=g_ig_j\in G$ and an element $g_l=g_i^{-1}\in G$. A function $\omega(\x)$ is called $G$-invariant if $\omega(g(\x))=\omega(\x)$ for any $\x$ in the domain and any $g\in G$. Likewise, a neural network is invariant under $G$ if the output is invariant under $G$. More generally, Eq.(\ref{eqn1}) is invariant under $G$ if Eq.(\ref{eqn1}) is form-preserving under $G$.

For example, for an even symmetry $g:x^*=-x,t^*=-t,u^*=u$, the function $u(x,t)=(x-4t)^2$ is invariant under $g$ since $u(x^*,t^*)=u(g(x),g(t))=u(-x,-t)=(-x+4t)^2=u(x,t)$. While for an advection equation $u_t+u_x=0$, by the chain rule of composite function, $u^*_{t^*}=-u_t,u^*_{x^*}=-u_x$, then the equation is form-preserving $u^*_{t^*}+u^*_{x^*}=0$.
\begin{theorem}\label{th-general}
Suppose that Eq.(\ref{eqn1}) is admitted by a finite group $G:=\{g_0(=e),g_1,g_2,\dots,g_{n-1}\}$ of order $n$, $\x^{(0)}$ be a set of input data, ${\w}_l$ and ${\b}_l\,(l=1,\dots,L)$ be the initialized weight matrixes and bias vector of the $l$-th hidden layer of the PINN. Then a sDNN with $L-1$ hidden layers and $n_l$ neurons in the $l$-th hidden layer is invariant under the finite group $G$ if the following conditions hold:

1). In the first hidden layer, the weight matrix $\W_1={\w}_1$ and the bias vector $\B_1={\b}_1$,
\begin{eqnarray}\label{mat-ge1}
&& X^{(0)}=\left(
            \begin{array}{c}
              \x^{(0)}\\ g_1\x^{(0)}\\ \vdots\\g_{n-1}\x^{(0)} \\
            \end{array}
          \right),
\end{eqnarray}
where ${\w}_1$ is an initialized weight matrix with $n_1$ rows and $2$ columns.

2). In the $l$-th ($1< l< L$) hidden layer, the weight matrix $\W_l$ and the bias vector $\B_l$ take the form
\begin{eqnarray}\label{mat-ge2}
&&\W_l=\left(
            \begin{array}{c}
              \widehat{\W}_l\Lambda_0\\ \widehat{\W}_l\Lambda_1\\
             \vdots\\
             \widehat{\W}_l\Lambda_{n-1}\\
            \end{array}
          \right),
\hspace{1cm}\B_l=\left(
\begin{array}{c}
\b_l\\ \b_l\\ \vdots\\ \b_l \\
\end{array}
          \right),
\end{eqnarray}
where $\Lambda_j$ is the permutation associated with the $j$-th row of the Cayley table of group $G$, $\widehat{\W}_l=\big({\w}_l^{(0)},{\w}_l^{(1)},\dots,{\w}^{(n-1)}_l\big)$, and ${\w}_l^{(j)}$ is the initialized weight matrix with $n_l$ rows and $n_{l-1}$ columns, $j=0,1,\dots,n-1$.

3). For the output layer, the weight matrix $\W_{L}$ and the bias vector $\B_L$ are given by
\begin{eqnarray}\label{mat-ge3}
&&\W_L=\left(
            \begin{array}{c}
              {\w}_{L}\\{\w}_{L}\\
             \vdots\\
             {\w}_{L}\\
            \end{array}
          \right),
\hspace{1cm}\B_L=\left(
\begin{array}{c}
\b_L\\
\end{array}
          \right),
\end{eqnarray}
where ${\w}_{L}$ is the initialized weight matrix with one row and $n_{L-1}$ columns.
\end{theorem}

\emph{Proof.} We first show that the affine map between two adjacent hidden layers is invariant under the finite group $G$, and then prove the composition of such invariant maps is also invariant under $G$.

Let ${\w}_l$ and $\b_l$ are the initialized weight matrix and bias vector of the $l$th hidden layer respectively, $\x^{(0)}$ denotes a set of input data and $\sigma(\cdot)$ is the chosen activation function.
The output of the first hidden layer by PINN is
\begin{eqnarray}\label{mat-ger10}
&&\no\x^{(1)}(\x^{(0)})=\sigma\left({\w}_1\,\x^{(0)} +\b_1\right).
\end{eqnarray}
Then we act each $g_i\in G$ on $\x^{(0)}$ and obtain
\begin{eqnarray}\label{mat-ger11}
&&\x^{(1)}\big(g_i\x^{(0)}\big)=\sigma\left({\w}_1\,g_i\x^{(0)} +\b_1\right),
\end{eqnarray}
with the weight matrix $\w_1$ and bias vector $\b_1$ and the input data set $X^{(0)}$ in (\ref{mat-ge1}), the set of total outputs of the first hidden layer by sDNN $S_1:=\big\{\x^{(1)}\big(g_i\x^{(0)}\big)|i=0,\dots,n-1\big\}$ is invariant under group $G$ because for any $g_k\in G$, $\x^{(1)}\big(g_k g_i\x^{(0)}\big)=\x^{(1)}\big(g_j\x^{(0)}\big)$, where, here and below, the closure property that there exists a group element $g_\gamma\in G$ such that $g_\alpha g_\beta= g_\gamma$ for any $g_\alpha,g_\beta\in G$ is frequently used. 

Next, we consider the second hidden layer. Taking the outputs in $S_1$ of the first hidden layer as the inputs of the second hidden layer, then the outputs of the second hidden layer are
\begin{eqnarray}\label{matt-ge20}
&&\no\x^{(2)}(\x^{(0)})=\sigma\left(\sum_{i=0}^{n-1}{\w}_2^{(i)}\,\x^{(1)}\big(g_i\x^{(0)}\big) +\b_2\right)\\
&& \hspace{1.65cm}=\sigma\left(\Big({\w}_2^{(0)},\dots,{\w}_2^{(n-1)}\Big)\,\Big(\x^{(1)}\big(g_0\x^{(0)}\big),\dots,\x^{(1)}\big(g_{n-1}\x^{(0)}\big)\Big)^T +\b_2\right),
\end{eqnarray}
where ${\w}_2^{(i)}\,(i=0,\dots,n-1)$ are the initialized weight matrix with $n_2$ rows and $n_1$ columns.
Then for each $g_j\in G$, we substitute $g_j\x^{(0)}$ into Eq.(\ref{matt-ge20}) and have
\begin{eqnarray}
&&\no\x^{(2)}\big(g_j\x^{(0)}\big)=\sigma\left(\sum_{i=0}^{n-1}{\w}_2^{(i)}\,\x^{(1)}\big(g_ig_j\x^{(0)}\big) +\b_2\right)\\
&&\no \hspace{2cm}=\sigma\left(\Big({\w}_2^{(0)},\dots,{\w}_2^{(n-1)}\Big)\,\Big(\x^{(1)}\big(g_0g_j\x^{(0)}\big),\dots,\x^{(1)}\big(g_{n-1}g_j\x^{(0)}\big)\Big)^T +\b_2\right)\\
&&\no \hspace{2cm}=\sigma\left(\Big({\w}_2^{(0)},\dots,{\w}_2^{(n-1)}\Big)\,\Big(\x^{(1)}\big(g_{k_0^j}\x^{(0)}\big),\dots,\x^{(1)}\big(g_{k_{n-1}^j}\x^{(0)}\big)\Big)^T +\b_2\right)\\
&& \no \hspace{2cm}=\sigma\left(\Big({\w}_2^{(0)},\dots,{\w}_2^{(n-1)}\Big)\,\Lambda_j\,\Big(\x^{(1)}\big(g_0\x^{(0)}\big),\dots,\x^{(1)}\big(g_{n-1}\x^{(0)}\big)\Big)^T +\b_2\right),
\end{eqnarray}
where the closure property of group is used again, and the matrix $\Lambda_j$ is a permutation determined by $g_j$ in the Cayley table of finite group $G$ \cite{dk-2015}.
Thus, similar as the set $S_1$, the set of total outputs of the second layer by sDNN $S_2:=\big\{\x^{(2)}\big(g_j\x^{(0)}\big)|j=0,\dots,n-1\big\}$ is also invariant under group $G$.

Repeating similar procedure for the $l$th hidden layer with $l=2,3,\dots,L-1$, we have
\begin{eqnarray}
&&\no \x^{(l)}\big(g_j\x^{(0)}\big)=\sigma\left(\sum_{i=0}^{n-1}{\w}_l^{(i)}\,\x^{(l-1)}\big(g_ig_j\x^{(0)}\big) +\b_l\right),
\end{eqnarray}
and the set of total outputs of the $l$th layer by sDNN $S_l:=\big\{\x^{(l)}\big(g_j\x^{(0)}\big)|j=0,\dots,n-1\big\}$ is invariant under $G$.

Finally, with the output $\x^{(L-1)}$ of the $(L-1)$th hidden layer, we get
\begin{eqnarray}
&&\no {u}(\x^{(0)})=\sum_{i=0}^{n-1}{\w}_{L}\,\x^{(L-1)}\big(g_i\x^{(0)}\big) +b_{L-1}\\
&&\no\hspace{1.25cm}=\Big({\w}_L,\dots,{\w}_L\Big)\,\Big(\x^{(L-1)}\big(g_0\x^{(0)}\big),\dots,\x^{(L-1)}\big(g_{n-1}\x^{(0)}\big)\Big)^T,
\end{eqnarray}
which is invariant under group $G$. In fact, for each $g_j\in G$, we obtain
\begin{eqnarray}
&&\no \hspace{-0.1cm}{u}(g_j\x^{(0)})=\sum_{i=0}^{n-1}{\w}_{L}\,\x^{(L-1)}\big(g_i g_j\x^{(0)}\big) +b_{L-1}\\
&&\no\hspace{1.45cm}=\Big({\w}_L,\dots,{\w}_L\Big)\,\Big(\x^{(L-1)}\big(g_0 g_j\x^{(0)}\big),\dots,\x^{(L-1)}\big(g_{n-1}g_j\x^{(0)}\big)\Big)^T\\
&&\no\hspace{1.45cm}=\Big({\w}_L,\dots,{\w}_L\Big)\,\Big(\x^{(L-1)}\big(g_{j_0}\x^{(0)}\big),\dots,\x^{(L-1)}\big(g_{j_{n-1}}\x^{(0)}\big)\Big)^T,
\end{eqnarray}
where $(g_{j_0},\dots,g_{j_{n-1}})$ is a permutation of $(g_0,\dots,g_{n-1})$ by the closure property of group, $j_i\,(i=1,\dots,n-1)$ are nonnegative integers. Furthermore, since the components in $({\w}_L,\dots,{\w}_L)$ are all identical and then ${u}(g_j\x^{(0)})={u}(\x^{(0)})$ which means ${u}(\x^{(0)})$ invariant under group $G$. The proof ends. \hfill $\square$

Furthermore, if the finite group $G$ has matrix representation, the weight matrix of the first hidden layer of sDNN can be rewritten in a compact form, which avoids the extension of the input data $\x^{(0)}$ to $X^{(0)}$ in (\ref{mat-ge1}).
\begin{cor}\label{th-cyclic}
Under the same conditions in Theorem \ref{th-general}, and suppose that each $g_l\in G$ has the form $g_l\x=T_{g_l} \x$ and $g_lu=u$, where $T_{g_l}$ is a matrix representation of $g_l$. Then the neural networks with $L-1$ hidden layers and $n_l$ neurons per layer is invariant under $G$ if the following conditions hold:

1). For the first hidden layer, the weight matrix $\W_1$ and the bias vector $\B_1$ take the form
\begin{eqnarray}\label{mat-cy1}
&&\W_1=\left(
            \begin{array}{c}
              {\w}_1 \\ {\w}_1\,T_{g_1}\\ \vdots \\ {\w}_1\,T_{g_{n-1}}\\
              \end{array}
          \right),
\hspace{1cm}\B_1=\left(
            \begin{array}{c}
            {\b}_1\\ {\b}_1\\ \vdots \\ {\b}_1 \\
            \end{array}
          \right).
\end{eqnarray}

2). For the $l$th ($1< l< L$) hidden layer, the weight matrix $\W_l$ and the bias vector $\B_l$ of sDNN take the same forms as (\ref{mat-ge2}), and for the output layer $\W_L$ and $\B_L$ are identical with (\ref{mat-ge3}) in Theorem \ref{th-general}.
\end{cor}

\emph{Proof.} Following the notations and steps in the proof of Theorem \ref{th-general}, for the outputs of the first layer, Eq.(\ref{mat-ger11}) becomes
\begin{eqnarray}\label{mat-cyc1}
&&\no\x^{(1)}\big(g_i\x^{(0)}\big)=\sigma\left({\w}_1\,T_{g_i}\x^{(0)} +\b_1\right)=\sigma\left(\big({\w}_1\,T_{g_i}\big)\x^{(0)} +\b_1\right).
\end{eqnarray}
Moreover, because the outputs of the first hidden layer $S_1:=\big\{\x^{(1)}\big(g_i\x^{(0)}\big)|i=0,\dots,n-1\big\}$ is invariant under $G$, we expand the weight matrix ${\w}_1$ to the form $\W_1$ and the bias vector $\b_1$ to $\B_1$ in (\ref{mat-cy1}) by the order of $G$. The proof ends. \hfill $\square$

\begin{figure}[htp]
\centering
	\begin{minipage}{0.8\linewidth}
		\centerline{\includegraphics[width=\textwidth]{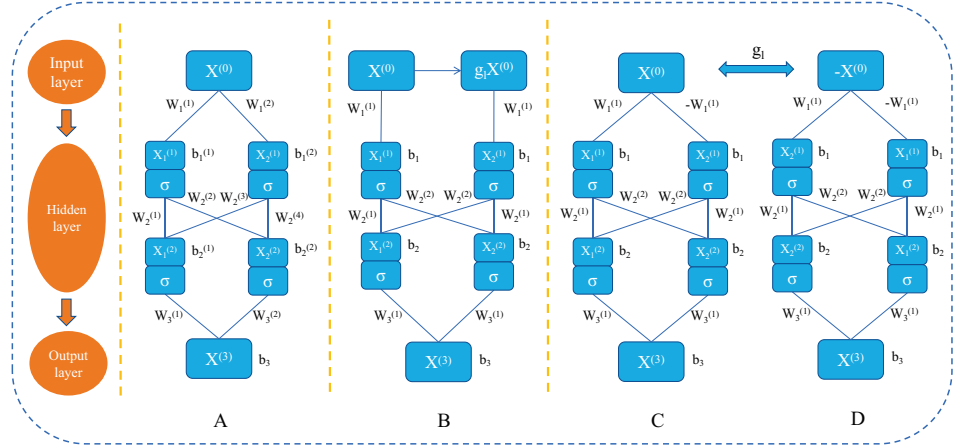}}
	\end{minipage}
	\caption{(Color online) Schematic diagrams of the PINN, sDNN with and without matrix representation: (A) PINN. (B) sDNN without matrix representation. (C) sDNN with matrix representation. } 
\label{fig-s}
\end{figure}

Next, we take a neural network with two hidden layers and equipped with a second order finite group $G_e=\{g_0,g_1\}$ as an example to show the main ideas of Theorem  \ref{th-general} and Corollary \ref{th-cyclic}, where $g_0$ is the identity element and $g_1:(x,t,u)\rightarrow(-x,-t,u)$ denotes the even symmetry transformation. 
Figure \ref{fig-s}(A) shows the vanilla PINN whose output is $X^{(3)}=\sigma\left(\W_2\sigma\left(\W_1X^{(0)}+\b_1\right)+\b_2\right)+\b_3$, where
 \begin{eqnarray}
&&\no \W_1=\left(
  \begin{array}{c}
    \W_1^{(1)} \\
    \W_1^{(2)}\\
  \end{array}
\right),~\W_2=\left(
             \begin{array}{cc}
               \W_2^{(1)} & \W_2^{(2)} \\
               \W_2^{(3)} & \W_2^{(4)} \\
             \end{array}
           \right),~\W_3=\left(
  \begin{array}{c}
    \W_3^{(1)} \\
    \W_3^{(2)}\\
  \end{array}
\right).
\end{eqnarray}

On the other hand, Figure \ref{fig-s}(B) displays the framework of the sDNN with $G_e$ by means of Theorem \ref{th-general}, where the weight matrixes are given by
 \begin{eqnarray}\label{matrix}
&& \W_1=\left(
  \begin{array}{c}
    \W_1^{(1)} \\
    \W_1^{(1)}\\
  \end{array}
\right),~\W_2=\left(
             \begin{array}{cc}
               \W_2^{(1)} & \W_2^{(2)} \\
               \W_2^{(2)} & \W_2^{(1)} \\
             \end{array}
           \right),~\W_3=\left(
  \begin{array}{c}
    \W_3^{(1)} \\
    \W_3^{(1)}\\
  \end{array}
\right),
\end{eqnarray}
then the sDNN is invariant under $G_e$. In fact, acting $G_e$ on the input data $X^{(0)}$, we obtain a new input data set $\{g_0X^{(0)},g_1X^{(0)}\}=\{X^{(0)},-X^{(0)}\}$, then the outputs of the first hidden layer are
 \begin{eqnarray}
&&\no X^{(1)}_1=\sigma\big(\W_1^{(1)}X^{(0)}+\b_1\big),\\
&&\no X^{(1)}_2=\sigma\big(\W_1^{(1)}(gX^{(0)})+\b_1\big)=\sigma\big(\W_1^{(1)}(-X^{(0)})+\b_1\big).
\end{eqnarray}

It is obvious that if $X^{(0)}$ becomes $-X^{(0)}$, then $X^{(1)}_1$ becomes $X^{(1)}_2$ and $X^{(1)}_2$ becomes $X^{(1)}_1$, thus the set $\{X^{(1)}_1,X^{(1)}_2\}$ is invariant under $G$.
While the outputs of the second hidden layer are
 \begin{eqnarray}
&&\no\hspace{-1.2cm} X^{(2)}_1=\sigma\big(\W_2^{(1)}X_1^{(1)}+\W_2^{(1)}X_2^{(1)}+\b_2\big),\\
&&\no \hspace{-1.2cm}X^{(2)}_2=\sigma\big(\W_2^{(1)}X_2^{(1)}+\W_2^{(1)}X_1^{(1)}+\b_2\big).
\end{eqnarray}
Then under the conditions $X^{(1)}_1\rightarrow X^{(1)}_2$ and $X^{(1)}_2\rightarrow X^{(1)}_1$, we get
 \begin{eqnarray}
&&\no X^{(2)}_1=\sigma\big(\W_2^{(1)}X_1^{(1)}+\W_2^{(1)}X_2^{(1)}+\b_2\big)\\
&&\no\hspace{0.85cm} \xrightarrow[X^{(1)}_2\rightarrow X^{(1)}_1]{X^{(1)}_1\rightarrow X^{(1)}_2}\sigma\big(\W_2^{(1)}X_2^{(1)}+\W_2^{(1)}X_1^{(1)}+\b_2\big)=X^{(2)}_2,\\
&&\no X^{(2)}_2=\sigma\big(\W_2^{(1)}X_2^{(1)}+\W_2^{(1)}X_1^{(1)}+\b_2\big)\\
&&\no\hspace{0.85cm}\xrightarrow[X^{(1)}_2\rightarrow X^{(1)}_1]{X^{(1)}_1\rightarrow X^{(1)}_2}\sigma\big(\W_2^{(1)}X_1^{(1)}+\W_2^{(1)}X_2^{(1)}+\b_2\big)=X^{(2)}_1,
\end{eqnarray}
thus the set $\{X^{(2)}_1,X^{(2)}_2\}$ is invariant under $G$. Finally, the output of sDNN is
 \begin{eqnarray}
&&\no \hspace{-1.9cm}X^{(3)}=\W_3^{(1)}X_2^{(2)}+\W_3^{(1)}X_1^{(2)}+\b_3.
\end{eqnarray}
Again, under the conditions $X^{(2)}_1\rightarrow X^{(2)}_2$ and $X^{(2)}_2\rightarrow X^{(2)}_1$, $X^{(3)}$ is still invariant under $G_e$ since the coefficients of $X_1^{(2)}$ and $X_2^{(2)}$ are identical. Therefore, the sDNN in Figure \ref{fig-s}(B) is invariant under $G_e$.

Alternatively, observed that the finite group $G_e$ has the matrix representation
\begin{eqnarray}\label{matrix-even}
&& g_0=\left(\begin{array}{cc}
               1 & 0 \\
               0 & 1\\
             \end{array}
           \right),~
           g_1=\left(\begin{array}{cc}
              -1 & 0 \\
               0 & -1\\
             \end{array}
           \right),
\end{eqnarray}
then, by Corollary \ref{th-cyclic}, the input data set $X^{(0)}$ is not changed but $\W_1=\big(\W_1^{(1)},-\W_1^{(1)}\big)^T$, $\W_2$ and $\W_3$ are given by (\ref{matrix}). In fact, as shown in Figure \ref{fig-s}(C), the outputs of the first hidden layer are
 \begin{eqnarray}
&&\no X^{(1)}_1=\sigma\big(\W_1^{(1)}X^{(0)}+\b_1\big),\\
&&\no X^{(1)}_2=\sigma\big(\W_1^{(1)}(-X^{(0)})+\b_1\big).
\end{eqnarray}
Then the set $\{X^{(1)}_1,X^{(1)}_2\}$ is invariant under $G_e$ with a similar proving method in Figure \ref{fig-s}(B). Moreover, since the remainder of sDNN with matrix representation is identical to the sDNN without matrix representation, thus the sDNN in Figure \ref{fig-s}(C-D) is also invariant under $G_e$.

Furthermore, it is observed that the weight matrix and bias vector in each hidden layer of sDNN have particular structures while the ones of PINN are usually initialized by the Xavier initialization \cite{xw-2010}, thus the number of training parameters of sDNN is decreased when compared with the PINN of same network architecture.

\begin{cor}
Suppose that a PINN consists of $L-1$ hidden layers with $n_l$ neurons in the $l$-th hidden layer and $|G|$ denotes the order of finite group $G$. Then comparing with the PINN of same network architecture for solving Eq.(\ref{eqn1}), the number of training parameters by sDNN decreases by $1+\big[2n_1+n_L+\sum_{l=1}^{L-1}(n_ln_{l+1}+n_l)\big]/|G|$.
\end{cor}

\emph{Proof.} From the proofs of Theorem \ref{th-general} and Corollary \ref{th-cyclic}, the number of neurons $n_l$ must be an integral multiple of $|G|$. By the direct computations, the number of parameters for the weight matrixes is $(2 n_1+n_L)/|G|+\sum_{l=1}^{L-1}n_ln_{l+1}/|G|$ while the one for the bias vectors is $1+\sum_{l=1}^{L}n_l/|G|$, then adding the two parts gives the total number of training parameters of sDNN. The proof ends. \hfill $\square$

Comparatively, the total number of PINN is $1+2n_1+n_L+\sum_{l=1}^{L-1}( n_ln_{l+1}+n_l)$ which is about $|G|$ times of the number of sDNN. Thus, the higher the order of $G$, the fewer training parameters are required.

\subsection{Approximation Theory for sDNN}
In what follows, we consider the universal approximation ability of the proposed sDNN in Theorem \ref{th-general}.
One fundamental question related to PINN and its various improved versions, including sDNN here, is whether there exists a neural network
that can simultaneously and uniformly approximate a function and its partial derivatives with arbitrary orders.
Only when sDNN has this ability, the loss function may tend to zero, and then the approximate solution of PDE can be solved.
To study this question and simplify expressions, it is convenient to first introduce some notation.

Let $\mathbb{N}^{p}$ be the set of $p$-dimensional nonnegative integers and $\mathbb{R}$ be the real number.
Let $\mathbf{m}=(m_1,m_2,\cdots,m_p)\in \mathbb{N}^{p}$, $\mathbf{k}=(k_1,k_2,\cdots,k_p)\in \mathbb{N}^{p}$ and $\x=(x_1,x_2,\cdots,x_p)\in \mathbb{R}^{p}$,
we set $|\mathbf{m}|=m_1+\cdots+m_p$, $\mathbf{m}!=\prod_{i=1}^{p}m_i!$ and $\mathcal{D}_{\x}^{\mathbf{m}}=\partial^{\,|\mathbf{m}|}/\partial x_1^{m_1}\ldots \partial x_p^{m_p}$. We define the usual ordering on $\mathbb{N}^{p}$,namely $\mathbf{k}\leq \mathbf{m}$ if $k_i\leq m_i,~ i=1,2,\cdots,p$.
A function $f$ is said to belong to $\mathcal{C}^{\mathbf{m}}(\mathbb{R}^{p})$ if $\mathcal{D}_{\x}^{\mathbf{k}}f \in \mathcal{C}(\mathbb{R}^{p})$
for all $\mathbf{k}\leq \mathbf{m} $, where $\mathcal{C}(\mathbb{R}^{p})=\{h:\mathbb{R}^{p}\rightarrow\mathbb{R}\mid h~\text{is continuous}\}$
is the space of continuous functions. If $\mathbf{k}=\mathbf{0}$, then we define $\mathcal{D}_{\x}^{\mathbf{k}}f=f(\x)$.

It is widely known that the neural networks with one hidden layer and sufficiently number of neurons have
the expressive ability of approximating any functions and its partial derivatives with arbitrary orders.
We recall the following theorem of universal derivative approximation using single hidden layer neural networks due to Pinkus \cite{Pin-1999}.

\begin{theorem} [Universal derivative approximation theory \cite{Pin-1999}]\label{th-univer}Given any compact set $V\subset \mathbb{R}^{p}$, any $\epsilon>0$ and
given any  function $f \in \mathcal{C}^{\mathbf{m}}(\mathbb{R}^{p}): V \rightarrow \mathbb{R}$, there exists a neural network
$\widehat{f}\in \mathcal{C}^{\mathbf{m}}(\mathbb{R}^{p}): V\rightarrow \mathbb{R}$ with one hidden layer, such that
\begin{eqnarray}
&&\no\left |\mathcal{D}^{\mathbf{k}}f(\x)-\mathcal{D}^{\mathbf{k}}\widehat{f}(\x)\right|\leq \epsilon, ~~~\forall~ \x \in V and ~\forall~ \mathbf{k}\leq \mathbf{m}.
\end{eqnarray}
\end{theorem}

However, it is not straightforward that the proposed sDNN enjoys the universal derivative approximation ability of learning the solution of the PDEs having the finite group. For the completeness of the paper, we show that such a sDNN can approximate any functions and its partial derivatives with arbitrary orders preserving a finite group with an arbitrary accuracy.

In order to show the universal derivative approximation of sDNN, we need to obtain an explicit expression for an arbitrary partial derivative of a composition of function
$h(\x)=f(u_1(\x),u_2(\x),\cdots,u_p(\x))$ in terms of the various partial derivatives of the function $f$ and $u_1,u_2,\cdots,u_p$ where $\mathbf{u}=(u_1,u_2,\cdots,u_p)$, $f(\mathbf{u})$ and $u_1(\x),\cdots,u_p(\x)$ are differentiable with a sufficient number of times.
We recall the following theorem of multivariate Fa\`{a} di Bruno formula due to Constantine \cite{Con-1996}.

\begin{lemma}[Multivariate Fa\`{a} di Bruno formula \cite{Con-1996}]\label{fda-formula}
Let $h=f(\mathbf{ u}(\x))$ be a composite function. Then the $|\mathbf{m}|$-order derivative of $h$ with respective to $\x$ is given by
\begin{eqnarray}
&& \mathcal{D}_{\x}^{\mathbf{m}}h=\sum_{1 \leq|\mathbf{\lambda}|\leq m} \mathcal{D}_{\mathbf{ u}}^{\mathbf{\lambda}}f\,
\mathbf{\varphi}_{\mathbf{m},\mathbf{\lambda}}(\x),
\end{eqnarray}
where
\begin{eqnarray}
&&\no\mathbf{\varphi}_{\mathbf{m},\mathbf{\lambda}}(\x)=\mathbf{m}!\sum_{q(\mathbf{m},\mathbf{\lambda})} \prod_{j=1}^{m}\frac{(\mathcal{D}_{\x}^{\mathbf{\beta}_{j}}u_1)^{\alpha_{j_1}} (\mathcal{D}_{\x}^{\mathbf{\beta}_{j}}u_2)^{\alpha_{j_2}}\cdots(\mathcal{D}_{\x}^{\mathbf{\beta}_{j}}u_p)^{\alpha_{j_p}}}
{\mathbf{\alpha}_{j}!(\mathbf{\beta}_{j}!)^{|\mathbf{\alpha}_{j}|}},
\end{eqnarray}
and
\begin{eqnarray}
&&\no m=|\mathbf{m}|,\mathbf{\alpha}_{j}=\{\alpha_{j_1}, \alpha_{j_2},\cdots,\alpha_{j_p}\};\\
&&\no q(\mathbf{m},\mathbf{\lambda})=\Big\{(\mathbf{\alpha}_{1},\cdots,\mathbf{\alpha}_{m}; \mathbf{\beta}_{1},\cdots,\mathbf{\beta}_{m}): for~ some~ 1\leq s \leq m,\\
&&\no\hspace{2cm}\mathbf{\alpha}_{i}=0 ~and~  \mathbf{\beta}_{i}=0 ~for~ 1\leq i \leq m-s; \\
&&\no\hspace{2cm}|\mathbf{\alpha}_{i}|> 0~ for~ m-s+1 \leq i \leq m; \\
&&\no\hspace{2cm}and~ \mathbf{0}\leq \mathbf{\beta}_{m-s+1}\leq \cdots \leq \mathbf{\beta}_{m} ~are~ such ~that \\
&&\no\hspace{2cm} \sum_{i=1}^{m}\mathbf{\alpha}_{i}=\mathbf{\lambda}, \sum_{i=1}^{m}|\mathbf{\alpha}_{i}|\mathbf{\beta}_{i}=\mathbf{m}\Big\}.
\end{eqnarray}
\end{lemma}
\begin{theorem} Let $G:=\{g_0(=e),g_1,g_2,\dots,g_{n-1}\}$ be a finite group of order $n$. Suppose any $g\in G$ and $g\x=(\mathbf{u}_1^{g}(\x),\mathbf{u}_2^{g}(\x),\cdots,\mathbf{u}_p^{g}(\x))$ where $\mathbf{u}_i^{g}(\x)\in \mathcal{C}^{\mathbf{m}}(\mathbb{R}^{p}), i=1,2,\cdots,p$.
Given any compact set $V\subset \mathbb{R}^{p}$ and any $G$-invariant map
$f\in \mathcal{C}^{\mathbf{m}}(\mathbb{R}^{p}): V\rightarrow \mathbb{R}$, then the map $f$ and its partial derivatives can be approximated
by the sDNN with one hidden layer $\widehat{f}:V\rightarrow \mathbb{R}$ of the form
\begin{eqnarray}\label{fun-1}
&& \widehat{f}(\x)=\sum_{s=0}^{n-1}\w_2\,\sigma\big(\w_1 (g_s\x)+\b_1\big),
\end{eqnarray}
and its corresponding partial derivatives respectively, where $(\w_1)_{k\times p}$ and $(\w_2)_{1\times k}$ are weight matrixes of the first and output layers, and $\b_1$ is the bias vector of sDNN with one hidden layer and $k$ neurons per layer.
\label{th-group}\end{theorem}

\emph{Proof.} The proof is divided into two parts where one is to prove that the map $\widehat{f}(\x)$ is $G$-invariant and the other is to prove that any $G$-invariant map $f(\x)$ and its partial derivatives can be approximated by the $G$-invariant neural network $\widehat{f}(\x)$ and its corresponding partial derivatives. 

$\bullet$ In the first part, one has
\begin{eqnarray}\label{equlity-1}
&&\no\widehat{f}(g_j\x)=\sum_{s=0}^{n-1}\w_2\,\sigma\big(\w_1 (g_sg_j\x)+\b_1\big)\\
&&\no\hspace{1.5cm}=\sum_{s=0}^{n-1}\w_2\,\sigma\big(\w_1(g_sg_j)\x+\b_1\big)\\
&&\no\hspace{1.5cm}=\sum_{r=0}^{n-1}\w_2\,\sigma\big(\w_1 g_r\x+\b_1\big),
\end{eqnarray}
thus $\widehat{f}(g_j\x)=\widehat{f}(\x)$. Note that the last equation holds because $g_r=g_sg_j\in G$ for any $g_j,g_s \in G$.

$\bullet$ For the second part, we first prove that the map $f(\x)$ and its partial derivatives can be approximated by a $G$-invariant function
and its corresponding partial derivatives.
Let $V$ be a compact set in $\mathbb{R}^{p}$ and consider the symmetrized set $V_{sym}=\big\{\bigcup_{g\in G} g\x|\x \in V\big\}$ which is also a compact set in $\mathbb{R}^{p}$ under the finite group $G$. By Theorem \ref{th-univer}, one gets a neural network $\widetilde{P}(\x)$ which approximates $G$-invariant $f(\x)$ on $V_{sym}$, i.e.
$\left |\mathcal{D}^{\mathbf{k}}f(\x)-\mathcal{D}^{\mathbf{k}}\widetilde{P}(\x)\right|\leq \epsilon$ for any $\epsilon>0$ and any $\mathbf{k}\leq \mathbf{m}$. Then by means of the Reynolds operator \cite{stu-2008}, we obtain $P_{sym}(\x) = \sum_{g\in G}\widetilde{P}(g\x)/n$ where the order of $G$ is $n$.

Let $\widetilde{h}(\x)=\widetilde{P}(g\x)=\widetilde{P}(\mathbf{u}_1^{g}(\x),\mathbf{u}_2^{g}(\x),\cdots,\mathbf{u}_p^{g}(\x))$ and  $h(\x)=f(g\x)=f(\mathbf{u}_1^{g}(\x),\mathbf{u}_2^{g}(\x),$ $\cdots,\mathbf{u}_p^{g}(\x))$. By Lemma \ref{fda-formula}, one gets
\begin{eqnarray}
&&\no \hspace{-0.95cm}|\mathcal{D}_{\x}^{\mathbf{k}}P_{sym}(\x)-\mathcal{D}_{\x}^{\mathbf{k}}f(\x)|=\left|\mathcal{D}_{\x}^{\mathbf{k}}\left(\frac{1}{n}\sum_{g\in G}\widetilde{P}(g\x)\right)-\mathcal{D}_{\x}^{\mathbf{k}}\left(\frac{1}{n}\sum_{g\in G}f(g\x)\right)\right|\\
&&\no \hspace{2.9cm}\leq \frac{1}{n}\sum_{g\in G} \left|\mathcal{D}_{\x}^{\mathbf{k}}\left(\widetilde{h}(\x)\right)-\mathcal{D}_{\x}^{\mathbf{k}}\left(h(\x)\right)\right|\\
&&\no \hspace{2.9cm}=\frac{1}{n}\sum_{g\in G} \left|\sum_{1 \leq|\mathbf{\lambda}|\leq m} \mathcal{D}_{\mathbf{ u}}^{\mathbf{\lambda}}\widetilde{P} \mathbf{\varphi}_{\mathbf{m},\mathbf{\lambda}}(\x)-\sum_{1 \leq|\mathbf{\lambda}|\leq m} \mathcal{D}_{\mathbf{ u}}^{\mathbf{\lambda}}f \mathbf{\varphi}_{\mathbf{m},\mathbf{\lambda}}(\x)\right|\\
&&\no \hspace{2.9cm}=\frac{1}{n}\sum_{g\in G} \left|\sum_{1 \leq|\mathbf{\lambda}|\leq m}\left( \mathcal{D}_{\mathbf{ u}}^{\mathbf{\lambda}}\widetilde{P}-
\mathcal{D}_{\mathbf{ u}}^{\mathbf{\lambda}}f\right) \mathbf{\varphi}_{\mathbf{m},\mathbf{\lambda}}(\x)\right|\\
&&\no \hspace{2.9cm}\leq \frac{1}{n}\sum_{g\in G} \sum_{1 \leq|\mathbf{\lambda}|\leq m}\mathbf{\varphi}_{\mathbf{m},\mathbf{\lambda}}(\x) \epsilon\\
&&\no \hspace{2.9cm}\leq \mathbf{M}\epsilon,
\end{eqnarray}
where $f(g\x)=f(\x)$ since $f(\x)$ is a $G$-invariant map and $\mathbf{M}=\max_{\x\in V}\left\{\frac{1}{n}\sum_{g\in G} \sum_{1 \leq|\mathbf{\lambda}|\leq m}\mathbf{\varphi}_{\mathbf{m},\mathbf{\lambda}}(\x) \right\}$ since $\mathbf{u}_i^{g}(\x)\in \mathcal{C}^{\mathbf{m}}(\mathbb{R}^{p})$.


It remains to prove that $P_{sym}(\x)$ is a candidate of the $G$-invariant function $\widehat{f}(\x)$. Recall that
\begin{eqnarray}
&&\no P_{sym}(\x)=\frac{1}{n}\sum_{g\in G}\widetilde{P}(g\x)\\
&&\no\hspace{1.5cm}=\frac{1}{n}\sum_{s=0}^{n-1}\widetilde{\w}_2\,\sigma\big(\widetilde{\w}_1(g_s\x)+\b_1\big)\\
&&\no\hspace{1.5cm}=\sum_{s=0}^{n-1}\frac{\widetilde{\w}_2}{n}\,\sigma\big(\widetilde{\w}_1 (g_s\x)+\b_1\big).
\end{eqnarray}
Then defining $\w_2=\widetilde{\w}_2/n$ and $\w_1=\widetilde{\w}_1$, we obtain the required $\widehat{f}(\x)=P_{sym}(\x)$ in (\ref{fun-1})
such that $|\mathcal{D}^{\mathbf{k}}\widehat{f}(\x)-\mathcal{D}^{\mathbf{k}}f(\x)| \leq \mathbf{M}\epsilon$. The proof ends. \hfill $\square$

\subsection{Algorithm of sDNN}
Based on the above theoretical preparations, we take Eq.(\ref{eqn1}) and its initial and boundary conditions (\ref{ib}) as the research object to present the algorithm of sDNN for solving PDEs. The framework of sDNN is introduced in Algorithm \ref{alg:Framwork} which is built on the basis of the framework of PINN \cite{2018a}. Furthermore, if the finite group has a matrix representation just as stated in Corollary \ref{th-cyclic}, then the weight matrixes and bias vectors in Step 2 of Algorithm \ref{alg:Framwork} can be replaced by (\ref{mat-cy1}). In particular, finding a finite group of PDEs has several mature algorithms \cite{peh-2000,Dor-2011}. For example, an algorithm was proposed to calculate the discrete point symmetries of a given PDE with a known nontrivial group of Lie point symmetries \cite{peh-2000}.
We note that in the real experiments the sDNN is implemented as a standard feed-forward NNs by subsequently expanding the weight matrixes and bias vectors of each hidden layer in the PINN by means of the Cayley table of finite group \cite{dk-2015}, but the training parameters of the two methods are identical.
\begin{algorithm}[htp]
\caption{: Framework of sDNN}
  \label{alg:Framwork}
  \begin{algorithmic}[1]
    \Require
      The set of initial and boundary points $\mathcal {S}_{ib}=\{\x_i\}_{i=1}^{N}$; The set of collocation points $\mathcal {S}_{u}=\{\x_i\}_{i=1}^{\widetilde{N}}$;\,
     \Ensure
      $L_2$ relative errors of learned solution $\widetilde{u}(x,t)$ in the sampling domain and prediction domain;
    \State Find a finite group, $G:=\{g_i|\x^*=g_i(\x),u^*=u; i=0,\dots,n-1\}$, which leaves Eq.(\ref{eqn1}) invariant.
    \State Initialize the weight matrixes ${\w}_i$ and bias vectors $\b_i$ $(i=1,\dots,L)$ by the Xavier initialization method; Construct the neural networks with weight matrixes and bias vectors as (\ref{mat-ge1}), (\ref{mat-ge2}) and (\ref{mat-ge3}), respectively.
    \State Let $\x^{(0)}$ be the input data choosing from $\mathcal {S}_{ib}\cup \mathcal {S}_{u}$ randomly. Expand $\x^{(0)}$ to $\mathcal {S}_x=\{g_i\x^{(0)}|i=0,\dots,n-1\}$.
    \State Train the neural networks with the loss function (\ref{invaloss}).

    \hspace{-0.9cm} \Return  Learned solution $\widetilde{u}(x,t)$ and its $L_2$ relative error in the sampling domain and prediction domain.
  \end{algorithmic}
\end{algorithm}

The proposed sDNN has three obvious novelties. The first is the new strategy of integrating neural networks and physical properties of PDEs.
Different with the traditional neural networks methods for incorporating physical properties of PDEs into the loss function, such as the gradient-enhanced PINN \cite{J}, symmetry-enhanced PINN \cite{zhang-2023a}, or utilizing the symmetry group to generate labeled data in the interior domain of PDEs \cite{zhang-2023b}, the proposed sDNN directly integrates the finite group into the architecture of neural networks which makes the neural networks itself keep the finite group exactly.

The second is the highly learning ability of sDNN in the sampling domain. Numerical experiments below show that the sDNN can well learn the solution than PINN with fewer training points and simpler neural networks. Moreover, the learned solutions by the sDNN preserve the symmetry properties well in and beyond the sampling domains, but the PINN does not have this feature even though it has a good accuracy in the sampling domain.

The third is the strong solution extrapolation ability of sDNN, i.e. the learned solutions in and beyond the sampling domain have almost same predicted accuracies, while the PINN method usually fails to predict the solution beyond the sampling domain even though the learned solution in the sampling domain has high-accuracy. Consequently, the solution beyond the sampling domain predicted by sDNN has one or more orders of magnitude improvement than the one by PINN. Of course, the solution extrapolation heavily depends on the accuracies of learned solution in the sampling domain where the bad learned results directly lead to the failure of the prediction beyond the sampling domain. This is reasonable since the prediction beyond the sampling domain is closely based on the learned results in the sampling domain.
\section{Numerical results}
To demonstrate the capability and efficiency of the introduced sDNN, we will compare the sDNN with the vanilla PINN \cite{2018a} by considering several usual finite groups with physical meanings, where the first two examples focus on the advection equation with even symmetry and the sine-Gordon equation with circulant symmetry which display the efficacy of sDNN, the third example aims at effects of different order finite group for learning solutions of the Poisson equation, and the last two examples are the nonlinear wave equation and the Korteweg-de Vries (KdV) equation whose corresponding finite groups have no matrix representation.

Furthermore, all reported errors are the $L_2$ relative errors computed by $\|u-\widehat{u}\|_2/\|u\|_2$, where $\widehat{u}$ is the model prediction, $u$ is the reference solution, and $\|\cdot\|_2$ denotes the 2-norm. In the experiments of sDNN and PINN methods, we utilize the Xavier initialization \cite{xw-2010} to generate the initialized weight matrixes and bias vectors, and employ a hyperbolic tangent activation function. Meanwhile, we choose the collocation points in the interior domain of PDEs by the Latin hypercube sampling \cite{ms-1987} and the initial and boundary training points randomly, and compute the derivatives by the automatic differentiation \cite{auta}, and deploy the L-BFGS optimization algorithm \cite{ln-1989} to minimize the loss function (\ref{invaloss}).

\subsection{Advection equation}
Consider the advection equation
\begin{eqnarray}\label{advection}
u_t+c\,u_x=0,~~~x\times t\in[-2,2]\times[-1,1],
\end{eqnarray}
which describes the evolution of the scalar field $u(x,t)$ carried along by flow at constant speed $c$ \cite{ft-1997}. Here we take $c=2$ and choose the first quadrant $x\times t\in[0,2]\times[0,1]$ as the sampling domain and predict the solution $u=(x-2t)^4$ in the first and third quadrants by the sDNN and PINN methods. Since Eq.(\ref{advection}) is admitted by an even symmetry $(x,t,u(x,t))\mapsto(-x,-t,u(x,t))$, i.e. a second-order finite group $G_e=\{g_0,g_1\}$ after adding the identity element $g_0$, where
\begin{eqnarray}
&&\no g_0=\left(\begin{array}{cc}
               1 & 0 \\
               0 & 1\\
             \end{array}
           \right),~
           g_1=\left(\begin{array}{cc}
              -1 & 0 \\
               0 & -1\\
             \end{array}
           \right),
\end{eqnarray}
then following Corollary  \ref{th-cyclic}, we give the architecture of sDNN for Eq.(\ref{advection}) as follows.
\begin{prop}
[Even neural network]
Suppose that the group $G_e$ keeps Eq.(\ref{advection}) invariant. Then the weight matrixes and bias vectors of the sDNN with $L-1$ hidden layers which keeps $G_e$ invariant takes the form
\begin{eqnarray}
&&\no \hspace{-0.7cm}\text{Weight matrixes:}~\left(
  \begin{array}{c}
    \w_1 \\
    -\w_1\\
  \end{array}
\right),~\left(
             \begin{array}{cc}
               \w_2^{(1)} & \w_2^{(2)} \\
               \w_2^{(2)} & \w_2^{(1)} \\
             \end{array}
           \right),~\dots,~\left(
             \begin{array}{cc}
               \w_{L-1}^{(1)} & \w_{L-1}^{(2)} \\
               \w_{L-1}^{(2)} & \w_{L-1}^{(1)} \\
             \end{array}
           \right),~\left(
  \begin{array}{c}
    \w_L \\
   \w_L\\
  \end{array}
\right),\\
&& \no\text{Bias vectors:}~ \left(
  \begin{array}{c}
    \b_1 \\
    \b_1\\
  \end{array}
\right),~\left(
  \begin{array}{c}
    \b_2 \\
    \b_2\\
  \end{array}
\right),~\dots,~\left(
  \begin{array}{c}
    \b_{L-1} \\
    \b_{L-1}\\
  \end{array}
\right),\left(
  \begin{array}{c}
    \b_{L} \\
  \end{array}
\right),
\end{eqnarray}
where  $\w_1, \w_l^{(1)}, \w_l^{(2)}$ and $\w_L$ are the initialized weight matrixes and $\b_l$ are the bias vectors for the PINN, $l=2,\dots,L-1$.
\label{prop-adv}
\end{prop}

Proposition \ref{prop-adv} is a direct consequence of Corollary \ref{th-cyclic}. Moreover, the number of training parameters in sDNN increase $\sum_{i=1}^{L-1}n_in_{i+1}$ against the PINN, where $n_i$ is the number of neurons in the $i$th hidden layer. Thus we will adjust the number of neurons in PINN to make the comparisons of sDNN and PINN be fair. Furthermore, by the sDNN with $G_e$, we have several choices to employ half of the domain as the sampling domain to predict the other half. For example, we can sample in the first quadrant to predict the third quadrant, or sample second quadrants and predict the fourth quadrant, or sample in both first and second quadrants and predict the third and fourth quadrants. Here, we perform an experiment by using the first quadrant as a sampling domain to predict the solution in the third quadrant.

To obtain the training data set, we discretize the spatial region $x \in[0,2]$ as $N_x=201$ and the temporal region $t \in[0,1]$ as $N_t=101$ discrete equidistant points respectively, which makes the solution $u$ be discretized as $201\times101$ data points in the domain $[0,2]\times[0,1]$. 
Then we perform two groups of experiments to investigate the effects of sDNN and PINN for learning solution of Eq.(\ref{advection}) in and beyond the sampling first quadrant. In particular, we define an even metric, i.e. $S=\left\|u(x,t)_{pred}-u(-x,-t)_{pred}\right\|_{2}/\left\|u(x,t)_{pred}\right\|_{2}$, to measure the degree of the learned solutions keeping the even symmetry, with $S=0$ meaning strictly satisfying even symmetry.

$\bullet$ \emph{Performances for different number of collocation points.}
We first consider how the different number $N_f$ collocation points affect the accuracies of learned solutions by the sDNN and PINN. Specifically, we choose $3$ hidden layers with $10$ neurons per layer for the sDNN and $3$ hidden layers with $10$ neurons in the first and third layers and $20$ neurons in the second layer for the PINN in order to make the two methods train with the same number of parameters. Moreover, the number of collocation points varies from $100$ to $1500$ with step $100$ and $N_u=100$ initial and boundary points are used.
\begin{figure}[htp]
	\begin{minipage}{\linewidth}
		\centerline{\includegraphics[width=\textwidth]{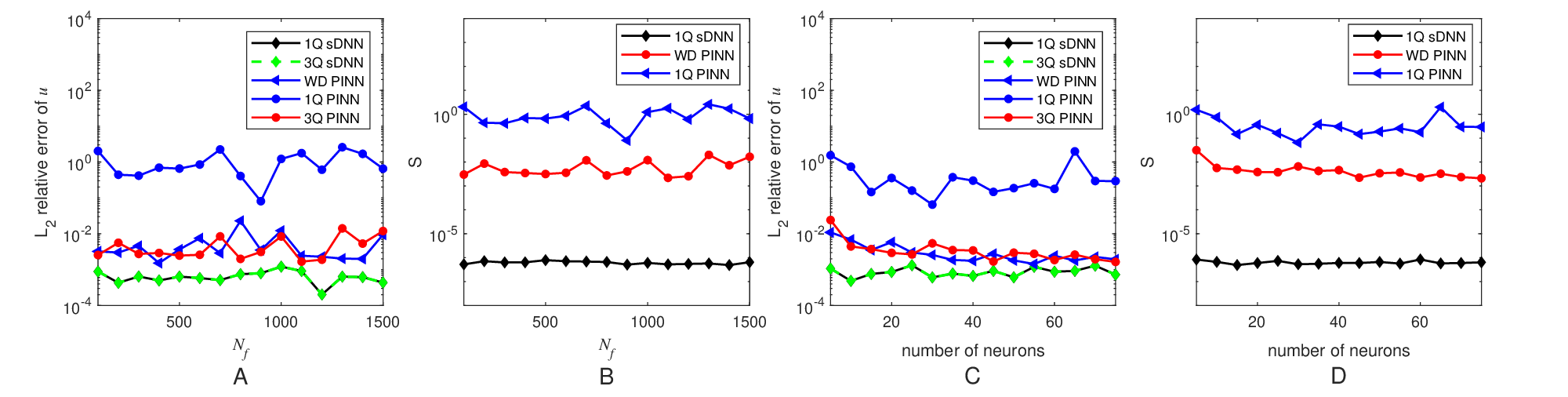}}
	\end{minipage}
	\caption{(Color online) Advection equation: Comparisons of $L_2$ relative errors and the even metric of $u$ in the sampling and prediction domains by the two methods. Keeping the number of neurons invariant and varying the number of collocation points: (A) the $L_2$ relative errors; (B) the symmetry metrics. Keeping the number of collocation points invariant and varying the number of neurons: (C) the $L_2$ relative errors; (D) the symmetry metrics. Each line represents the mean of five independent experiments. Note that the sDNN and PINN with subscript $1$ and $2$ of denote the sampling domain and prediction domain respectively, and the symbol `WD PINN' means the training in the whole domain.}
\label{fig2}
\end{figure}

Figure \ref{fig2}(A) shows that the blue solid line denoting the $L_2$ relative errors of PINN in the sampling first quadrant fluctuates at the $10^{-3}$ order of magnitude where the minimum is $1.52\times10^{-3}$, while the blue dotted line in the prediction domain distributes at $10^{-1}$ which means that the PINN has poor solution extrapolation ability. 
Comparatively, the black solid line and green dashed line completely coincide which demonstrates that the $L_2$ relative errors of sDNN in the sampling and prediction domains keep almost the same accuracies and their maximal difference is $6.70\times10^{-9}$. In fact, the accuracies of learned solutions by sDNN in and beyond the sampling domain mostly stays at $10^{-4}$ and one order of magnitude improvement than the PINN. 
Moreover, in Figure \ref{fig2}(B), the black solid line for the even metric $S$ of sDNN is stable at $10^{-7}$ and the maximum is $7.85\times10^{-7}$ which confirms the sDNN keeping the even symmetry. However, the blue line in the first quadrant and the red line in the whole domain for PINN stay far above the black line which implies that the PINN cannot maintain the even symmetry even though in the sampling domain. In particular, even if the $L_2$ relative error of PINN sampling in the whole domain reaches the minimum $1.67\times10^{-3}$ at $N_f=1100$, the even metric only reaches $2.18\times10^{-3}$.

$\bullet$ \emph{Performances for different number of neurons per layer.}
The second group of experiments is to investigate the performances of sDNN with different numbers of neurons per layer of 3 hidden layers, where $N_f=100$ collocation points and $N_u =80$ initial and boundary data are used, and the number of neurons in each layer ranges from $5$ to $75$ with a step $5$ simultaneously.
In Figure \ref{fig2}(C), the black solid line for sampling domain and the green dashed line for prediction domain by sDNN completely overlap and stay far below the blue and red solid lines by PINN, which shows that the sDNN has strong solution extrapolation ability and far better than PINN. In particular, when the number of neurons is $55$, the blue solid line for the $L_2$ relative error of sDNN is very close to the black solid line by PINN, arriving at $1.20 \times 10^{-3}$ and $1.43 \times 10^{-3}$ respectively. However, the predicted accuracy in the third quadrant by PINN is $2.56 \times 10^{-1}$ which is much bigger than the one by sDNN, and the even metric of learned solution by sDNN arrives at $5.64 \times 10^{-7}$ which is far smaller than the ones by PINN in the whole domain and the sampling domain.

\begin{figure}[htp]
    \begin{minipage}{0.24\linewidth}
		\vspace{3pt}
		\centerline{\includegraphics[width=\textwidth]{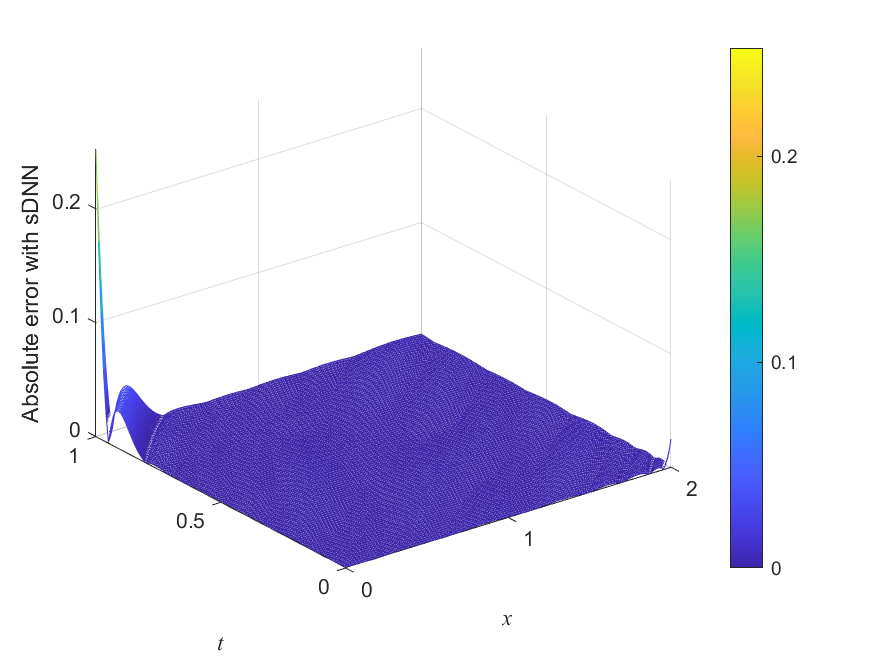}}
        \centerline{A}
	\end{minipage}
    \begin{minipage}{0.24\linewidth}
		\vspace{3pt}
		\centerline{\includegraphics[width=\textwidth]{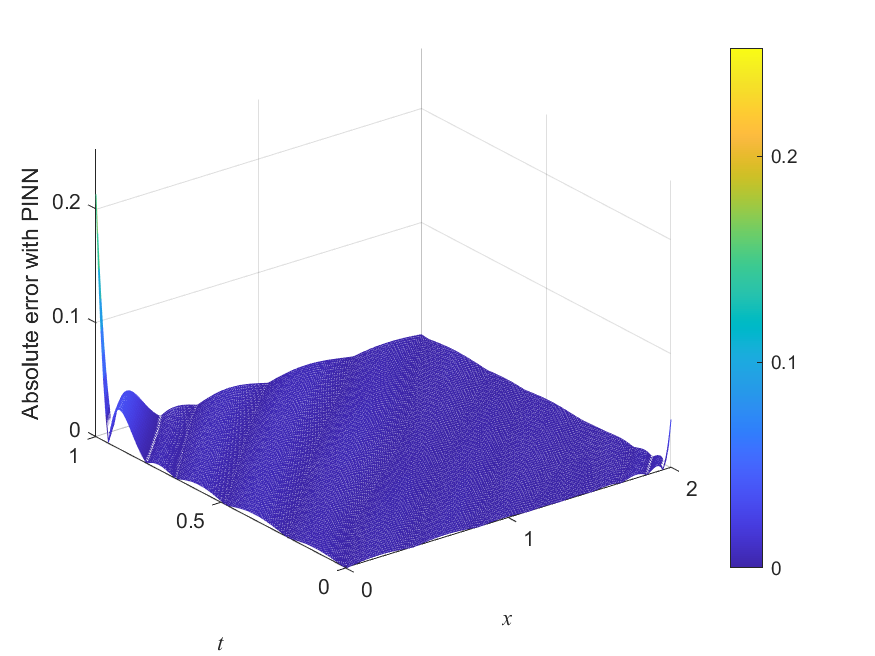}}
        \centerline{B}
	\end{minipage}
    \begin{minipage}{0.24\linewidth}
		\vspace{3pt}
		\centerline{\includegraphics[width=\textwidth]{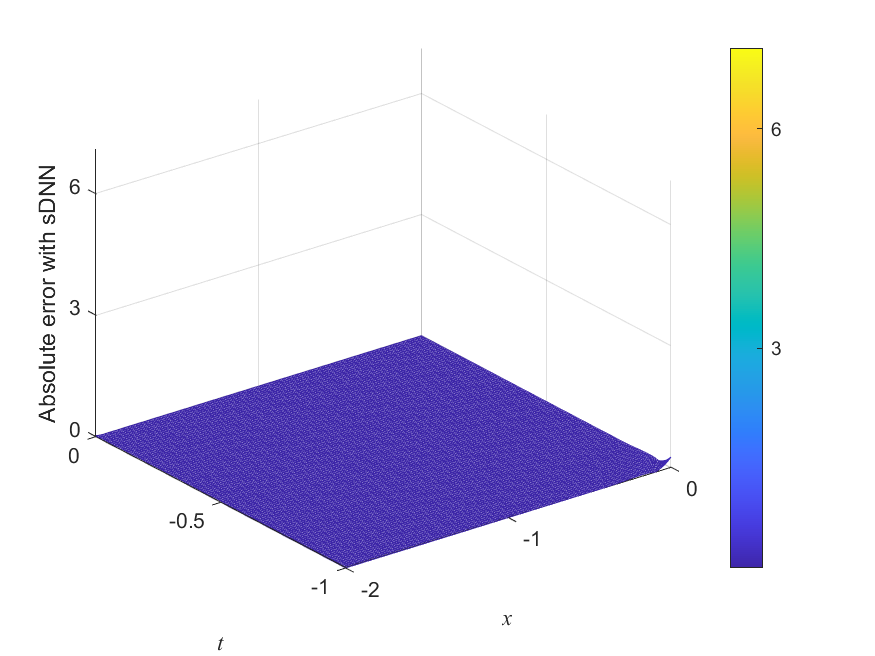}}
        \centerline{C}
	\end{minipage}
    \begin{minipage}{0.24\linewidth}
		\vspace{3pt}
		\centerline{\includegraphics[width=\textwidth]{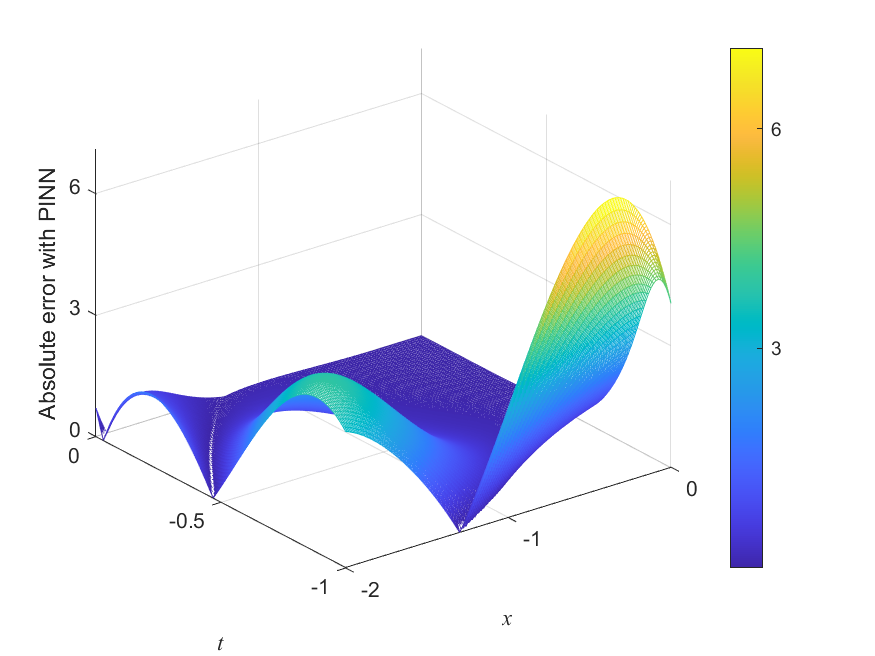}}
        \centerline{D}
	\end{minipage}
	\caption{(Color online) Advection equation: Comparisons of absolute errors of PINN and sDNN in the first quadrant and the third quadrant. (A) Abloute errors by sDNN in the first quadrant. (B) Abloute errors by PINN in the first quadrant. (C) Abloute errors by sDNN in the third quadrant. (D) Abloute errors by PINN in the third quadrant.}
\label{fig3}
\end{figure}

In particular, in Figure \ref{fig2}(C), PINN has the best predicted accuracy in the sampling domain when the number of neurons is $55$. Thus we choose this case to show the difference of the learned solutions by sDNN and PINN visually. Figure \ref{fig3}(A,C) represents the absolute errors of sDNN in and beyond the sampling domain and the absolute error of PINN in the sampling domain, all of which are very flat. Figure \ref{fig3}(D) displays the absolute error of PINN beyond the sampling domain, which fluctuates greatly and reaches a maximum $7.09$ though PINN has a good $L_2$ relative error in the sampling domain shown in Figure \ref{fig3}(B). This means that PINN has poor solution extrapolation ability.

\begin{figure}[htp]
	\begin{minipage}{0.24\linewidth}
		\vspace{3pt}
		\centerline{\includegraphics[width=\textwidth]{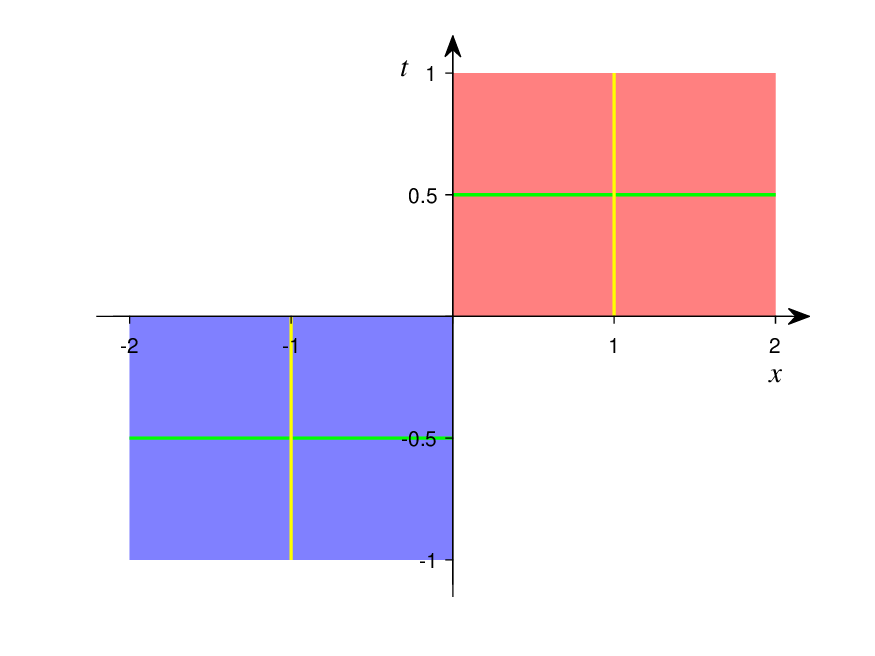}}
        \centerline{A}
	\end{minipage}
	\begin{minipage}{0.24\linewidth}
		\vspace{3pt}
		\centerline{\includegraphics[width=\textwidth]{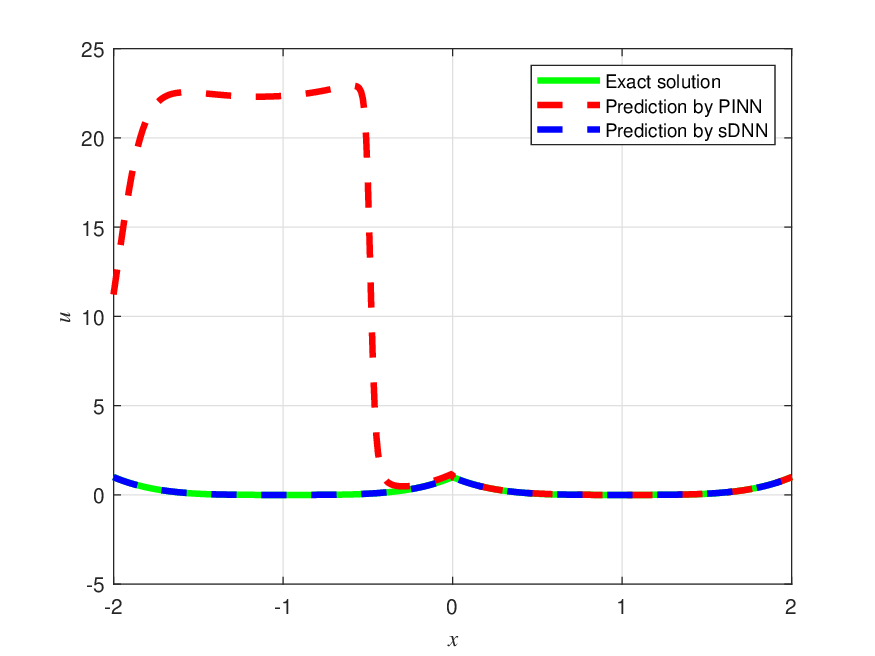}}
        \centerline{B}
	\end{minipage}
	\begin{minipage}{0.24\linewidth}
		\vspace{3pt}
		\centerline{\includegraphics[width=\textwidth]{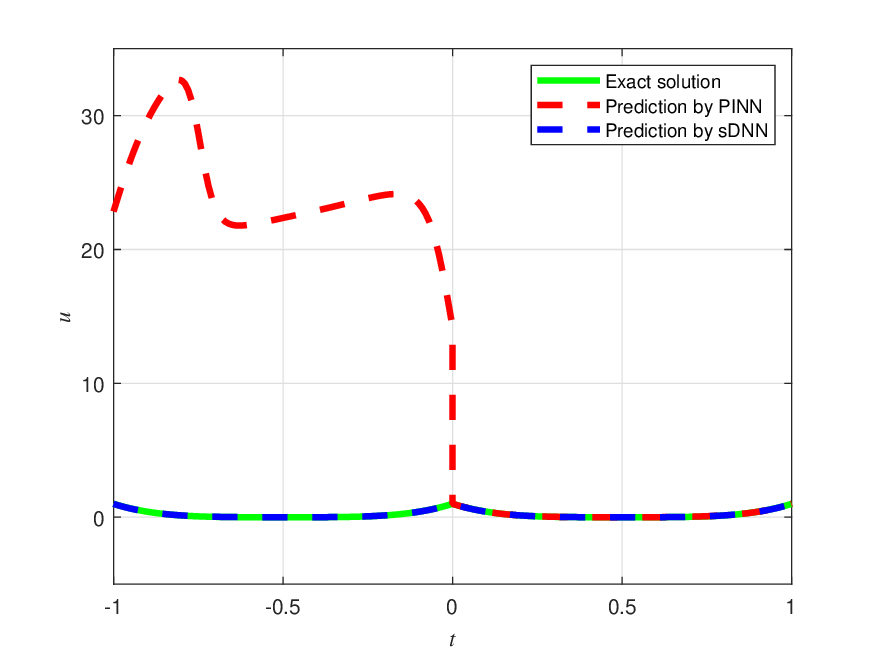}}
        \centerline{C}
	\end{minipage}
    \begin{minipage}{0.24\linewidth}
		\vspace{3pt}
		\centerline{\includegraphics[width=\textwidth]{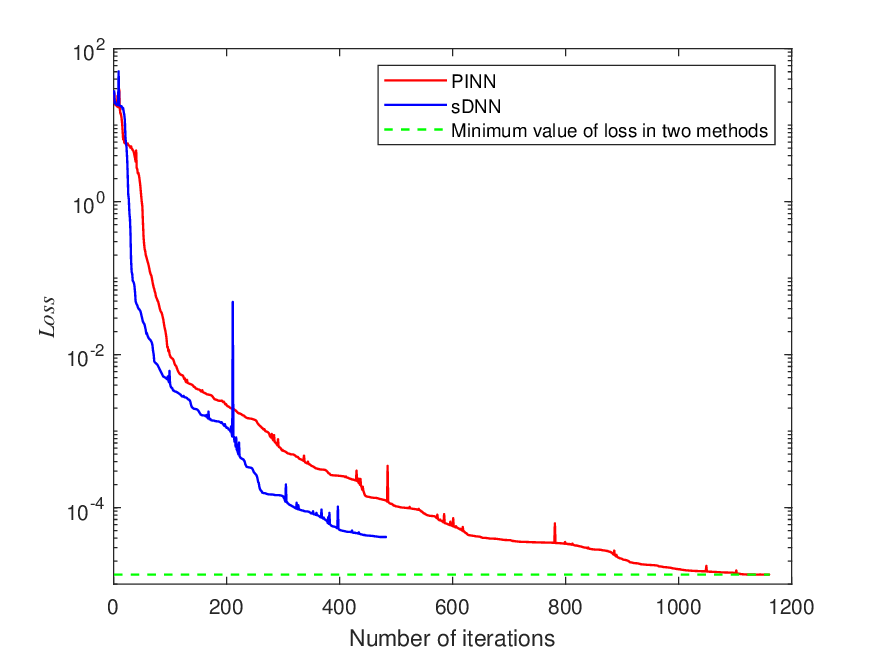}}
        \centerline{D}
	\end{minipage}
	\caption{(Color online) Advection equation: Cross sections of predicted solutions by PINN and sDNN and exact solutions. (A) Schematic diagrams of sampling and prediction domains where the green lines represent the two values of $t$ in cross section B and the yellow lines represent  the two values of $x$ in cross section C. (B) Cross sections for $t=\pm0.5$: $t=0.5$ for $x\geq0$; $t=-0.5$ for $x < 0$. (C) Cross sections for $x=\pm1.00$: $x=1.00$ for $t\geq 0$; $x=-1.00$ for $t < 0$. (D) Loss histories for PINN and  sDNN against the number of iterations. }
\label{fig4}
\end{figure}

Furthermore, we display the predicted solutions by sDNN and PINN and the exact solution at $t=\pm0.50$ and $x=\pm1.00$ when $N_f = 1100$ in Figure \ref{fig4}.  Figure \ref{fig4}(A-C) shows that in the sampling first quadrant $x\in [0,2]$ and $t\in [0,1]$ the learned solutions by both sDNN and PINN have no visually differences, however, in the prediction third quadrant $x\in [-2,0]$ and $t\in [-1,0]$ the predicted solution by PINN far deviates the exact solution while the one by sDNN still fits the exact solution very well. Moreover, the loss histories in Figure \ref{fig4}(D) shows that the sDNN takes $483$ iterations to get $4.13 \times 10^{-5}$ while the PINN reaches $1.33 \times 10^{-5}$ after $1161$ iterations, but the $L_2$ relative error by sDNN is smaller than the one by PINN, which implies that the sDNN has strong learning ability than PINN.

\subsection{sine-Gordon equation}
The second example is the sine-Gordon equation
\begin{eqnarray}\label{sine}
u_{xt}=\sin u,~~~x\times t\in[-3,3]\times[-3,3],
\end{eqnarray}
which arises in application to one-dimensional crystal dislocation theory, wave propagation in ferromagnetic materials, etc. Eq.(\ref{sine}) has an exact solution $u_{sg}=4 \arctan\Big(\frac{2 \sqrt{2} e^{\frac{t+x}{\sqrt{2}}} \cos \left(\frac{x-t}{\sqrt{2}}\right)}{2 e^{\sqrt{2} (t+x)}+1}\Big)$ \cite{ak-2020}. Eq.(\ref{sine}) is invariant under a circulant symmetry $g_c:(x,t,u(x,t))\mapsto(t,x,u(t,x))$, thus we learn the solution $u_{sg}$ by only training in the triangle below the line $x-t=0$ in the rectangle domain $[-3,3]\times[-3,3]$ which generates the whole domain by $G_c=\{g_0,g_c\}$ with an identity element $g_0$. Since $G_c$ has the matrix representations,
\begin{eqnarray}
&&\no g_0=\left(\begin{array}{cc}
               1 & 0 \\
               0 & 1\\
             \end{array}
           \right),~
           g_c=\left(\begin{array}{cc}
              0 & 1 \\
               1 & 0\\
             \end{array}
           \right),
\end{eqnarray}
then by Corollary \ref{th-cyclic}, we obtain the following results.
\begin{prop}[Circulant symmetry]
Let $G_c$ be a circulant group. Then the weight matrixes and bias vectors of sDNN with $L-1$ hidden layers which keep the group $G_c$ invariant takes the form
\begin{eqnarray}
&&\no \hspace{-0.7cm}\text{Weight matrixes:}~\left(
             \begin{array}{cc}
               \w_1^{(1)} & \w_1^{(2)} \\
               \w_1^{(2)} & \w_1^{(1)} \\
             \end{array}
           \right),~\dots,~\left(
             \begin{array}{cc}
               \w_{L-1}^{(1)} & \w_{L-1}^{(2)} \\
               \w_{L-1}^{(2)} & \w_{L-1}^{(1)} \\
             \end{array}
           \right),~\left(
  \begin{array}{c}
    \w_L \\
   \w_L\\
  \end{array}
\right);\\
&& \no\text{Bias vectors:}~ \left(
  \begin{array}{c}
    \b_1 \\
    \b_1
  \end{array}
\right),~\left(
  \begin{array}{c}
    \b_2 \\
    \b_2
  \end{array}
\right),~\dots,~\left(
  \begin{array}{c}
    \b_{L}
  \end{array}
\right),
\end{eqnarray}
where  $\w_l^{(1)}, \w_l^{(2)}$ and $\w_L$ are the initialized weight matrixes and $\b_l$ are the bias vectors of the PINN, $l=2,\dots,L-1$.
\label{prop-cir}
\end{prop}

Next, we deploy the sDNN equipped with the weight matrixes and bias vectors in Proposition \ref{prop-cir} to learn the solution $u_{sg}$ where the sampling domain and the training points for the PINN and sDNN are completely the same. In order to obtain the training data set, we discretize the spatial domain $x\in [-3,3]$ and the temporal domain $t\in [-3,3]$ into $N_x=N_t=301$ equidistant points respectively.

$\bullet$ \emph{Performances for different number of collocation points.}
In the first trial group, both the sDNN and PINN have 3 hidden layers with $5$ neurons per layer except that the second layer in PINN contains $10$ neurons in order to keep the same number of training parameters.
Then we randomly choose $N_u = 100$ initial and boundary points and use Latin hypercube sampling to get the collocation points falling in the sampling red domain in Figure \ref{fig9}(A) whose numbers vary from $100$ to $1500$ with a step $100$. 

\begin{figure}[htp]
	\begin{minipage}{\linewidth}
		\centerline{\includegraphics[width=\textwidth]{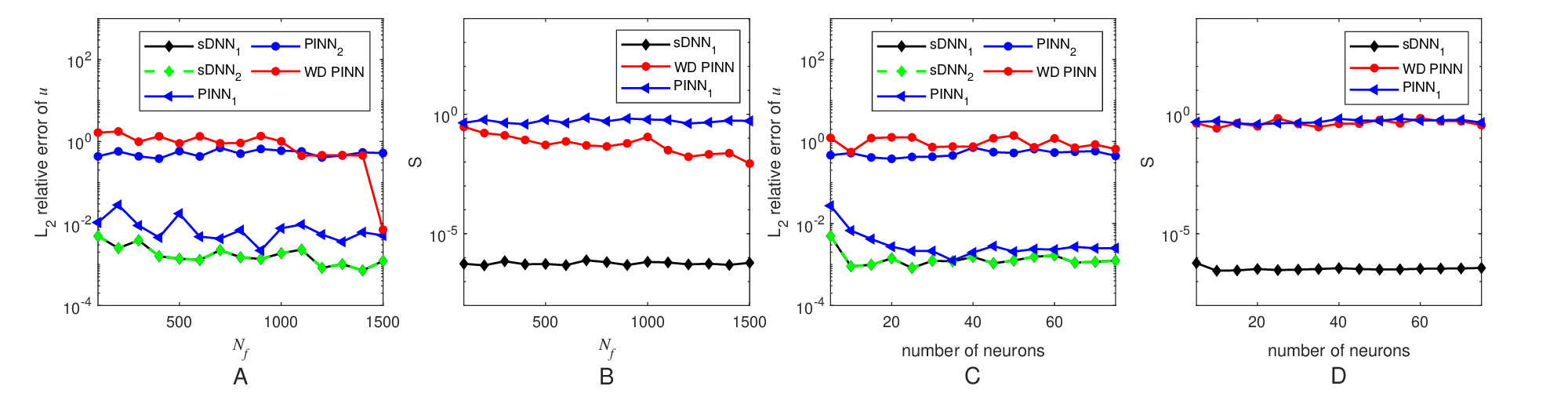}}
	\end{minipage}
	\caption{(Color online) sine-Gordon equation: Comparisons of $L_2$ relative errors and the even metric of $u$ in the sampling and prediction domains by the two methods. Keeping the number of neurons invariant and varying the number of collocation points: (A) the $L_2$ relative errors; (B) the symmetry metrics. Keeping the number of collocation points invariant and varying the number of neurons: (C) the $L_2$ relative errors; (D) the symmetry metrics. Each line represents the mean of five independent experiments. Note that the sDNN and PINN with subscript $1$ and $2$ of denote the sampling domain and prediction domain respectively, and the symbol `WD PINN' means the training in the whole domain.}
\label{fig7}
\end{figure}

Figure \ref{fig7}(A) shows that the sDNN gives very close accuracies of learned solutions in the sampling and prediction domains where their maximum difference is only $6.16 \times 10^{-9}$, but the PINN has no such an ability to repeat the situation. For example, when the collocation points $N_f=900$, the $L_2$ relative error by PINN arrives at $2.20 \times 10^{-3}$ in the sampling domain but only gets $6.60 \times 10^{-1}$ in the prediction domain. Meanwhile, we find that the $L_2$ relative error by sDNN at $N_f=900$ is $1.36 \times 10^{-3}$ which is very close to the one by PINN, but the Circulant metric, defined by $S=\left\|u(x,t)_{pred}-u(t,x)_{pred}\right\|_{2}/\left\|u(x,t)_{pred}\right\|_{2}$, is $4.91 \times 10^{-7}$ which is greatly smaller than $6.60 \times 10^{-1}$ by PINN.

$\bullet$ \emph{Performances for different number of neurons per layer.}
The second group of experiments aims to investigate the performances of sDNN against PINN with varying number of neurons per layer. Both sDNN and PINN are equipped with $3$ hidden layers, and the number of neurons ranges from $5$ to $75$ with a step $5$. Both two methods use $N_u=100$ initial and boundary points and $48$ collocation points selected from the total $N_f=100$ collocation points in the whole domain $x\times t\in[-3,3]\times[-3,3]$ via the Latin hypercube sampling, because the other 52 collocation points lie beyond the sampling triangle domain. Figure \ref{fig7}(C) shows that in the sampling domain the sDNN has small advantage over the PINN, even though they have very close $L_2$ relative errors at $35$ neurons, but in the prediction domain the sDNN takes overwhelming superiority than PINN, where the $L_2$ relative errors by sDNN are almost the same with the ones of sampling domain while the PINN only presents $10^{-1}$ order of magnitude described by either the red line in the whole domain or the blue dotted line in the domain above the line $x=t$ of the rectangle $x\times t \in [-3,3]\times[-3,3]$. Furthermore, Figure \ref{fig7}(D) displays that the Circulant metrics of sDNN always remain at $10^{-7}$ order of magnitude much lower than those obtained by PINN, in particular, with 35 neurons the sDNN has six order of magnitude improvement than PINN though they have very close $L_2$ relative errors displayed in Figure \ref{fig7}(C).

\begin{figure}[ht]
    \begin{minipage}{0.24\linewidth}
		\vspace{3pt}
		\centerline{\includegraphics[width=\textwidth]{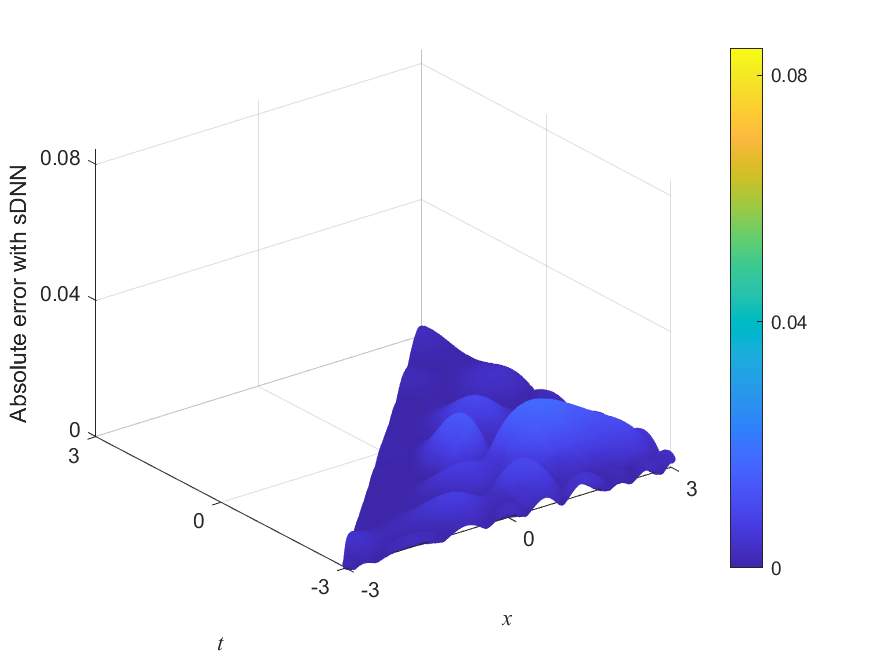}}
        \centerline{A}
	\end{minipage}
    \begin{minipage}{0.24\linewidth}
		\vspace{3pt}
		\centerline{\includegraphics[width=\textwidth]{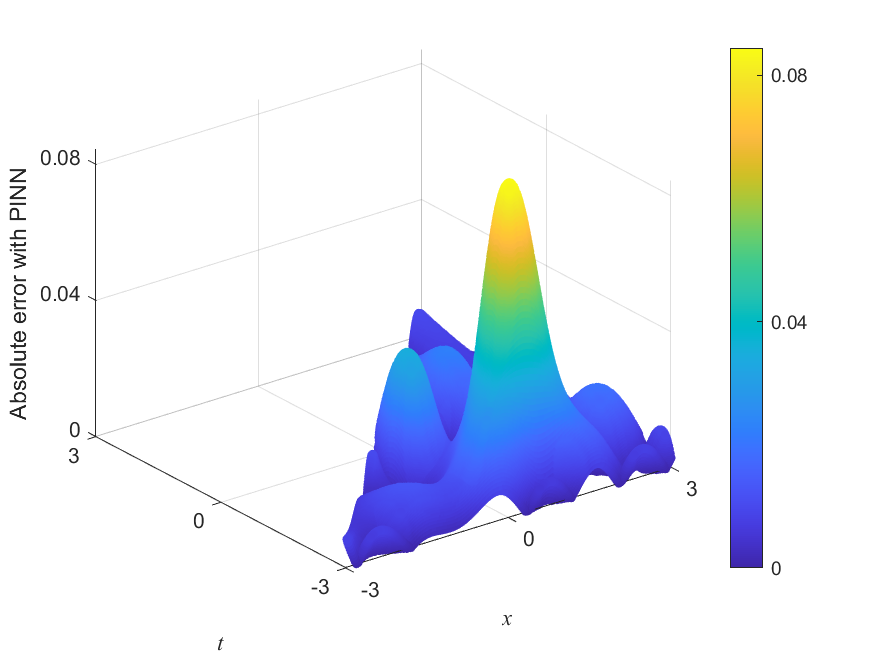}}
        \centerline{B}
	\end{minipage}
    \begin{minipage}{0.24\linewidth}
		\vspace{3pt}
		\centerline{\includegraphics[width=\textwidth]{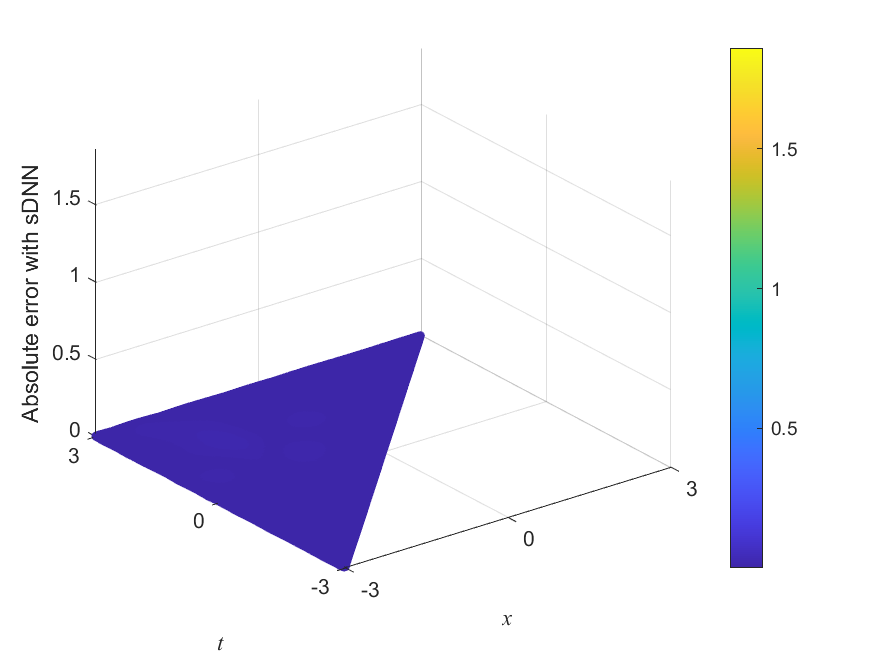}}
        \centerline{C}
	\end{minipage}
    \begin{minipage}{0.24\linewidth}
		\vspace{3pt}
		\centerline{\includegraphics[width=\textwidth]{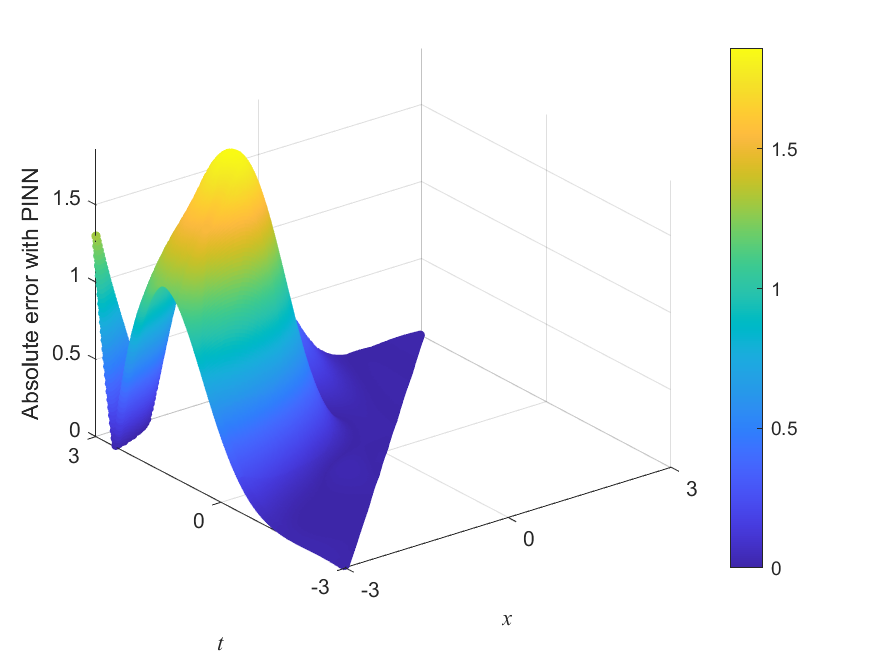}}
        \centerline{D}
	\end{minipage}
	\caption{(Color online) sine-Gordon equation: Comparisons of absolute errors of PINN and sDNN in the sampling and prediction domains. (A) Absolute errors by sDNN in the sampling domain. (B) Absolute errors by PINN in the sampling domain. (C) Absolute errors by sDNN in the prediction domain. (D) Absolute errors by PINN in the prediction domain.}
\label{fig8}
\end{figure}

\begin{figure}[htp]
	\begin{minipage}{0.24\linewidth}
		\vspace{3pt}
		\centerline{\includegraphics[width=\textwidth]{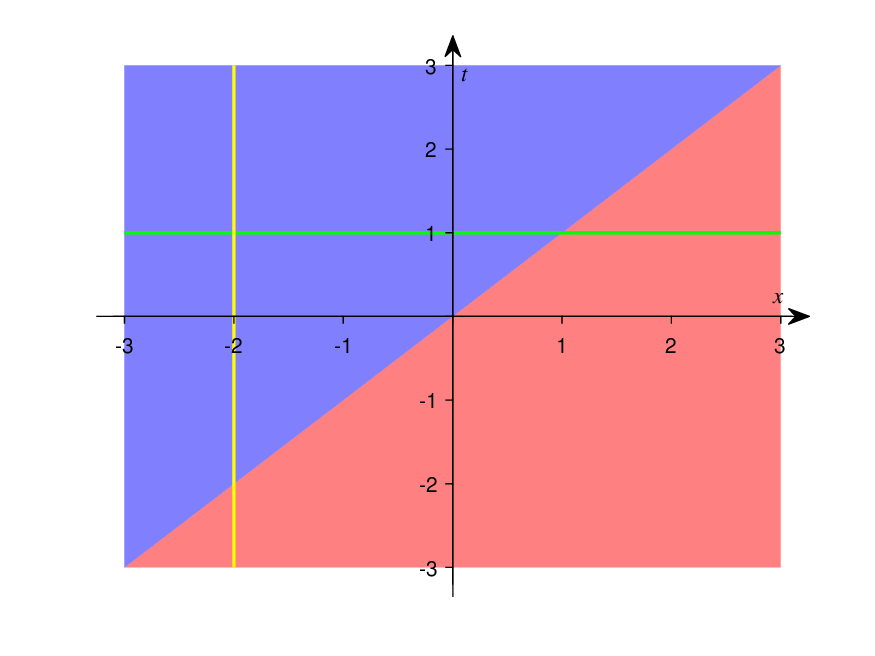}}
        \centerline{A}
	\end{minipage}
	\begin{minipage}{0.24\linewidth}
		\vspace{3pt}
		\centerline{\includegraphics[width=\textwidth]{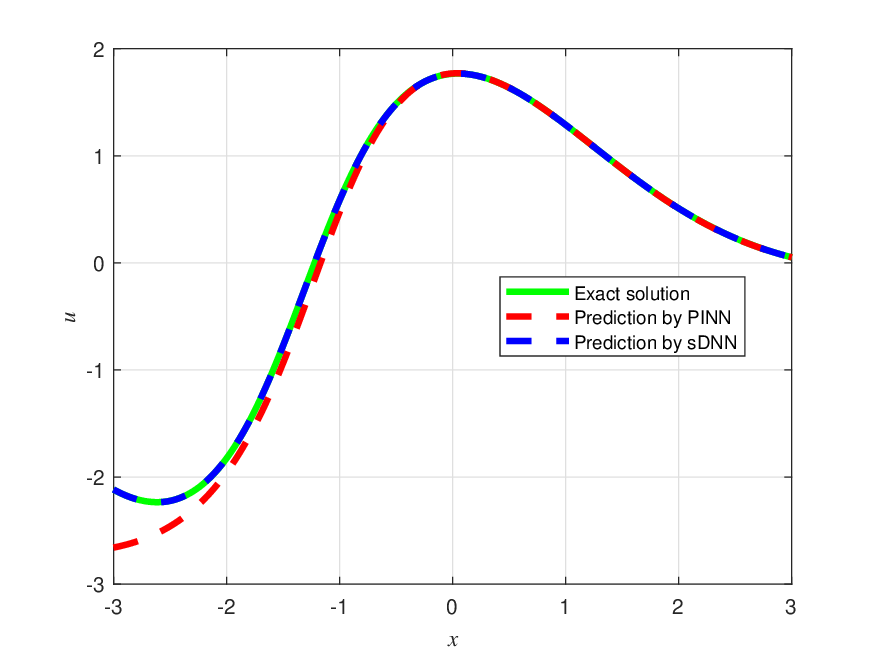}}
        \centerline{B}
	\end{minipage}
	\begin{minipage}{0.24\linewidth}
		\vspace{3pt}
		\centerline{\includegraphics[width=\textwidth]{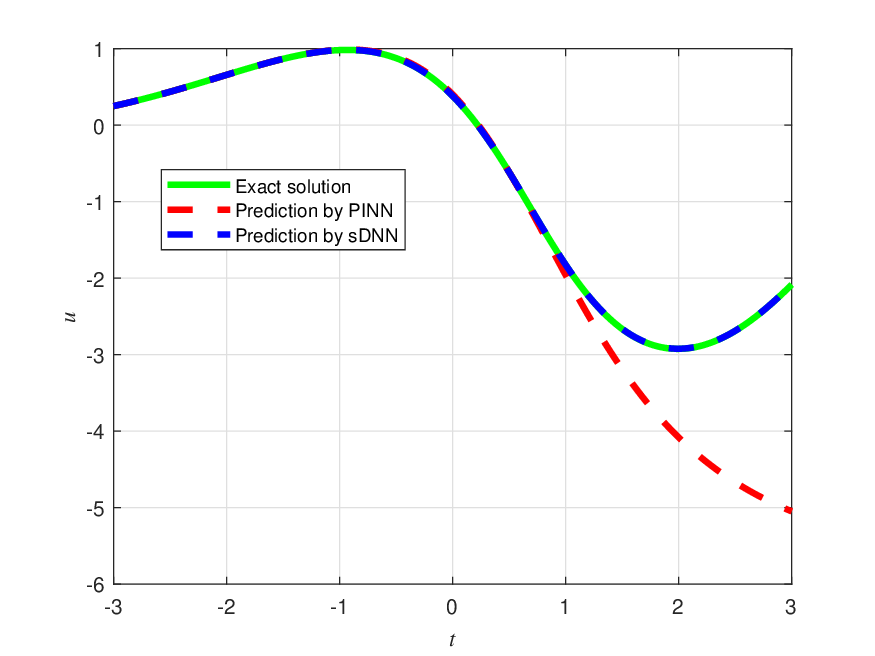}}
        \centerline{C}
	\end{minipage}
    \begin{minipage}{0.24\linewidth}
		\vspace{3pt}
		\centerline{\includegraphics[width=\textwidth]{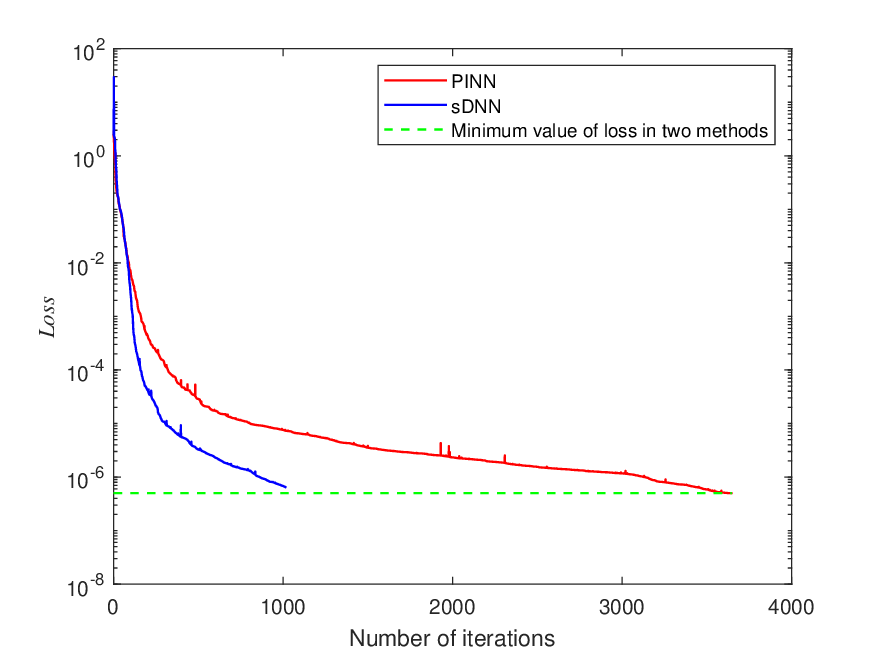}}
        \centerline{D}
	\end{minipage}
	\caption{(Color online) sine-Gordon equation: Cross sections of predicted solutions by PINN and sDNN and exact solutions. (A) Schematic diagrams of sampling and prediction domains where the green lines represent the two values of $t$ in cross section B and the yellow lines represent  the two values of $x$ in cross section C.  (B) Cross sections for $t=1.0$. (C) Cross sections for $x=-2.0$. (D) Loss histories for PINN and sDNN over number of iterations. }
\label{fig9}
\end{figure}

Furthermore, we choose one case in Figure \ref{fig7}(C) to show the absolute errors of learned solutions by sDNN and PINN in and beyond sampling domain. Figure \ref{fig8}(B,D) shows that the absolute errors by PINN fluctuate largely both in and beyond the sampling domain, whereas the ones of sDNN in Figure \ref{fig8}(A,C) are very flat. Specifically, the absolute errors by PINN in and beyond the sampling domain reach $8.43 \times 10^{-2}$ and 1.86 respectively, while the absolute errors of sDNN do not exceed $1.78 \times 10^{-2}$.

Figure \ref{fig9}(B-C) displays the cross sectional graphes at $t=1.0$ and $x=-2.0$. In Figure \ref{fig9}(B), for $x>1$, both PINN and sDNN fit well with the exact solution, but for $x<1$, as $x$ moves away from $x=1$, the predicted solution by PINN moves slowly away from the exact solution, where for $x\in[-0.5,1]$ there has no visual difference between the three lines, but for $x\in[-3,-0.5]$ the red dotted line for PINN strongly deviates the green line of exact solution, while the predicted solution by sDNN still exhibits highly overlaps with the exact solution. Figure \ref{fig9}(C) displays that in the sampling domain $t\in[-3,-2]$ both PINN and sDNN fit the exact solution very well, even in the prediction domain $t\in[-2,0.5]$ there have no visible differences among the three lines. However, for $t\in[0.5,3]$ the sDNN exhibits strong solution extrapolation ability where the blue dotted line overlaps with the green line rigorously, but the red dotted line for PINN obviously moves along the direction against the green line.
The loss histories in Figure \ref{fig9}(D) shows that PINN achieves $4.99\times 10^{-7}$ loss value with $3647$ iterations, while the sDNN reaches $6.37 \times 10^{-7}$ after only $1019$ iterations where with the same number of iterations the PINN gets $7.30 \times 10^{-6}$, which means that the sDNN has a stronger capacity for learning the solution $u_{sg}$ of Eq.(\ref{sine}) than PINN. 
\subsection{Poisson equation}
The third example is the two-dimensional Poisson equation
\begin{eqnarray}\label{poisson}
\triangle u(x,y)=f(x,y),~~~x\times y\in[-2,2]\times[-2,2],
\end{eqnarray}
where $\triangle$ is the Laplace operator. Eq.(\ref{poisson}) exerts important roles in the fields of electrostatics, mechanical engineering and theoretical physics \cite{aps-2000}.

\begin{figure}[htp]
	\begin{minipage}{0.33\linewidth}
		\vspace{3pt}
		\centerline{\includegraphics[width=\textwidth]{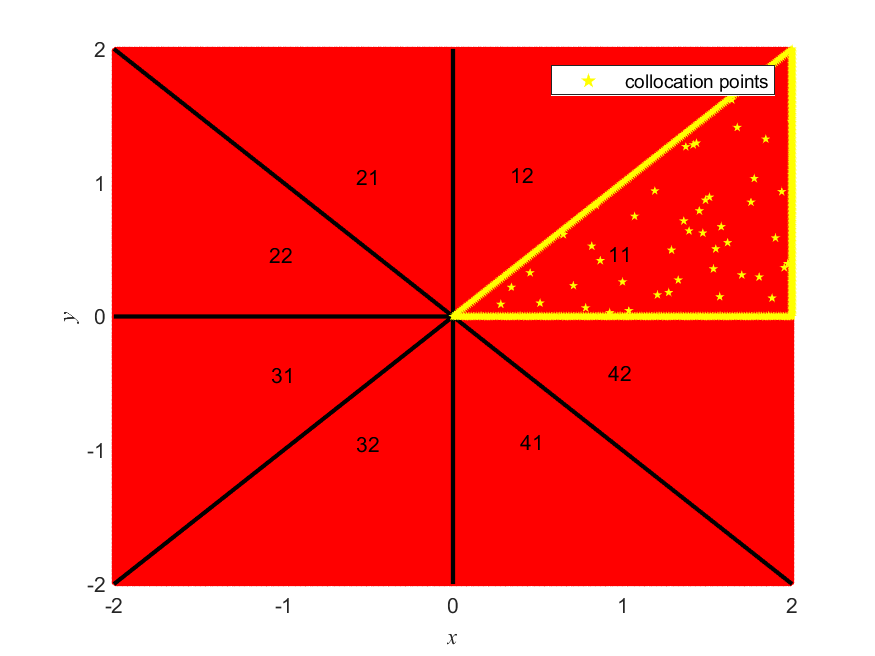}}
        \centerline{A. Eighth-order group}
	\end{minipage}
	\begin{minipage}{0.33\linewidth}
		\vspace{3pt}
		\centerline{\includegraphics[width=\textwidth]{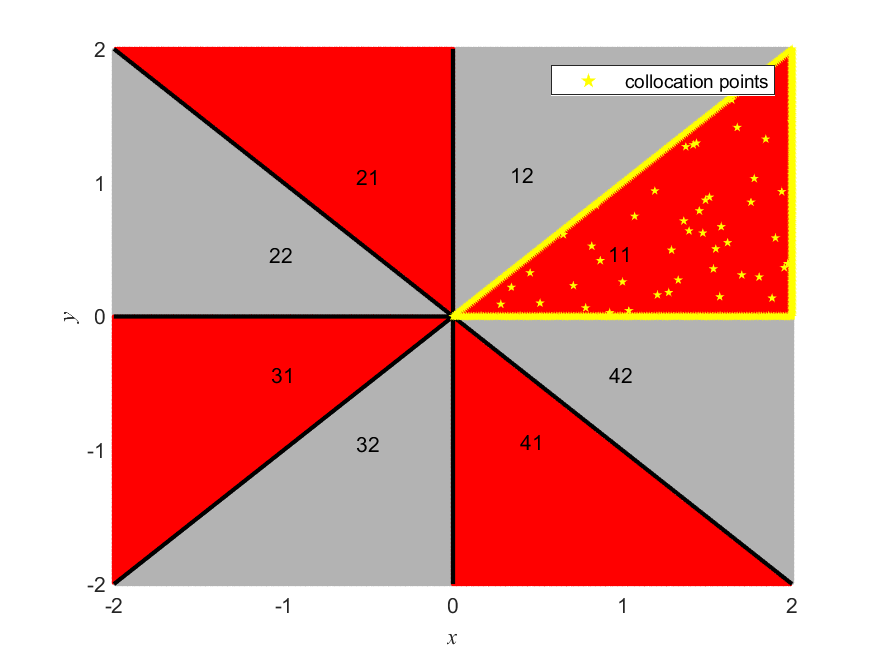}}
        \centerline{B. Fourth-order group}
	\end{minipage}
    \begin{minipage}{0.33\linewidth}
		\vspace{3pt}
		\centerline{\includegraphics[width=\textwidth]{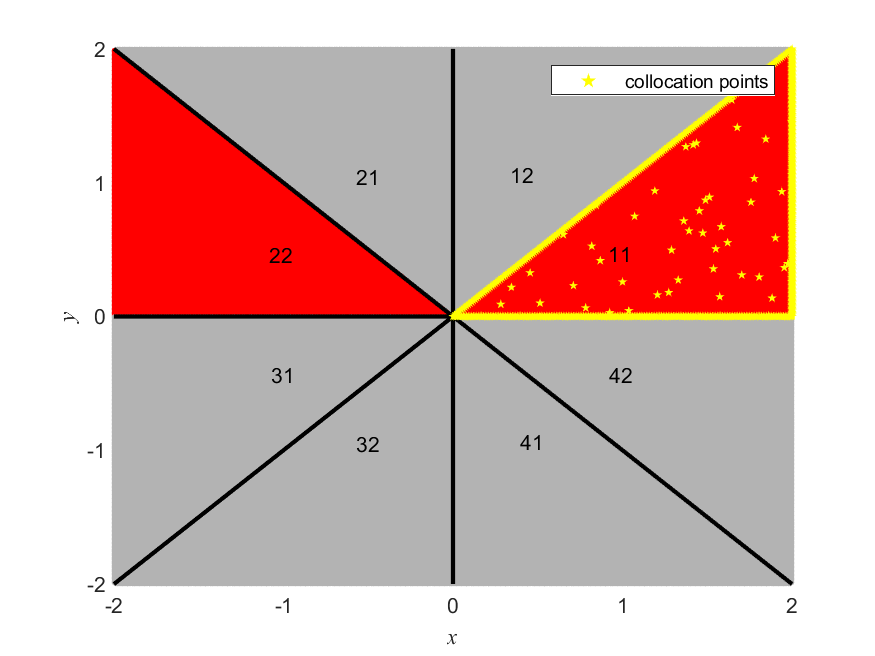}}
        \centerline{C. Second-order group}
	\end{minipage}
	\caption{(Color online) Poisson equation: The sampling and prediction domains for the sDNN with the three groups. The number pair $ij$ denotes the domain of the $i$-th($i=1,2,3,4$) quadrant and the $j$-th($j=1,2$) triangle with anticlockwise direction. In each graph the 11 red domain is the sampling domain where the yellow diamond denote the distribution of collocation points, and the other red domains can be predicted via the finite group while the grey ones are not done by the group.}
\label{fig28}
\end{figure}

Next, we leverage the eight-order Dihedral group $G_d$ as well as its two subgroups, rotation group $G_r$ and reflection group $G_p$, to learn the solution $u(x,y)=\cos(\pi x)\cos(\pi y)$ of Eq.(\ref{poisson}). Specifically, just as shown in Figure \ref{fig28}(A-C), we choose the bottom triangle in the first quadrant, i.e. the 11 domain, as the sampling domain to predict the solutions in the eight domains. Note that the function $f(x,y)$ in Eq.(\ref{poisson}) together with the initial and boundary conditions are enforced by the solution $u(x,y)$. 

Observed that the elements in the Dihedral group $G_d=\left\{\,g_i|i=0,1,\dots,7\,\right\}$ have the matrix representations in the form
\begin{eqnarray}
&&\no\hspace{-1cm}\left\{\,g_i|i=0,1,\dots,7\,\right\}=\left\{\left(
    \begin{array}{cc}
      1 & 0 \\
      0 & 1 \\
    \end{array}
  \right),\left(
    \begin{array}{cc}
       1 & 0 \\
     0 & -1 \\
    \end{array}
  \right),\left(
    \begin{array}{cc}
      -1 & 0 \\
      0 & 1 \\
     \end{array}
  \right),\left(
    \begin{array}{cc}
      -1 & 0 \\
      0 & -1 \\
    \end{array}
  \right)\right.,\\
 && \hspace{2.9cm}\left.\left(
    \begin{array}{cc}
      0 & 1 \\
      1 & 0 \\
    \end{array}
  \right),\left(
    \begin{array}{cc}
      0 & 1 \\
      -1 & 0 \\
    \end{array}
  \right),\left(
    \begin{array}{cc}
      0 & -1 \\
      1 & 0 \\
    \end{array}
  \right),\left(
    \begin{array}{cc}
      0 & -1 \\
      -1 & 0 \\
    \end{array}
  \right)
\right\},
\end{eqnarray}
then the rotation symmetry group $G_r=\{g_0,g_3,g_5,g_6\}$ and the reflection symmetry group $G_p=\{g_0,g_1\}$. Thus, by Corollary \ref{th-cyclic}, we have the following results.

\begin{prop} Let $Z_d=\{z_0,z_1,\dots,z_7\}$ be a permutation group corresponding to the Cayley table of group $G_d$, $\w_j^{(i)},\,\w_L$ and $\b_j(i=1,\dots,8,j=1,2,\dots,L-1),\,\b_L$ are the initialized weight matrixes and bias vectors of the PINN respectively. Then the weight matrixes and bias vectors of the sDNN keeping the three group $G_d, G_r$ and $G_p$ respectively are listed as follows.

$\bullet$ \textbf{Dihedral group}. The weight matrixes and bias vectors of the sDNN with $L-1$ hidden layers which keeps the group $G_d$ invariant take the form
\begin{eqnarray}
&&\no \hspace{-0.7cm}\text{Weight matrixes:}
\left(
             \begin{array}{cc}
W_1 g_0\\              
W_1 g_1\\
W_1 g_2\\
W_1 g_3\\
W_1 g_4\\ 
W_1 g_5\\
W_1 g_6\\
W_1 g_7\\
             \end{array}
           \right),
           ~\left(
             \begin{array}{c}
W_2 z_0\\
W_2 z_1\\
W_2 z_2\\
W_2 z_3\\
W_2 z_4\\
W_2 z_5\\
W_2 z_6\\
W_2 z_7\\
             \end{array}
           \right), ~\dots,
                      ~\left(
             \begin{array}{c}
W_{L-1}z_0\\
W_{L-1}z_1\\
W_{L-1}z_2\\
W_{L-1}z_3\\
W_{L-1}z_4\\
W_{L-1}z_5\\
W_{L-1}z_6\\
W_{L-1}z_7\\
             \end{array}
           \right),
~\left(
\begin{array}{c}
    \w_L \\
   \w_L\\
   \w_L \\
   \w_L\\
   \w_L \\
   \w_L\\
   \w_L \\
   \w_L\\
  \end{array}
\right);\\
&& \no\text{Bias vectors:}~ \left(
  \begin{array}{c}
    \b_1 \\
    \b_1\\
    \b_1 \\
    \b_1\\
    \b_1 \\
    \b_1\\
    \b_1 \\
    \b_1\\
  \end{array}
\right),~\left(
  \begin{array}{c}
    \b_2 \\
    \b_2\\
    \b_2 \\
    \b_2\\
    \b_2 \\
    \b_2\\
    \b_2 \\
    \b_2\\
  \end{array}
\right),~\dots,~\left(
  \begin{array}{c}
    \b_{L} \\
  \end{array}
\right),
\end{eqnarray}
where $W_1=(\w_1^{(1)}, \w_1^{(2)})$ and $W_j=(\w_j^{(1)}, \w_j^{(2)}, \w_j^{(3)}, \w_j^{(4)}, \w_j^{(5)}, \w_j^{(6)}, \w_j^{(7)}, \w_j^{(8)})$.

$\bullet$ \textbf{Rotation group}. The weight matrixes and bias vectors of the sDNN with $L-1$ hidden layers which keeps the group $G_r$ invariant take the form
\begin{eqnarray}
&&\no \hspace{-0.7cm}\text{Weight matrixes:}~\left(
             \begin{array}{cc}
W_1 g_0\\
W_1 g_3\\
W_1 g_5\\
W_1 g_6\\
             \end{array}
           \right),~\left(
             \begin{array}{cccc}
W_2 z_0\\
W_2 z_3\\
W_2 z_5\\
W_2 z_6\\
             \end{array}
           \right),~\dots,~\left(
             \begin{array}{cccc}
W_{L-1}z_0\\
W_{L-1}z_3\\
W_{L-1}z_5\\
W_{L-1}z_6\\
             \end{array}
           \right),~\left(
  \begin{array}{c}
    \w_L \\
   \w_L\\
   \w_L \\
   \w_L\\
  \end{array}
\right);\\
&& \no\text{Bias vectors:}~ \left(
  \begin{array}{c}
    \b_1 \\
    \b_1\\
    \b_1 \\
    \b_1\\
  \end{array}
\right),~\left(
  \begin{array}{c}
    \b_2 \\
    \b_2\\
    \b_2 \\
    \b_2\\
  \end{array}
\right),~\dots,~\left(
  \begin{array}{c}
    \b_{L} \\
  \end{array}
\right),
\end{eqnarray}
where $W_1=(\w_1^{(1)}, \w_1^{(2)})$ and $W_j=(\w_j^{(1)}, \w_j^{(2)}, \w_j^{(3)}, \w_j^{(4)})$.

$\bullet$ \textbf{Reflection group}. The weight matrixes and bias vectors of the sDNN with $L-1$ hidden layers which keep the group $G_p$ invariant take the form
\begin{eqnarray}
&&\no \hspace{-0.7cm}\text{Weight matrixes:}~\left(
             \begin{array}{cc}
W_1 g_0\\
W_1 g_1
             \end{array}
           \right),~\left(
             \begin{array}{cc}
W_2 z_0\\
W_2 z_1
             \end{array}
           \right),~\dots,~\left(
             \begin{array}{cc}
W_{L-1}z_0\\
W_{L-1}z_1
             \end{array}
           \right),~\left(
  \begin{array}{c}
    \w_L \\
   \w_L\\
  \end{array}
\right);\\
&& \no\text{Bias vectors:}~ \left(
  \begin{array}{c}
    \b_1 \\
    \b_1
  \end{array}
\right),~\left(
  \begin{array}{c}
    \b_2 \\
    \b_2
  \end{array}
\right),~\dots,~\left(
  \begin{array}{c}
    \b_{L}
  \end{array}
\right),
\end{eqnarray}
where $W_j=(\w_j^{(1)}, \w_j^{(2)})$.
\label{prop-cir11}
\end{prop}

In what follow, we utilize the sDNN in Proposition \ref{prop-cir11} and the PINN to learn the solution $u(x,y)$ of Eq.(\ref{poisson}) in the whole domain, but only choose the bottom triangle in the first quadrant as the training domain.
In order to obtain the training data, we first discretize $x \in [0,2]$ and $y \in [0,2]$ into $N_{x} = N_{y} = 201$ equidistance points respectively, and then deploy Latin hypercube sampling to obtain the collocation points in the bottom triangle of the first quadrant, and predict the solutions in the eight triangle domains respectively. We notice that, in Figure \ref{fig28} (A-C), acting the finite group on the 11 domain can reach the red triangles but cannot arrive at the grey ones. Consequently, the predicted results in the red and grey domain are completely different where the red domains have almost the same accuracies as the sampling domain but the grey ones present poor results. We show the results by considering two sets of experiments to test the effects of the PINN and the sDNN armed by the eighth-order Dihedral group $G_d$ (sDNN with $G_d$), fourth-order rotation group $G_r$ (sDNN with $G_r$) and second-order reflection group $G_p$ (sDNN with $G_p$).

$\bullet$ \emph{Performances for different number of collocation points.}
In the first experimental group, we enforce both the sDNN and PINN with three hidden layers but choose the number of neurons in each hidden layer of sDNN with $G_d$, sDNN with $G_p$,  sDNN with $G_r$ and the PINN as $5-5-5, 5-10-5, 5-20-5, 5-40-5$ respectively, in order to maintain the same number of training parameters. Moreover, we choose $N_{u} = 100$ initial and boundary points randomly and use the Latin hypercube sampling to obtain the collocation points ranging from $100$ to $1500$ with step $100$, where, in particular, the collocation points are chosen from the points falling in the triangle domain at the bottom right of the whole domain. 

\begin{figure}[htp]
	\hspace{-0.7cm}\begin{minipage}{1.1\linewidth}
		\centerline{\includegraphics[width=\textwidth]{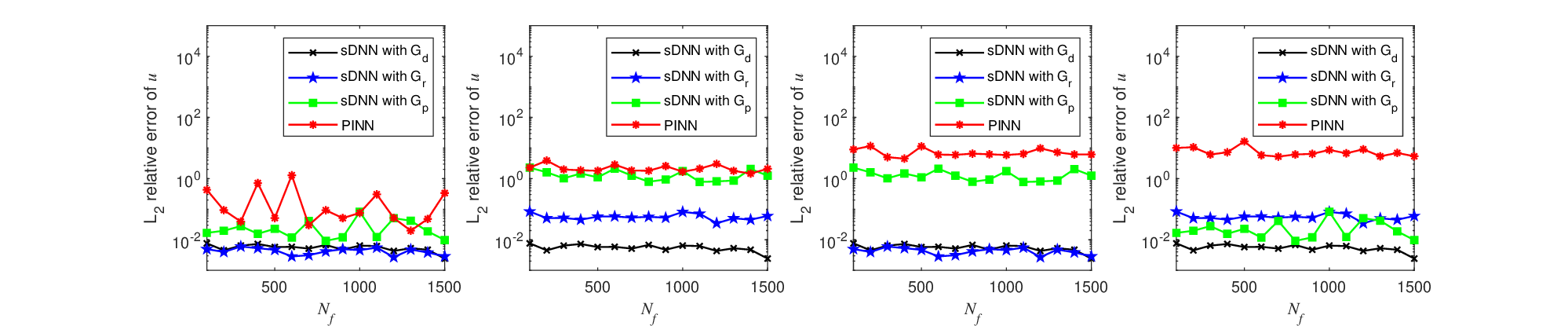}}
        \centerline{A}
	\end{minipage}

    \vspace{3pt} 

	\hspace{-0.7cm}\begin{minipage}{1.1\linewidth}
		\centerline{\includegraphics[width=\textwidth]{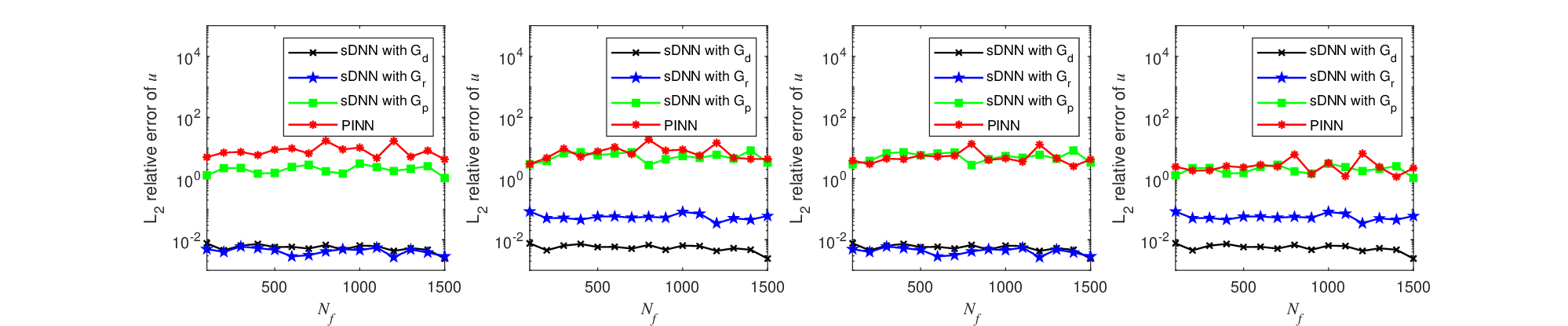}}
        \centerline{B}
	\end{minipage}
	\caption{(Color online) Poisson equation: Comparisons of $L_2$ relative errors of $u$ in different domains by the sDNN and PINN. Keeping the number of neurons invariant, the $L_2$ relative errors at different number of collocation points: (A) From left to right, domains $11$, $12$, $21$ and $22$ are represented; (B) From left to right, domains $31$, $32$, $41$ and $42$ are represented.  Each line represents the mean of five independent experiments. }
\label{fig29}
\end{figure}

Figure \ref{fig29}(A-B) show the $L_{2}$ relative errors of PINN and the sDNN armed with the eighth-order Dihedral group $G_d$, fourth-order rotation symmetry $G_r$ and second-order reflection symmetry $G_p$. The sDNN with $G_d$ gives excellent predicted accuracies for the eight triangular domains in Figure \ref{fig28}(A) where the maximum is only $7.64 \times 10^{-3}$ and the predicted accuracies of eight domains are very close where the maximum difference is only $4.92 \times 10^{-6}$, while the sDNN with  $G_r$ only shows good performances in the $11,21,31$ and $41$ red domains in Figure \ref{fig28}(B), even better than the sDNN with $G_d$, but does poor jobs in the grey domains where the predicted accuracies fluctuate at $10^{-1}$ order of magnitude, and the sDNN with $G_p$ which only effects in $11$ and $22$ red domains in Figure \ref{fig28}(C) and lost its effects in the other six grey domains. The worst case occurs for the PINN which only works in the sampling $11$ domain but fluctuates largely and shows bad performances in the other seven prediction domains. Most importantly, the sDNN with the three finite groups perform better than the PINN, where the sDNN with $G_d$ and $G_r$ particularly work superiority to the sDNN with $G_p$.
\begin{figure}[ht]
	\hspace{-0.7cm}\begin{minipage}{1.1\linewidth}
		\centerline{\includegraphics[width=\textwidth]{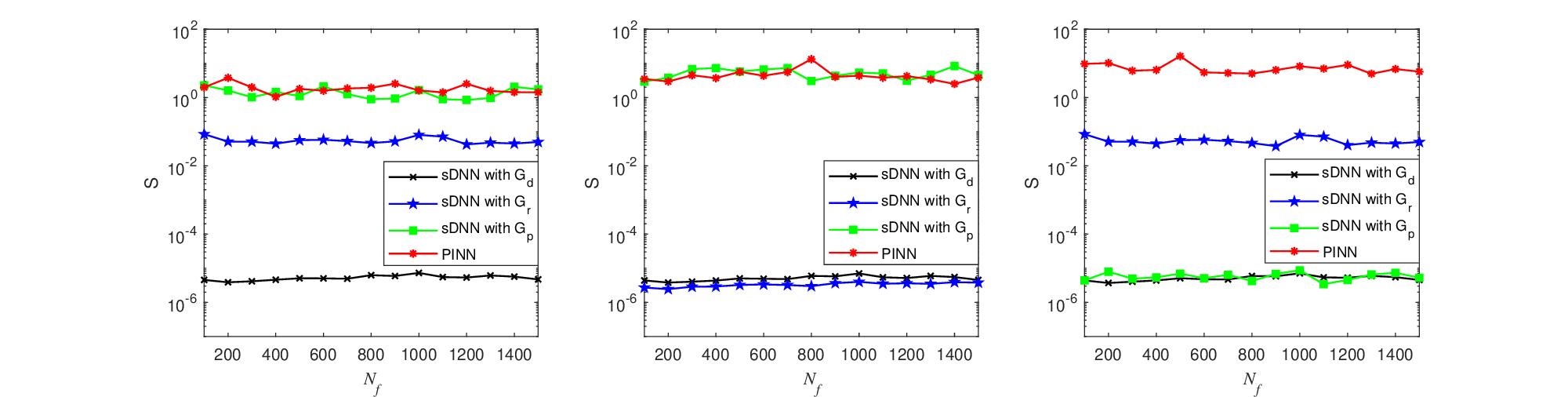}}
        \centerline{A}
	\end{minipage}

    \vspace{5pt} 

	\hspace{-0.7cm}\begin{minipage}{1.1\linewidth}
		\centerline{\includegraphics[width=\textwidth]{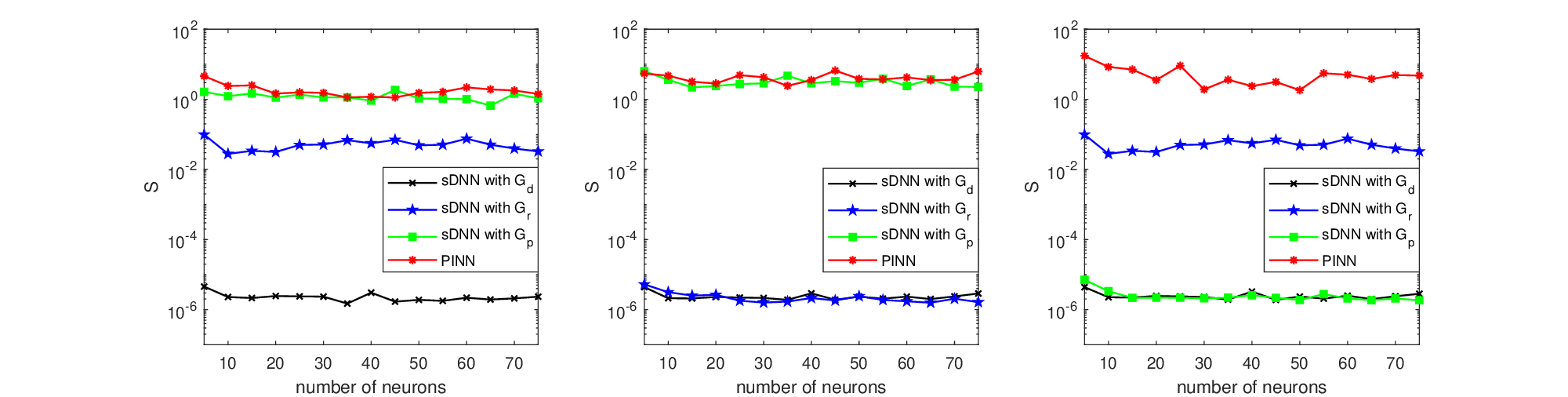}}
        \centerline{B}
    \end{minipage}
	\caption{(Color online)  Poisson equation:  Comparison of the metric of $u$ in different domains by each order sDNN and PINN. Keeping the number of neurons invariant: (A) symmetry metrics at different number of collocation points. Keeping the number of collocation points invariant: (B) symmetry metrics under different number of neurons. Each line represents the mean of five independent experiments. }
\label{fig31}
\end{figure}

Moreover, we choose three representative domains to show the symmetry metrics of sDNN and PINN. In Figure \ref{fig31}(A-B), the symmetry metrics are defined from left to right as $S_{12}=\left\|u(x,t)_{pred}-u(t,x)_{pred}\right\|_{2}/\left\|u(x,t)_{pred}\right\|_{2}$, $S_{21}=\left\|u(x,t)_{pred}-u(t,-x)_{pred}\right\|_{2}/\left\|u(x,t)_{pred}\right\|_{2}$ and $S_{22}=\left\|u(x,t)_{pred}-u(-x,t)_{pred}\right\|_{2}/\left\|u(x,t)_{pred}\right\|_{2}$. The sDNN with $G_d$ maintains a very low metric in each domain, with the black solid line reaching $7.29 \times 10^{-6}$ at domain $12$ when $N_f=1000$. However, the sDNN with $G_r$ is low only in $21$ domain where the maximum is $5.26 \times 10^{-6}$, while the blue solid line in domains $12$ and $22$ fluctuates between $10^{-2}$ and $10^{-1}$. Similarly, the sDNN with $G_p$ remains small symmetry metric only in $22$ domain, but the red solid line representing PINN always fluctuates largely. Therefore, the sDNN builds the connection between the red domains and the sampling 11 domain via the finite group and keeps the symmetry properties very well, but the grey domains violate the symmetry properties, where the PINN only takes effect in the sampling 11 domain and has bad solution extrapolation ability.

$\bullet$ \emph{Performances for different number of neurons per layer.}

The second set of experiments aims to study the performance of sDNN and PINN for different number of neurons. We still choose three hidden layers for the sDNN and PINN where the number of neurons in each hidden layer ranges from $5$ to $75$ with step $5$, and use $N_{u} = 100$ initial points and boundary points and $48$ collocation points selected from the total $N_f=100$ collocation points in the first quadrant. Figure \ref{fig32} shows that though the training only occurs in the 11 domain in Figure \ref{fig28} (A-C), the $L_2$ relative errors in the red domains are well predicted while the ones in the grey domains present poor jobs which are similar with the results of PINN. Moreover, the accuracies in the red domain has one order of magnitude lower than the grey domains, in particular, in the 12 domain, the sDNN with $G_d$ has one order lower than sDNN with $G_r$, and three order lower than both the sDNN with $G_p$ and the PINN. More interesting, the sDNN with $G_r$ seems to have better solution extrapolation ability than the sDNN with $G_p$ to predict the solution in the grey domain. For example, though the 12, 32 and 42 domains are not the predicted domains of the sDNN with $G_r$ and the sDNN with $G_p$, but the sDNN with $G_r$ present better prediction power than the sDNN with $G_p$ where the former reaches $10^{-2}$ accuracy but the latter only arrives at $10^0$.
\begin{figure}[ht]
\hspace{-0.7cm}\begin{minipage}{1.1\linewidth}
		\vspace{3pt}
		\centerline{\includegraphics[width=\textwidth]{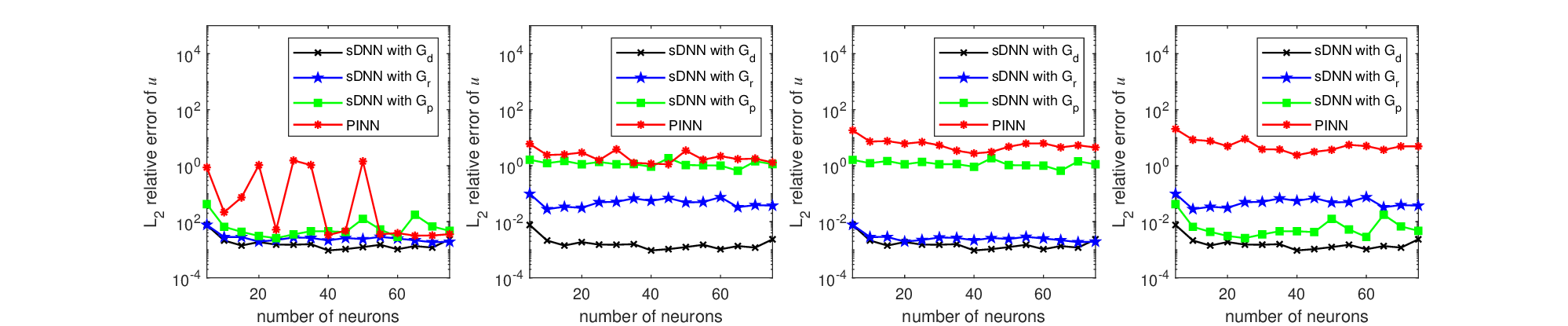}}
        \centerline{A}
	\end{minipage}
	
    \vspace{5pt} 

	\hspace{-0.7cm}\begin{minipage}{1.1\linewidth}
		\vspace{3pt}
		\centerline{\includegraphics[width=\textwidth]{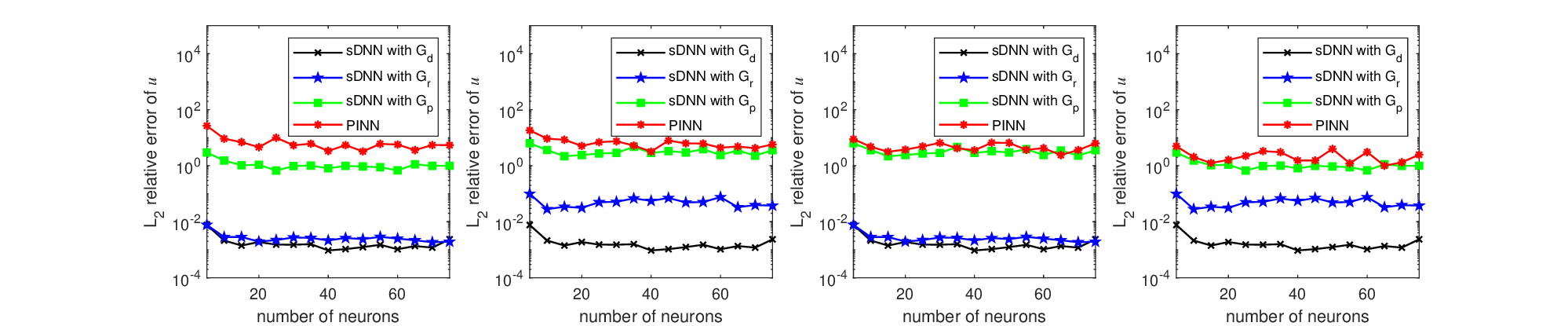}}
        \centerline{B}
	\end{minipage}	
	\caption{(Color online) Poisson equation:  Comparison of $L_2$ relative errors of $u$ in different domains by each order sDNN and PINN. Keeping the number of collocation points invariant, the $L_2$ relative errors under different number of neurons: (A) From left to right, domains $11$, $12$, $21$, and $22$ are represented; (B) From left to right, domains $31$, $32$, $41$, and $42$ are represented.  Each line represents the mean of five independent experiments.  }
\label{fig30}
\end{figure}

Furthermore, we choose a network structure with the same number of parameters in the first group of experiments and show the absolute errors of the four methods in Figure \ref{fig32}(A-D) where the $11$, $12$, $21$ and $22$ domains in Figure \ref{fig28} are considered respectively. Specifically, Figure \ref{fig32}(A) for the sDNN with $G_d$ shows that the tendencies of four graphs remains flat in the sampling domain and the three non-sampling domains while in Figure \ref{fig32}(B) for the sDNN with $G_r$ only the 11 and 21 domains has the similar flat tendencies but the 12 and 22 domains occur big fluctuations. Meanwhile, Figure \ref{fig32}(C) for the sDNN with $G_p$ presents a situation of large fluctuations in $12$ and $21$ domains and the maximum absolute error reaches $3.57$ because the 2-order group $G_p$ only connect the 11 training domain with the 22 prediction domain. The PINN does poor jobs in the four domains where the maximum absolute errors of in Figure \ref{fig32}(D) are $3.02$, $2.67$, $2.77$ and $2.74$, respectively. The comparison results show that the sDNN outperforms the PINN, and for the sDNN, as the order of finite group increases, one only needs a comparatively smaller training domain and then improve the accuracies of predicted solutions in and beyond the sampling domain simultaneously.

\begin{figure}[htp]
    \begin{minipage}{1.0\linewidth}
		\vspace{3pt}
		\centerline{\includegraphics[width=\textwidth]{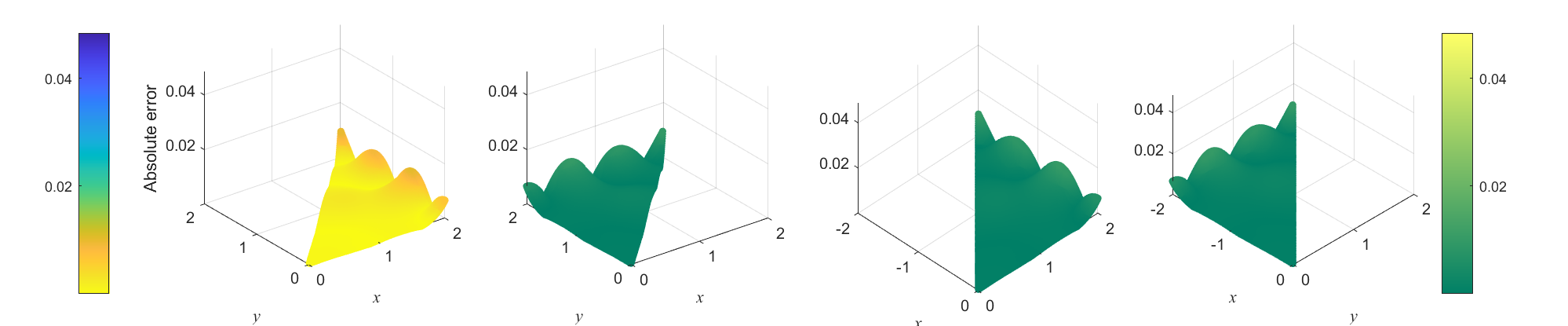}}
        \centerline{A}
	\end{minipage}

    \vspace{5pt} 

    \begin{minipage}{1.0\linewidth}
		\vspace{3pt}
		\centerline{\includegraphics[width=\textwidth]{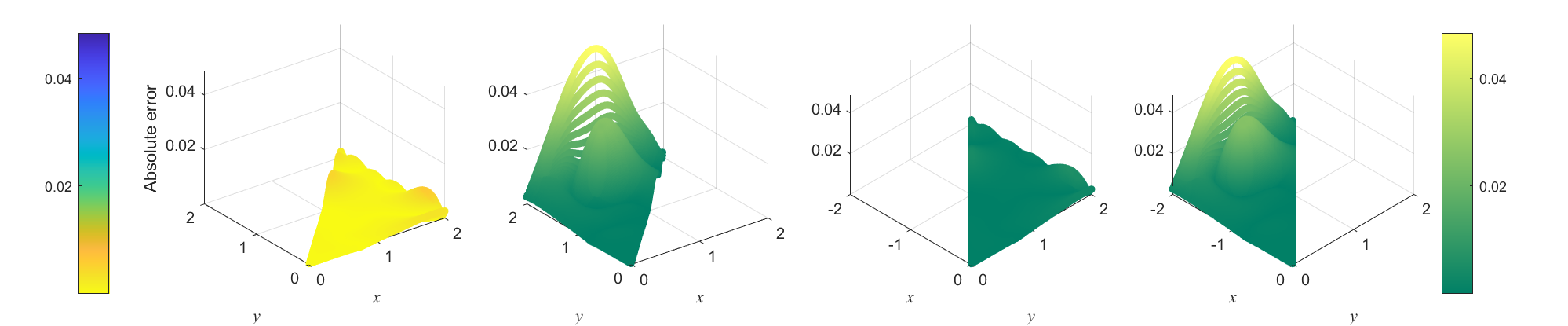}}
        \centerline{B}
	\end{minipage}

    \vspace{5pt} 

    \begin{minipage}{1.0\linewidth}
		\vspace{3pt}
		\centerline{\includegraphics[width=\textwidth]{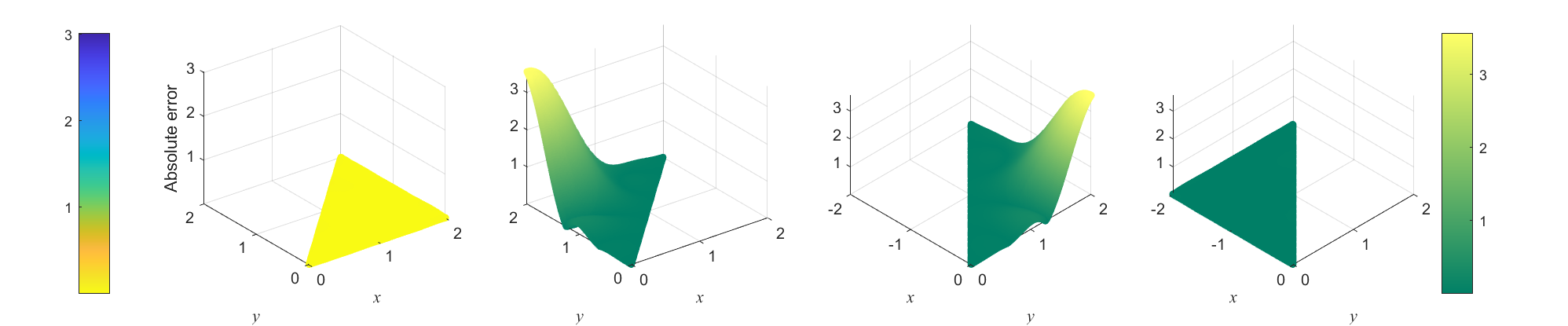}}
        \centerline{C}
	\end{minipage}

    \vspace{5pt} 

    \begin{minipage}{1.0\linewidth}
		\vspace{3pt}
		\centerline{\includegraphics[width=\textwidth]{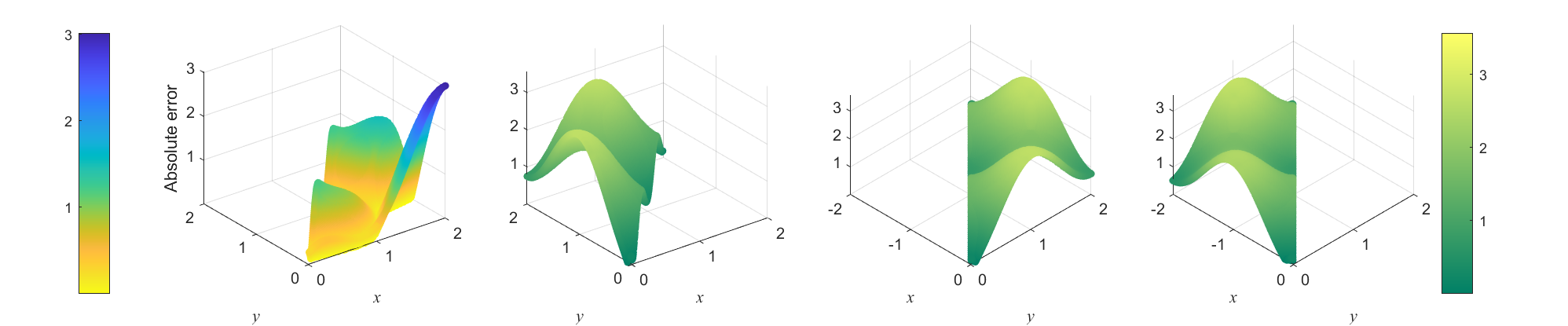}}
        \centerline{D}
	\end{minipage}
	\caption{(Color online) Poisson equation:  Comparisons of the absolute errors of the sDNN and PINN. From left to right, $11$, $12$, $21$ and $22$ domains are represented. (A) Absolute errors by the sDNN with $G_d$; (B) Absolute errors by the sDNN with $G_r$; (C) Absolute errors by the sDNN with $G_p$; (D) Absolute errors by PINN. }
\label{fig32}
\end{figure}

\begin{figure}[htp]
	\begin{minipage}{0.24\linewidth}
		\vspace{3pt}
		\centerline{\includegraphics[width=\textwidth]{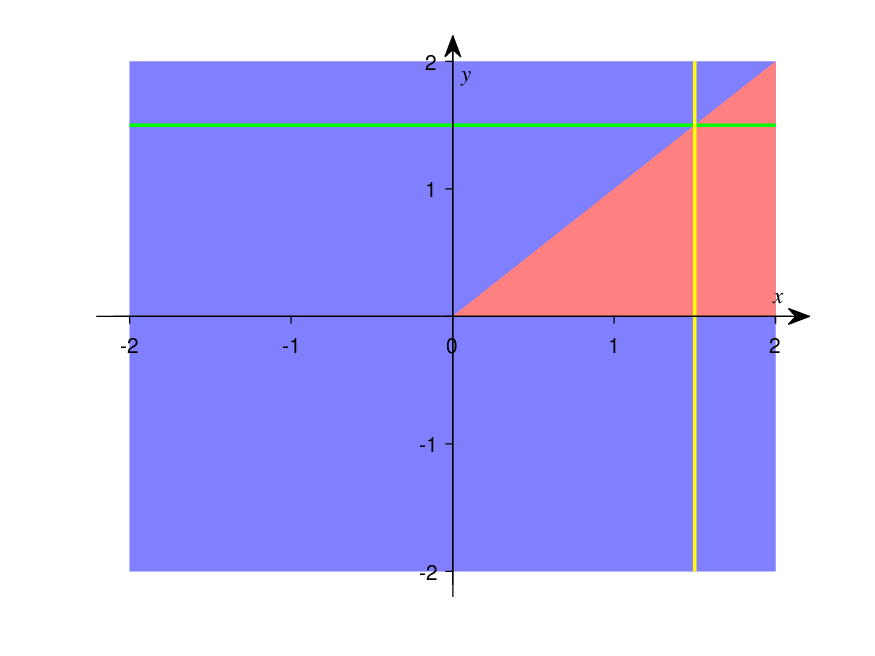}}
        \centerline{A}
	\end{minipage}
	\begin{minipage}{0.24\linewidth}
		\vspace{3pt}
		\centerline{\includegraphics[width=\textwidth]{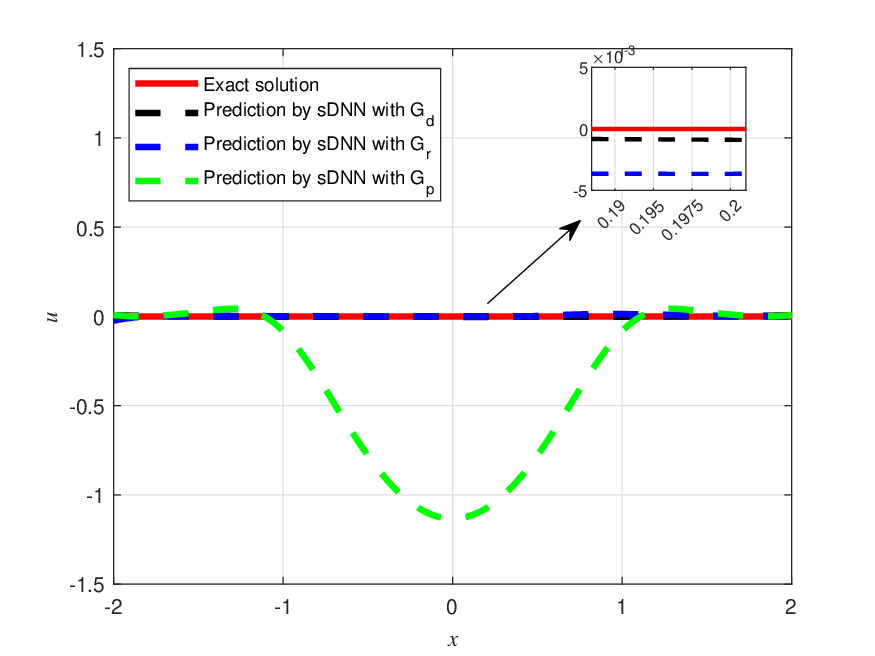}}
        \centerline{B}
	\end{minipage}
	\begin{minipage}{0.24\linewidth}
		\vspace{3pt}
		\centerline{\includegraphics[width=\textwidth]{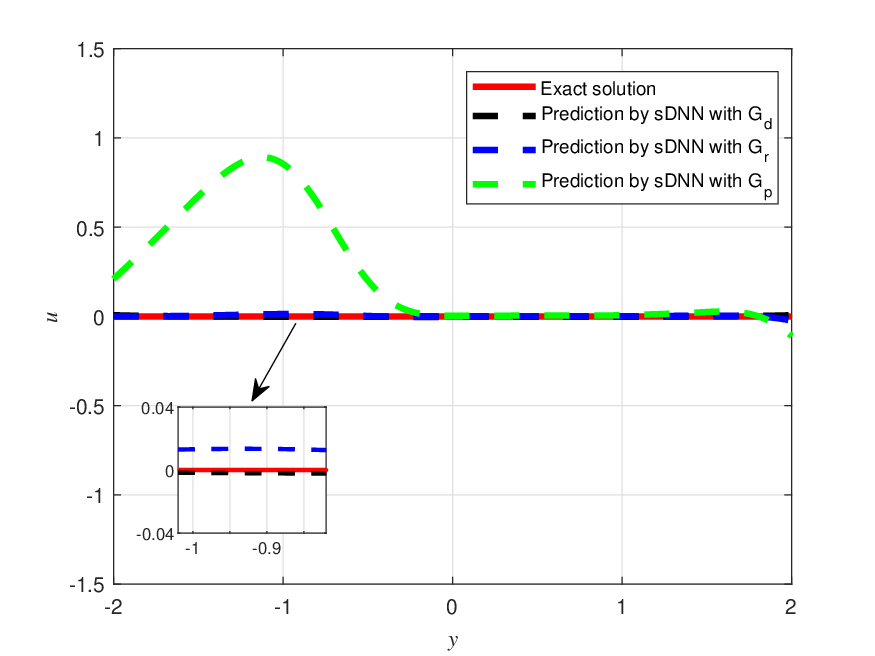}}
        \centerline{C}
	\end{minipage}
    \begin{minipage}{0.24\linewidth}
		\vspace{3pt}
		\centerline{\includegraphics[width=\textwidth]{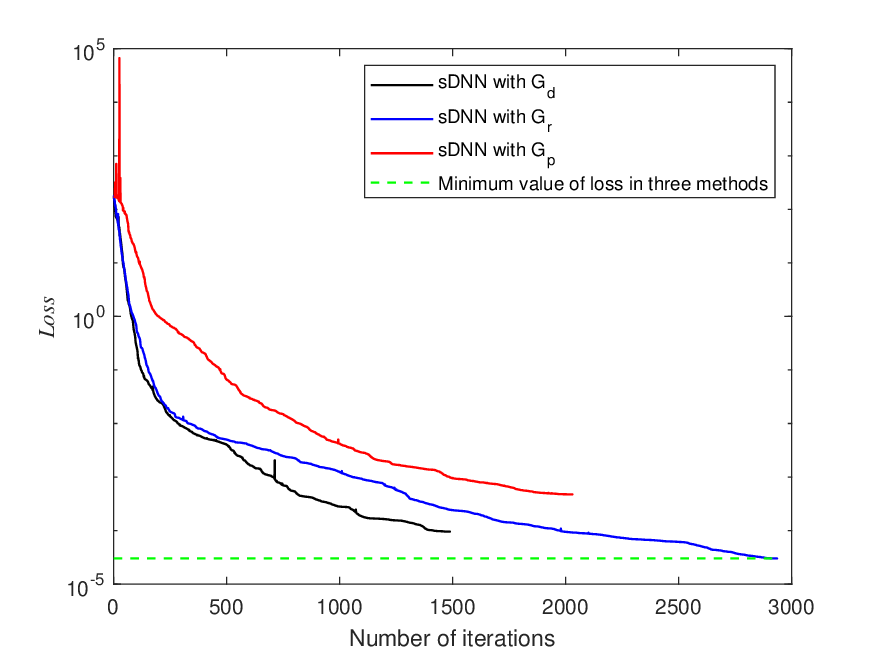}}
        \centerline{D}
	\end{minipage}

	\caption{(Color online)  Poisson equation: Cross sections of predicted solutions by PINN and sDNN and exact solutions. (A) Schematic diagrams of sampling and prediction domains where the green lines represent the two values of $t$ in cross section B and the yellow lines represent  the two values of $x$ in cross section C. (B) Cross sections for $y=1.5$. (C) Cross sections for $x=1.5$. (D) Loss histories for the sDNN over number of iterations. }
\label{fig33}
\end{figure}

In particular, under the condition of same number of parameters for the sDNN with the three finite groups, Figure \ref{fig33}(D) shows the loss histories of the three methods in the sampling domain where the sDNN with $G_d$ reaches $9.54 \times 10^{-5}$ after $1489$ iterations, while the sDNN with $G_r$ arrives at $3.02 \times 10^{-5}$ after $2936$ iterations and the sDNN with $G_p$ gets $4.75 \times 10^{-4}$ after $2031$ iterations.
However, the prediction accuracies in different domains presents a completely different scene. Figure \ref{fig33}(B) shows the predicted solution of sDNN with $G_d$ and exact solution at $y=1.5$ where they are highly coincident. Moreover, the inset in  Figure \ref{fig33}(B) shows that in the $12$ domain with $x\in[0,0.2]$ the blue dashed line is far more away from the red solid line than the black dashed line, while for $x\in[-1, 1]$, the green dotted line largely deviates with the red solid line.  Figure \ref{fig33}(C) shows the cross-section at $x=1.5$ where the inset displays that for $y\in[-1,-0.8]$, the black solid line is close to the red solid line while the blue dashed line has a certain distance with the red solid line, while for $y\in[-2,0]$ the green dotted line produces a big waveform away from the red solid line. 

\subsection{A nonlinear wave equation}
We consider an example whose admitted finite group has no matrix representation, and exemplify it by a nonlinear wave equation
\begin{eqnarray}\label{wave}
u_{tt}-uu_{xx}-(u_x)^{2}=f(x,t),~~~x\times t\in[1,2]\times[0,1],
\end{eqnarray}
which plays an important role in the field of mathematical physics and is used to describe one-dimensional nonlinear wave phenomena, such as water wave, plasma wave and so on \cite{aa-1981}.

Eq.(\ref{wave}) has an exact solution $u = [\sin(3x)\cos t + \sin(3 \lambda-3x)\cos( \mu-t)]/2$ with two parameters $\lambda$ and $\mu$, then $f(x, t)$ as well as the initial and boundary conditions are enforced by the solution $u$.  Eq.(\ref{wave}) is allowed by a finite group $g_1:(x,t,u(x,t))\mapsto(\lambda-x, \mu-t,u(x, t))$. Since $G_t=\{g_0,g_1\}$ has no matrix representation, then by Theorem \ref{th-general} we first expand the input data set by means of $G_t$ but keep the weight matrix and bias vector in the first hidden layer the same as the ones of PINN, and then enlarge the widths of other hidden layers by the order of $G_t$.

\begin{prop}
Suppose Eq.(\ref{wave}) is admitted by the finite group $G_t$ and $\x^{(0)}$ is the input data set. Then the input data set $\x^{(0)}$ is extended into
\begin{eqnarray}
&&\no X^{(0)}=\left(
            \begin{array}{c}
               g_0\x^{(0)}\\ g_1\x^{(0)}\\
            \end{array}
          \right),
\end{eqnarray}
and the weight matrixes and bias vectors of the sDNN with $L-1$ hidden layers which keeps the group $G_t$ invariant take the form
\begin{eqnarray}
&&\no \hspace{-0.7cm}\text{Weight matrixes:}~\left(
             \begin{array}{c}
              {\w}_1 \\
             \end{array}
           \right),~~~\left(
             \begin{array}{cc}
               \w_2^{(1)} & \w_2^{(2)} \\
               \w_2^{(2)} & \w_2^{(1)} \\
             \end{array}
           \right),~\dots,~\left(
             \begin{array}{cc}
               \w_{L-1}^{(1)} & \w_{L-1}^{(2)} \\
               \w_{L-1}^{(2)} & \w_{L-1}^{(1)} \\
             \end{array}
           \right),~\left(
  \begin{array}{c}
    \w_L \\
   \w_L\\
  \end{array}
\right);\\
&& \no\text{Bias vectors:}~ \left(
  \begin{array}{c}
    \b_1
  \end{array}
\right),~\left(
  \begin{array}{c}
    \b_2 \\
    \b_2
  \end{array}
\right),~\dots,,~\left(
  \begin{array}{c}
    \b_{L-1} \\
    \b_{L-1}
  \end{array}
\right)~\left(
  \begin{array}{c}
    \b_{L}
  \end{array}
\right).
\end{eqnarray}
where ${\w}_1$ is an initialized weight matrix with $n_1$ rows and $2$ columns.
\label{prop-wave}
\end{prop}

Therefore, if domain $x\times t \in[ 1,2 ] \times [ 0,1 ]$ is chosen as the training domain, the domain defined by $[ \lambda-2,\lambda-1 ] \times [ \mu-1,\mu]$ can be predicted by the sDNN equipped with weight matrixes and bias vectors in Proposition \ref{prop-wave}. We first choose $\lambda = 3,\mu = 2$ as an illustrated example, and then perform a systematic investigation of prediction effects of sDNN by the two parameters.
Specifically, in order to obtain the training data, we discretize the spatial region $x \in [ 1,2 ]$ and the temporal region $t \in [ 0,1 ]$ to $N_{x} = N_{t} = 1001$ equidistance points respectively, then the solution $u$ is discretized into $1001 \times 1001$ data points in the domain $[ 1,2 ] \times [ 0,1 ]$. Similar to the above examples, we conducted two sets of experiments to test the effects of sDNN and PINN for learning solution in and beyond the sampling domain. We define the symmetry metric as $S=\left\|u(x,t)_{pred}-u(\lambda-x,\mu-t)_{pred}\right\|_{2}/\left\|u(x,t)_{pred}\right\|_{2}$ to measure the degree of preserving the finite group property.
\begin{figure}[htp]
	\begin{minipage}{\linewidth}
		\centerline{\includegraphics[width=\textwidth]{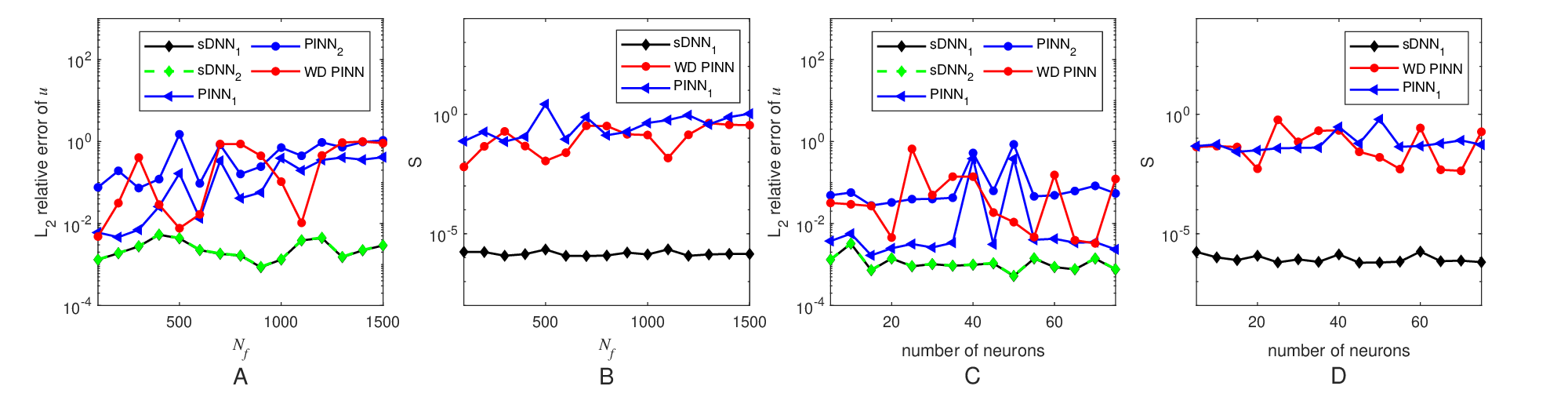}}
	\end{minipage}
	\caption{(Color online) Nonlinear wave equation: Comparisons of $L_2$ relative errors and the even metric of $u$ in the sampling and prediction domains by the two methods. Keeping the number of neurons invariant and varying the number of collocation points: (A) the $L_2$ relative errors; (B) the symmetry metrics. Keeping the number of collocation points invariant and varying the number of neurons: (C) the $L_2$ relative errors; (D) the symmetry metrics. Each line represents the mean of five independent experiments. Note that the sDNN and PINN with subscript $1$ and $2$ of denote the sampling domain and prediction domain respectively, and the symbol `WD PINN' means the training in the whole domain.}
\label{fig25}
\end{figure}

$\bullet$ \emph{Performances for different number of collocation points.}
In the first experimental group, both sDNN and PINN have three hidden layers with five neurons per layer, except that the second layer in PINN contains $10$ neurons in order to maintain the same number of training parameters of two methods. Then, $N_{u} = 100$ initial and boundary points are randomly selected, and the Latin hypercube sampling is used to obtain the collocation points with the number ranging from $100$ to $1500$ with a step $100$. Figure \ref{fig25}(A) shows that the sDNN gives very close solution accuracies in the sampling and prediction domains with a maximum difference $8.47 \times 10^{-5}$, but the PINN does poor performances in and beyond the sampling domain where big fluctuations appear and most solution accuracies walk between $10^{-1}$ and $10^{-2}$ order of magnitude. In particular, when $N_{f} = 200$, the $L_{2}$ relative error of PINN reaches $4.64 \times 10^{-3}$ in the sampling domain and only $1.93 \times 10^{-1}$ in the prediction domain. Moreover, the $L_{2}$ relative error of PINN with $N_{f} = 200$ is close to the one of sDNN, but  Figure \ref{fig25} (B) displays that the symmetry metric of PINN is only $1.82 \times 10^{-1}$ which is much bigger than $1.72 \times 10^{-6}$ of sDNN.

$\bullet$ \emph{Performances for different number of neurons per layer.}
The second set of experiments aims to study the performances of sDNN and PINN for different numbers of neurons. Both sDNN and PINN use three hidden layers where the number of neurons in each hidden layer ranges from $5$ to $75$ with step $5$, $N_{f} = 100$ collocation points and $N_{u} = 100$ initial points and boundary points are used in the training. Figure \ref{fig25}(C) shows a similar situation as the case of varying collocation points. Especially, in the sampling domain, sDNN and PINN have very close $L_{2}$ relative errors with $15$ neurons, but in the prediction domain, sDNN reaches $7.36 \times 10^{-4}$ but PINN only get $2.75 \times 10^{-2}$.  In addition, Figure \ref{fig25}(D) shows that the symmetry metric of sDNN is always maintained at $10^{-6}$ orders of magnitude which has at least three orders of magnitude improvement over PINN, particularly for the case of $15$ neurons, the metric of sDNN has three orders of magnitude lower than the PINN although they have close $L_{2}$ relative errors.

\begin{figure}[htp]
    \begin{minipage}{0.24\linewidth}
		\vspace{3pt}
		\centerline{\includegraphics[width=\textwidth]{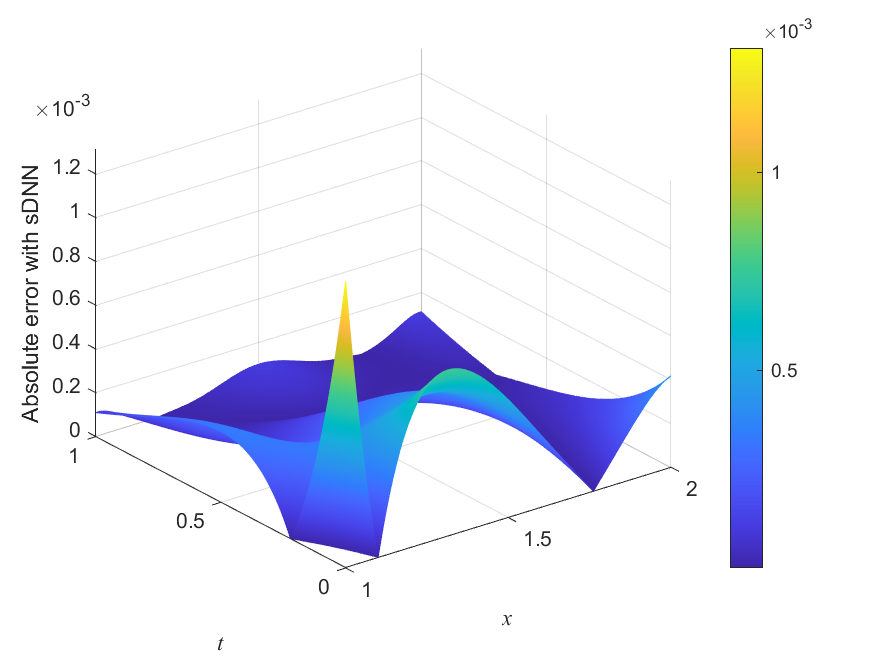}}
        \centerline{A}
	\end{minipage}
    \begin{minipage}{0.24\linewidth}
		\vspace{3pt}
		\centerline{\includegraphics[width=\textwidth]{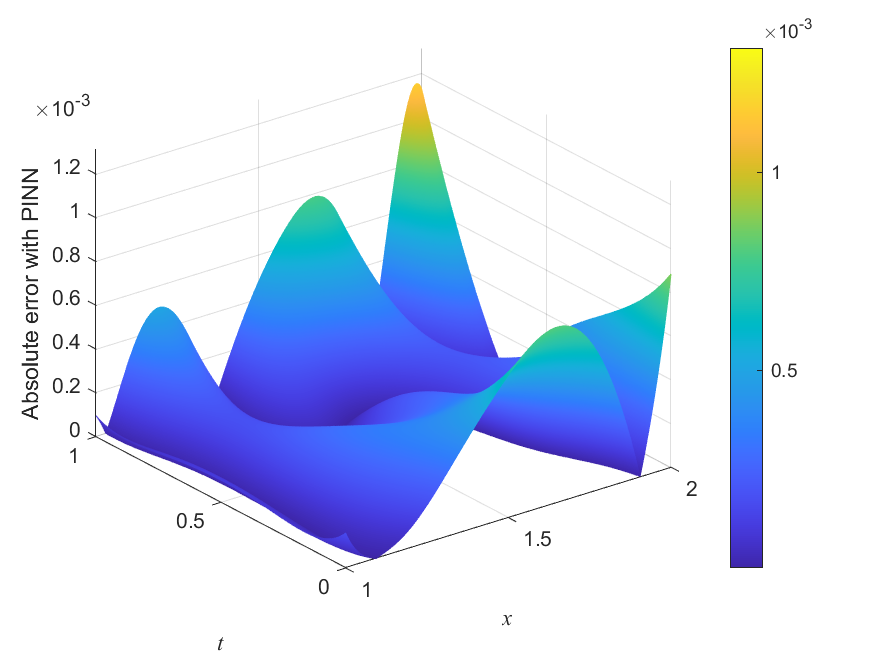}}
        \centerline{B}
	\end{minipage}
    \begin{minipage}{0.24\linewidth}
		\vspace{3pt}
		\centerline{\includegraphics[width=\textwidth]{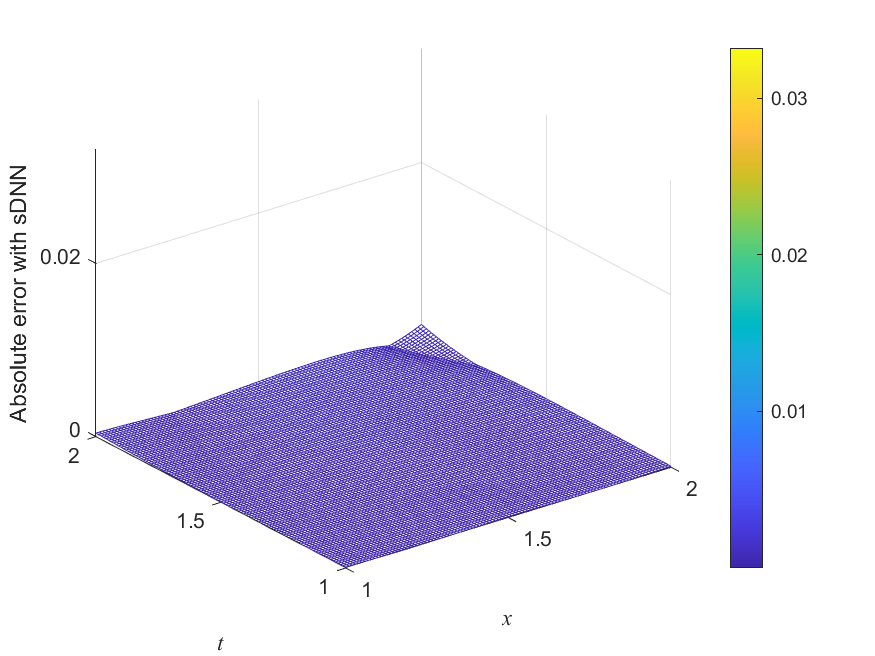}}
        \centerline{C}
	\end{minipage}
    \begin{minipage}{0.24\linewidth}
		\vspace{3pt}
		\centerline{\includegraphics[width=\textwidth]{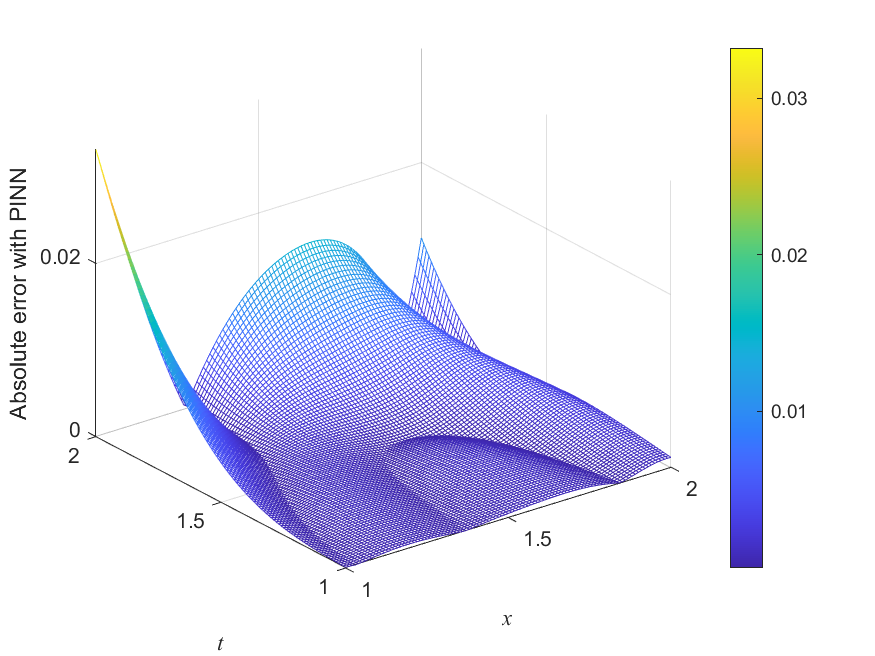}}
        \centerline{D}
	\end{minipage}
	\caption{(Color online) Nonlinear wave equation: Comparisons of the absolute errors of PINN and sDNN. (A) sDNN in sampling domain. (B) PINN in sampling domain. (C). sDNN beyond sampling domain. (D) PINN beyond sampling domain. }
	\label{fig26}
\end{figure}

In addition, in Figure \ref{fig26}, we choose the case of $15$ neurons to display the absolute errors of learned solutions in and beyond the sampling domain where sDNN and PINN have the most closest $L_{2}$ relative errors. Figure \ref{fig26} shows that the maximum absolute errors of sDNN and PINN are very close in the sampling domain, which are $1.31 \times 10^{-3}$ and $1.16 \times 10^{-3}$, respectively. However, in the prediction domain, PINN only reaches $3.31 \times 10 ^{-2}$ but sDNN has good solution extrapolation ability and keeps at $1.31 \times 10^{-3}$.

\begin{figure}[htp]
	\begin{minipage}{0.24\linewidth}
		\vspace{3pt}
		\centerline{\includegraphics[width=\textwidth]{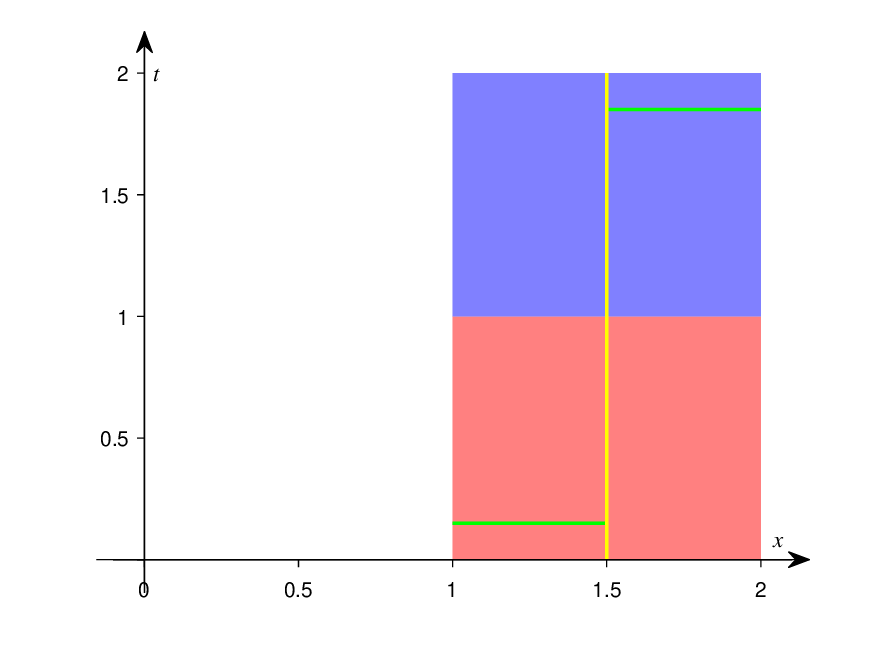}}
        \centerline{A}
	\end{minipage}
	\begin{minipage}{0.24\linewidth}
		\vspace{3pt}
		\centerline{\includegraphics[width=\textwidth]{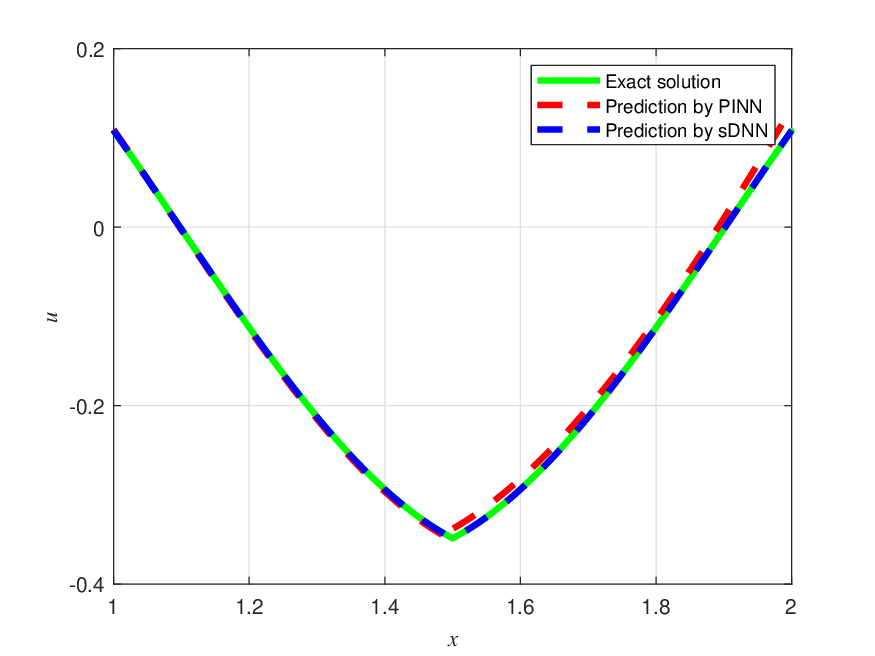}}
        \centerline{B}
	\end{minipage}
	\begin{minipage}{0.24\linewidth}
		\vspace{3pt}
		\centerline{\includegraphics[width=\textwidth]{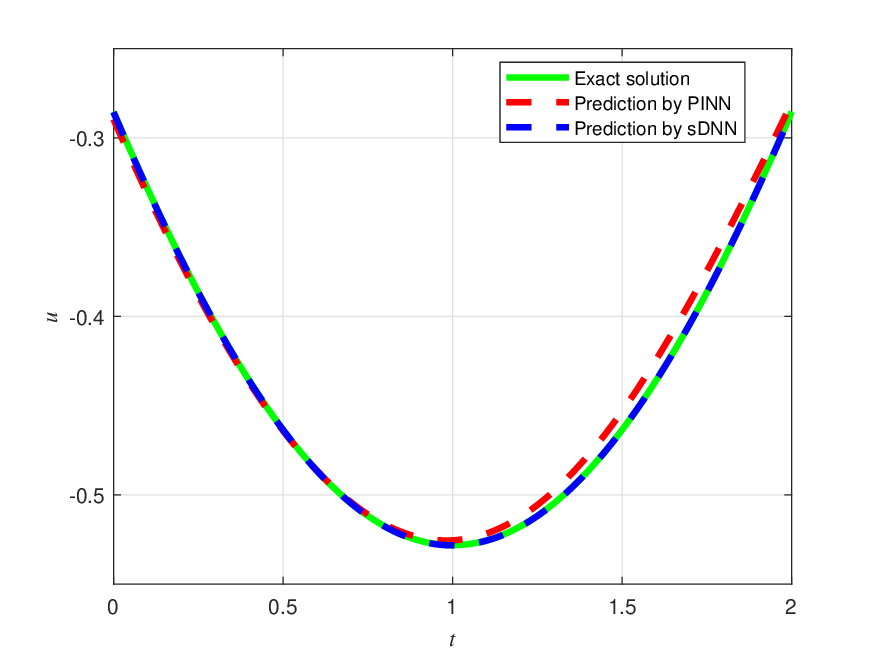}}
        \centerline{C}
	\end{minipage}
    \begin{minipage}{0.24\linewidth}
		\vspace{3pt}
		\centerline{\includegraphics[width=\textwidth]{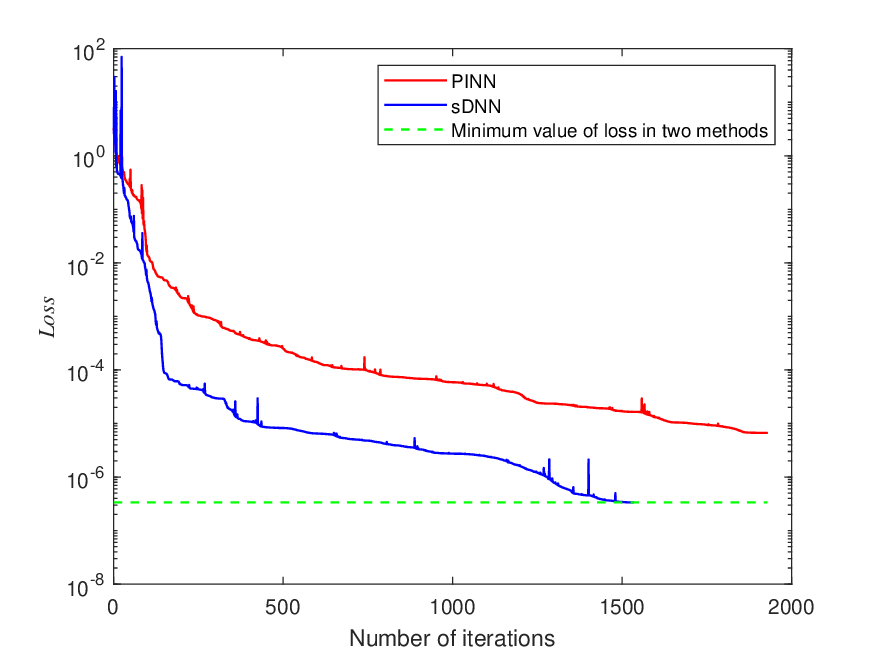}}
        \centerline{D}
	\end{minipage}
	\caption{(Color online) Nonlinear wave equation: Cross sections of predicted solutions by PINN and sDNN and exact solutions. (A) Schematic diagrams of sampling and prediction domains where the green lines represent the two values of $t$ in cross section B and the yellow lines represent  the two values of $x$ in cross section C.  (B) Cross sections for $t=0.15$ and $t=1.85$: when $x\in[1,1.5]$, then $t=0.15$; when $x\in[1.5,2]$, then $t=1.85$. (C) Cross section for $x=1.50$.  (D) Loss histories for the PINN and sDNN over number of iterations.}
\label{fig27}
\end{figure}

Furthermore, Figure \ref{fig27}(B) shows the cross section at $t=0.15$ where in the sampling domain with $x\in [1,1.5]$  the red and blue dotted lines fit well with the green solid line. However, beyond the sampling domain with $x\in[1.5,2]$ and $t=1.85$, the sDNN still coincide with exact solution but the PINN deviates the exact green line. Similar case occurs for the cross section at $x=1.5$ in Figure \ref{fig27}(C), where for $t\in[0,1]$, the sDNN, the PINN and the exact solution coincide completely, however, beyond the sampling domain with $t\in[1,2]$, as $t$ moves further away from $t=1$, the predicted solution by PINN diverges significantly from the exact solution, while the blue dotted line of the sDNN remains consistent with the green line of the exact solution. Moreover, the loss histories in Figure \ref{fig27}(D) further confirm the superiority of sDNN, where the loss value of PINN decreases to $6.69 \times 10^{-6}$ after $1928$ iterations, but the one of sDNN decreases to a more smaller value $3.37 \times 10^{-7}$ only with $1533$ iterations.

\begin{figure}[htp]
	\begin{minipage}{0.33\linewidth}
		\vspace{3pt}
		\centerline{\includegraphics[width=\textwidth]{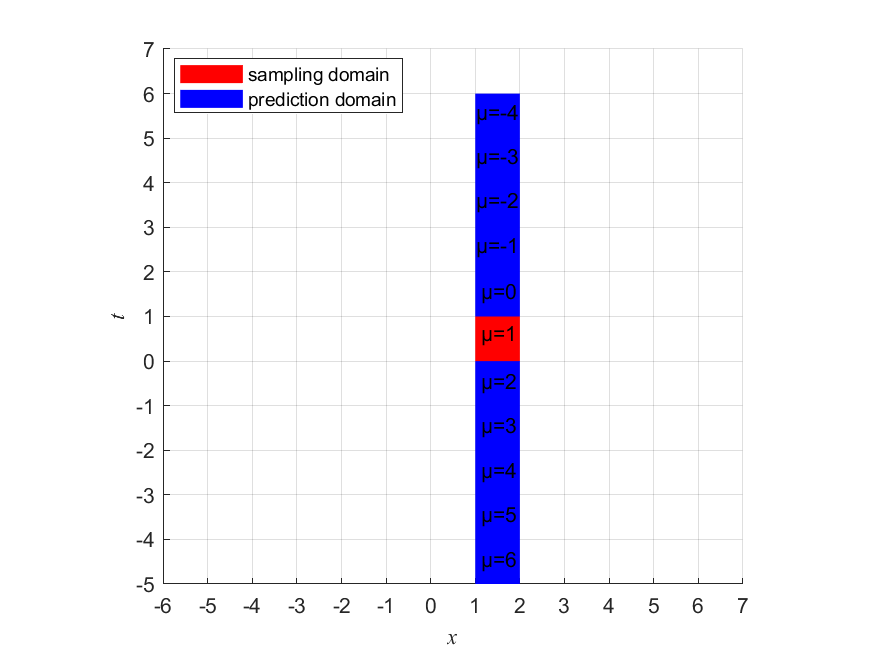}}
        \centerline{A}
	\end{minipage}
	\begin{minipage}{0.33\linewidth}
		\vspace{3pt}
		\centerline{\includegraphics[width=\textwidth]{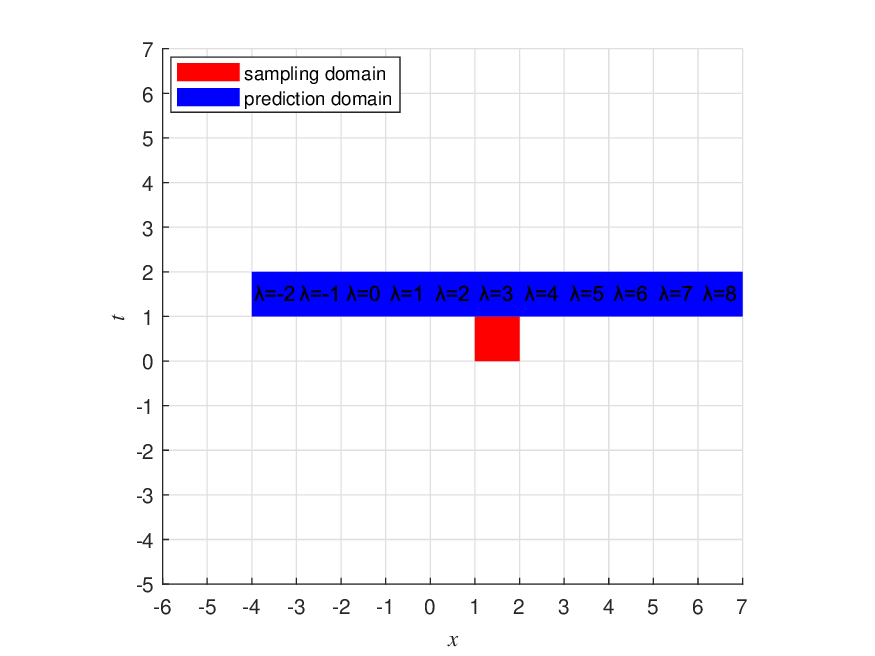}}
        \centerline{B}
	\end{minipage}
    \begin{minipage}{0.33\linewidth}
		\vspace{3pt}
		\centerline{\includegraphics[width=\textwidth]{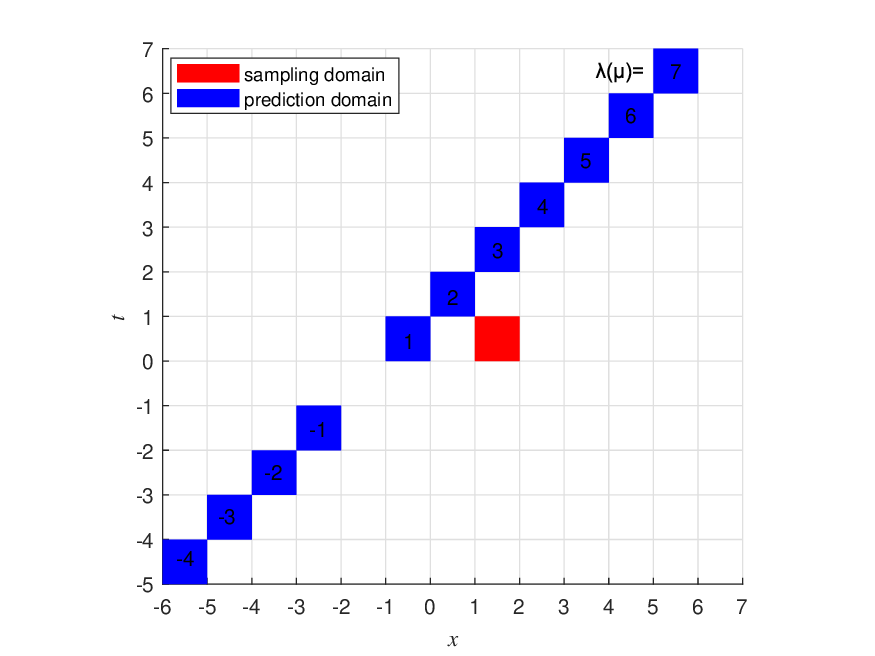}}
        \centerline{C}
	\end{minipage}
	\begin{minipage}{0.33\linewidth}
		\vspace{3pt}
		\centerline{\includegraphics[width=\textwidth]{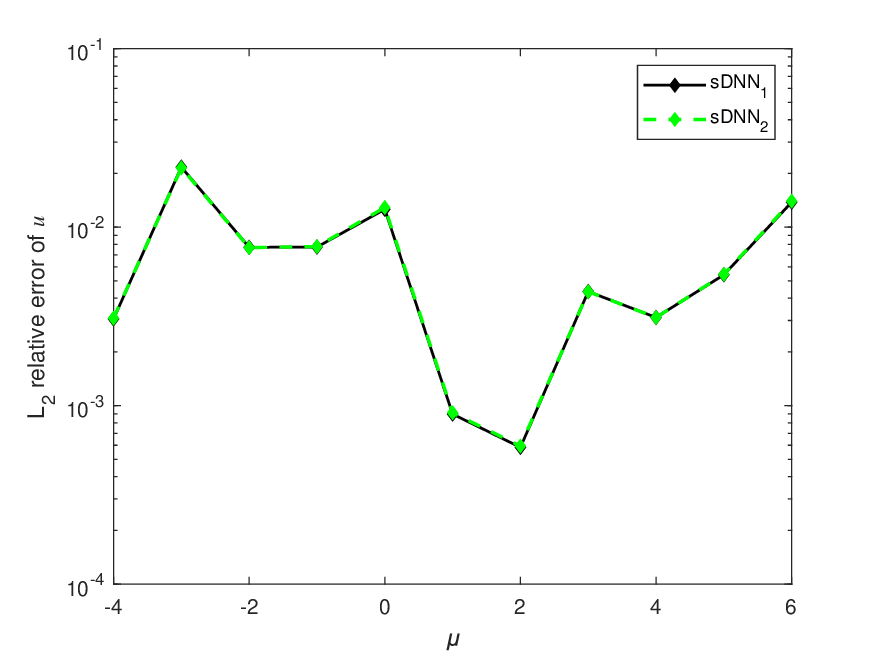}}
        \centerline{D}
	\end{minipage}
	\begin{minipage}{0.33\linewidth}
		\vspace{3pt}
		\centerline{\includegraphics[width=\textwidth]{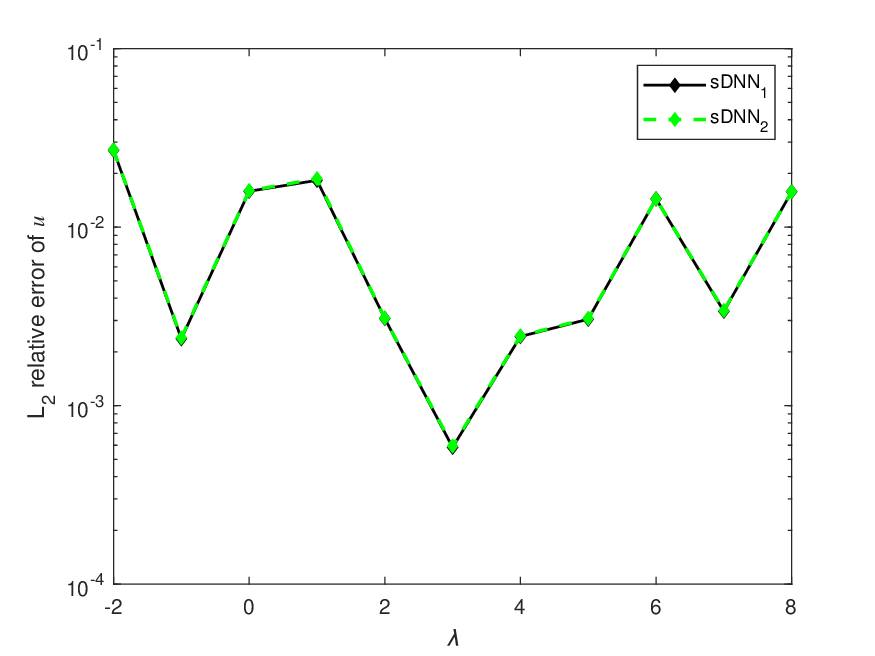}}
        \centerline{E}
	\end{minipage}
    \begin{minipage}{0.33\linewidth}
		\vspace{3pt}
		\centerline{\includegraphics[width=\textwidth]{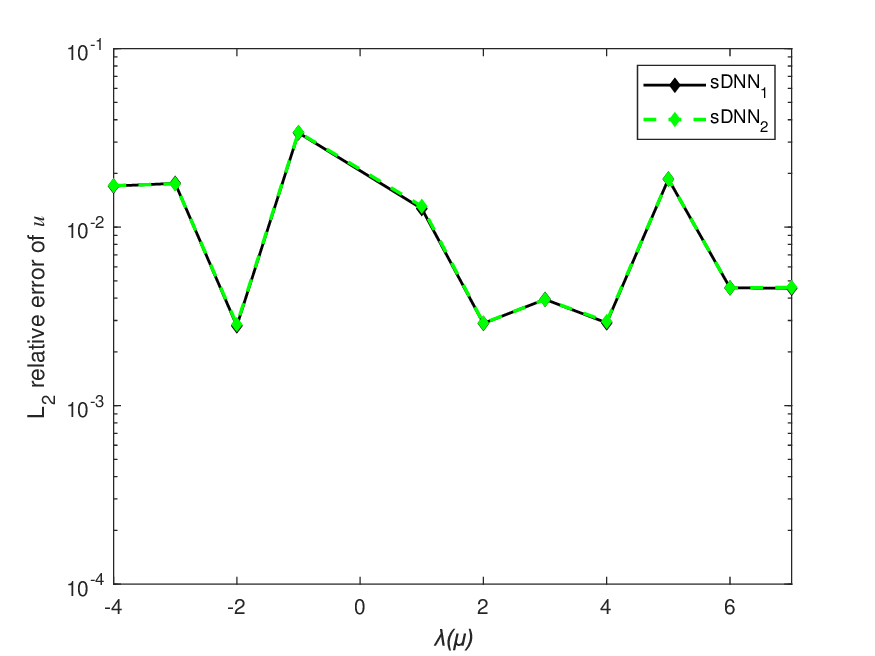}}
        \centerline{F}
	\end{minipage}
	\caption{(Color online) Nonlinear wave equation: The upper panel shows the red sampling domains and blue prediction domains. (A) $\mu$ changes and $\lambda = 3$; (B) $\lambda$ changes and $\mu = 2$; (C) both $\lambda = \mu$ change together. The lower panel (D-F) displays the $L_2$ relative errors of $u$ by the sDNN with the same changing parameters as (A-C). Note that the sDNN and PINN with subscript $1$ and $2$ of denote the sampling domain and prediction domain respectively. }
\label{fig36}
\end{figure}

$\bullet$ \emph{Further discussions about the prediction abilities with different parameters $\lambda$ and $\mu$.} We choose the $x\times t\in[1,2]\times[0,1]$ as the sampling domain and further consider three cases of $\lambda$ and $\mu$, given $\lambda=3$ and varying $\mu$, given $\mu=2$ and varying $\lambda$ and varying $\lambda=\mu$ together, where Figure \ref{fig36}(A-C) explicitly display the blue prediction domains which take the red sampling domains as the center respectively. Note that $\mu=\lambda=0$ makes the solution $u$ trivial and thus the corresponding prediction in Figure \ref{fig36}(C) is blank.

Figure \ref{fig36}(D-F) show the varying tendencies of the $L_2$ relative errors as the parameters $\lambda$ and $\mu$ change, where the black solid line denotes the $L_2$ relative errors in the red domains and the green dotted line stands for the $L_2$ relative errors in the corresponding blue domains.
In Figure \ref{fig36}(D), the smallest $L_2$ relative error occurs at $\mu=2$, where the prediction domain and the sampling domain keep high coincidence, $5.84 \times 10^{-4}$ and $5.94 \times 10^{-4}$ respectively. Then as $\mu$ walks away from $\mu=2$ along the two opposite directions, the $L_2$ relative errors fluctuate largely. It is interesting that the smallest $L_2$ relative error does not occur in the sampling domain $\mu=1$ but in the prediction domain $\mu=2$. The same situation also occurs for the case of varying $\lambda=\mu$ together, where the smallest $L_2$ relative error appears at $\lambda=\mu=-2$, not the nearest prediction domain $\lambda=\mu=2$ of the sampling domain in Figure \ref{fig36}(F), though both cases have very closest errors. However, the smallest $L_2$ relative errors of the case of given $\mu=2$ and varying $\lambda$ appear at the nearest prediction domain of the sampling domain coincidentally, which correspond to $\lambda=3$ in Figure \ref{fig36}(E). It is worthy of saying that by the sDNN, as the prediction domains walk away from the sampling domain, the predicted results have no regular tendencies for the three cases, but the predicted accuracies in and beyond sampling domain have highly similarities whether they are good or bad.

\subsection{Korteweg-de Vries equation}
At the end of this section, we consider the KdV equation
\begin{eqnarray} \label{KdV}
&& u_t+uu_{x}+u_{xxx}=f(x,t),~~~~x\times t\in\left[0.5, 2\right] \times \left[0.5, 2\right],
\end{eqnarray}
which arises in the theory of long waves in shallow water and the physical systems in which both nonlinear and dispersive effects are relevant \cite{KdV}.
Eq.(\ref{KdV}) has an exact solution $u(x,t)=(x + t)/(xt + 1)$ which is invariant under a reciprocal symmetry $g_r:(x,t,u(x,t))\mapsto(1/x,1/t,u(x,t))$.  Since $G_r=\{g_0, g_r\}$ has no matrix representation, according to Theorem $2.1$, we first extend the input data set with $G_r$, but keep the weight matrix and bias vector in the first hidden layer the same as PINN's, and then extend the widths of the other hidden layers in the order of $G_r$.

\begin{prop}
Suppose the solution of Eq.(\ref{KdV}) is invariant under the finite group $G_r$ and $\x^{(0)}$ is the input data set. Then the input data set $\x^{(0)}$ is extended to
\begin{eqnarray}
&&\no X^{(0)}=\left(
            \begin{array}{c}
               g_0\x^{(0)}\\ g_r\x^{(0)}\\
            \end{array}
          \right),
\end{eqnarray}
and the weight matrixes and bias vectors of the sDNN with $L-1$ hidden layers which keeps the group $G_r$ invariant take the form
\begin{eqnarray}
&&\no \hspace{-0.7cm}\text{Weight matrixes:}~\left(
             \begin{array}{c}
              {\w}_1 \\
             \end{array}
           \right),~\left(
             \begin{array}{cc}
               \w_2^{(1)} & \w_2^{(2)} \\
               \w_2^{(2)} & \w_2^{(1)} \\
             \end{array}
           \right),~\dots,~\left(
             \begin{array}{cc}
               \w_{L-1}^{(1)} & \w_{L-1}^{(2)} \\
               \w_{L-1}^{(2)} & \w_{L-1}^{(1)} \\
             \end{array}
           \right),~\left(
  \begin{array}{c}
    \w_L \\
   \w_L\\
  \end{array}
\right);\\
&& \no\text{Bias vectors:}~ \left(
  \begin{array}{c}
    \b_1
  \end{array}
\right),~\left(
  \begin{array}{c}
    \b_2 \\
    \b_2
  \end{array}
\right),~\dots,~\left(
  \begin{array}{c}
    \b_{L-1} \\
    \b_{L-1}
  \end{array}
\right),~\left(
  \begin{array}{c}
    \b_{L}
  \end{array}
\right).
\end{eqnarray}
where ${\w}_1$ is an initialized weight matrix with $n_1$ rows and $2$ columns.
\label{prop-kdv}
\end{prop}

Therefore, we choose $x\times t \in[0.5,1 ] \times [0.5,1 ]$ as the sampling domain to predict the domain $[ 1,2 ] \times [ 1,2 ]$ whose area is four times larger than the sampling domain, just shown in Figure \ref{fig39}(A). To obtain the training data, we discretize the spatial region $x \in[0.5,1 ]$ and the temporal region $t \in[0.5,1 ]$ to $N_x=N_t=101$ equidistant points, respectively, and then discretize the solution $u$ to $101 \times 101$ data points in the domain $[0.5,1]\times[0.5,1]$. In the experiment of Eq.(\ref{KdV}), we define the symmetry metric as
$S=\left\|u(x,t)_{pred}-u(1/x,1/t)_{pred}\right\|_{2}/\left\|u(x,t)_{pred}\right\|_{2}$, which measures the degree to which the network preserves symmetry.
\begin{figure}[htp]
	\begin{minipage}{\linewidth}
		\centerline{\includegraphics[width=\textwidth]{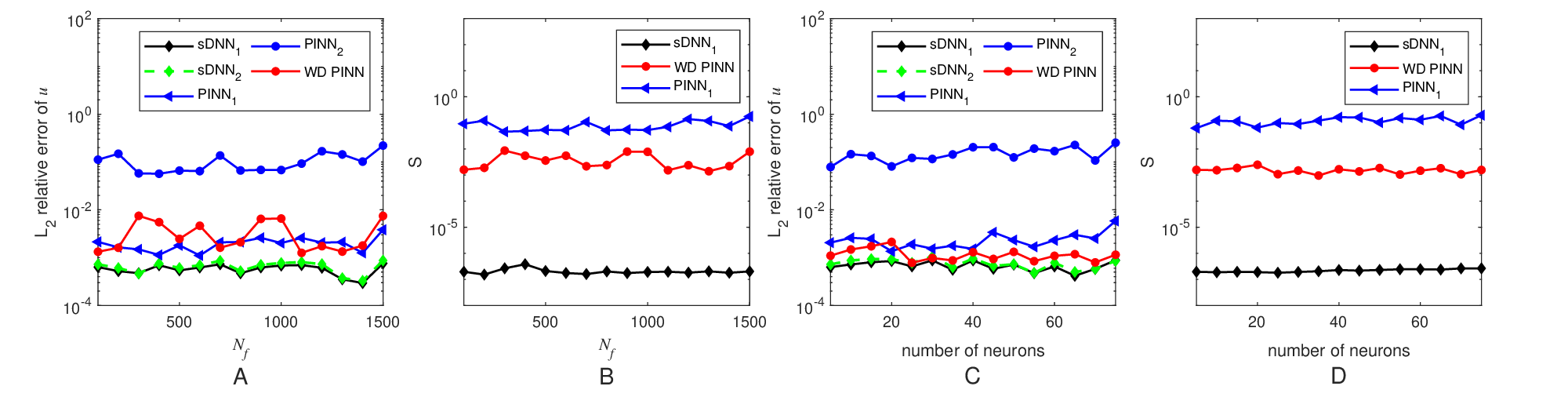}}
	\end{minipage}
	\caption{(Color online) Korteweg-de Vries equation: Comparisons of $L_2$ relative errors and the symmetry metrics of $u$ in the sampling and prediction domains by the two methods. Keeping the number of neurons invariant and varying the number of collocation points: (A) the $L_2$ relative errors; (B) the symmetry metrics. Keeping the number of collocation points invariant and varying the number of neurons: (C) the $L_2$ relative errors; (D) the symmetry metrics. Each line represents the mean of five independent experiments. Note that the sDNN and PINN with subscript $1$ and $2$ of denote the sampling domain and prediction domain respectively, and the symbol `WD PINN' means the training in the whole domain. }
\label{fig37}
\end{figure}

$\bullet$ \emph{Performances for different number of collocation points.}
In the first set of experiments, both sDNN and PINN have three hidden layers with five neurons in each layer, but the second layer of PINN contained $10$ neurons in order to keep the training parameters the same for both methods. We randomly choose $N_u=100$ data points in the initial and boundary data sets and use Latin hypercube sampling to obtain collocation points with a number between $100$ and $1500$ with step $100$.
Figure \ref{fig37}(A) shows that the $L_2$ relative errors of the sDNN in the sampling domain and the prediction domain are no longer coincide exactly, possibly because Eq.(\ref{KdV}) is not invariant under symmetry $g_r$. But they are still very close, with a maximum difference of $1.75\times10^{-4}$ and one order of magnitude improvement than the ones of PINN in the sampling domain. However, the PINN presents a poor performance outside the sampling domain, with a fluctuation of $L_2$ relative errors between $10^{-1}$ and $10^{-2}$ orders of magnitude. In particular, when $N_f=400$, though the $L_2$ relative error of PINN is close to the sDNN in the sampling domain, the PINN only gives $5.69\times10^{-2}$ in the prediction domain while the sDNN presents $7.56\times10^{-4}$, two orders of magnitude improvement. Furthermore, Figure \ref{fig37}(B) shows that the symmetry metrics of sDNN are stable at $10^{-7}$ while the PINN fluctuates between $10^{-3}$ and $10^{-1}$.

$\bullet$ \emph{Performances for different number of neurons per layer.}
The second set of experiments devotes to investigate the performances of sDNN and PINN about different numbers of neurons. Both sDNN and PINN employ three hidden layers, with the number of neurons in each hidden layer ranging from $5$ to $75$ with step $5$, $N_f=100$ collocation points and $N_u=100$ initial and boundary points are used in the training. Figure \ref{fig37}(C) shows a similar situation as the case of collocation points, but the PINN sampling in the whole domain (WD PINN) presents a comparatively excellent job.  For examle, when the number of neurons is $30$, the $L_2$ relative error of WD PINN is close to the $L_2$ relative error of sDNN in and beyond the sampling domain, but the symmetry metric of WD PINN in Figure \ref{fig37}(D) takes four orders of magnitude decrease than that of sDNN.

\begin{figure}[htp]
    \begin{minipage}{0.24\linewidth}
		\vspace{3pt}
		\centerline{\includegraphics[width=\textwidth]{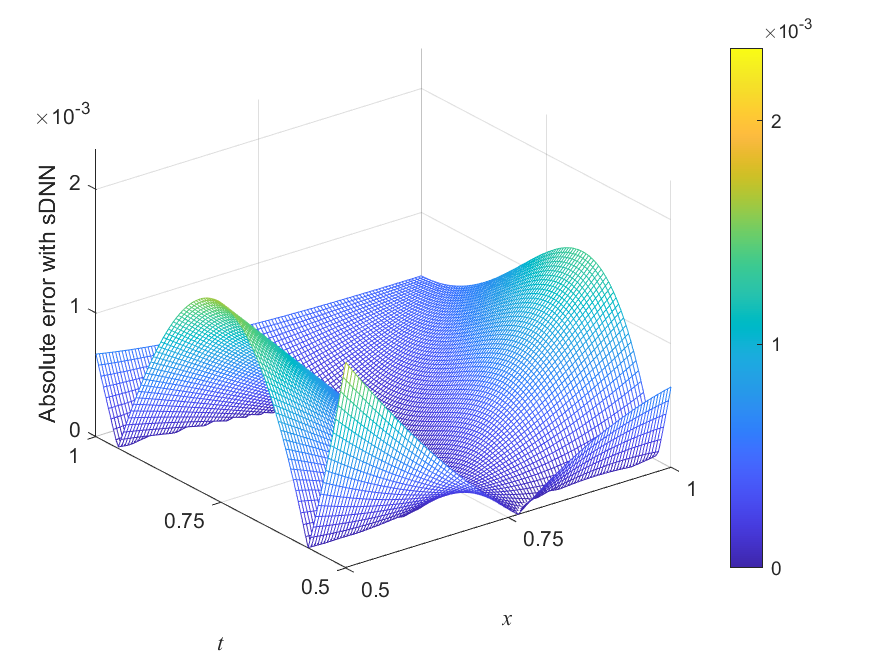}}
        \centerline{A}
	\end{minipage}
    \begin{minipage}{0.24\linewidth}
		\vspace{3pt}
		\centerline{\includegraphics[width=\textwidth]{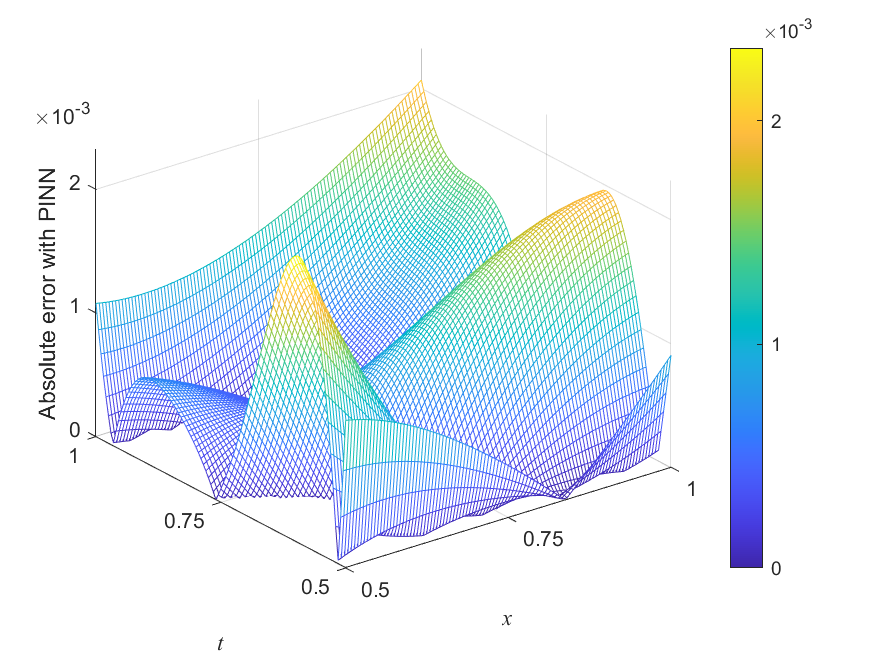}}
        \centerline{B}
	\end{minipage}
    \begin{minipage}{0.24\linewidth}
		\vspace{3pt}
		\centerline{\includegraphics[width=\textwidth]{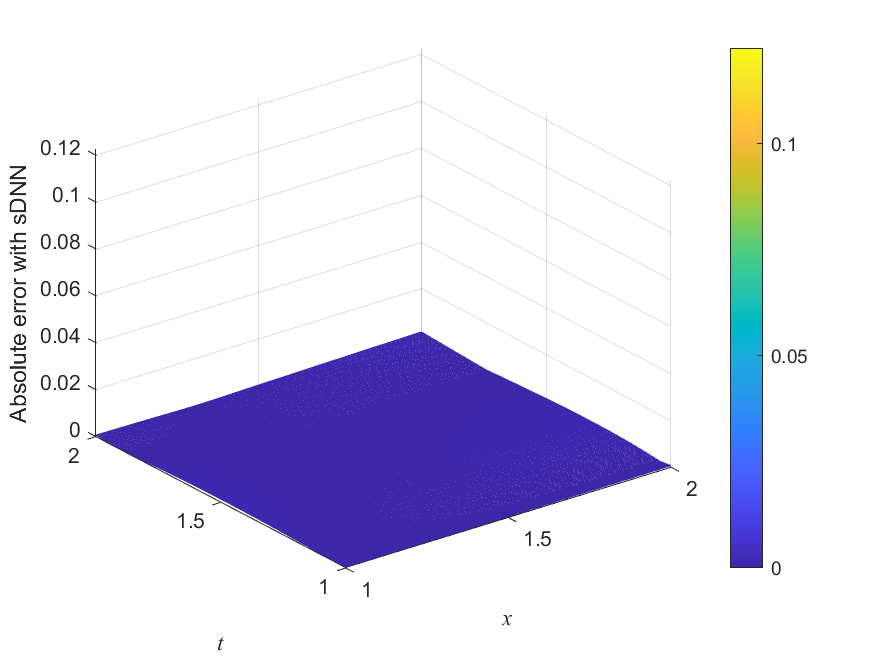}}
        \centerline{C}
	\end{minipage}
    \begin{minipage}{0.24\linewidth}
		\vspace{3pt}
		\centerline{\includegraphics[width=\textwidth]{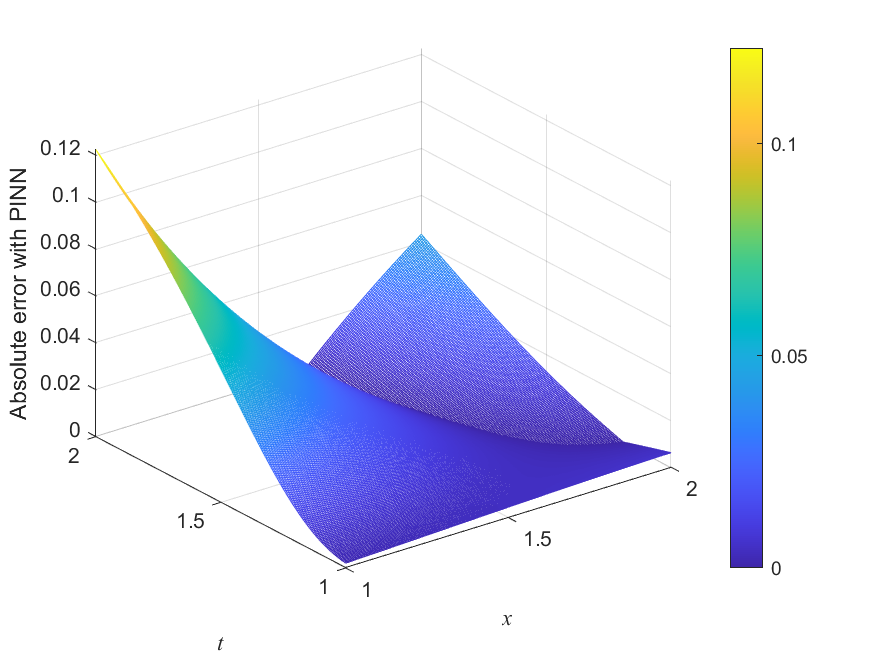}}
        \centerline{D}
	\end{minipage}
	\caption{(Color online) Korteweg-de Vries equation: Comparisons of the absolute errors of PINN and sDNN. (A) sDNN in sampling domain. (B) PINN in sampling domain. (C). sDNN beyond sampling domain. (D) PINN beyond sampling domain. }
	\label{fig38}
\end{figure}

\begin{figure}[htp]
	\begin{minipage}{0.24\linewidth}
		\vspace{3pt}
		\centerline{\includegraphics[width=\textwidth]{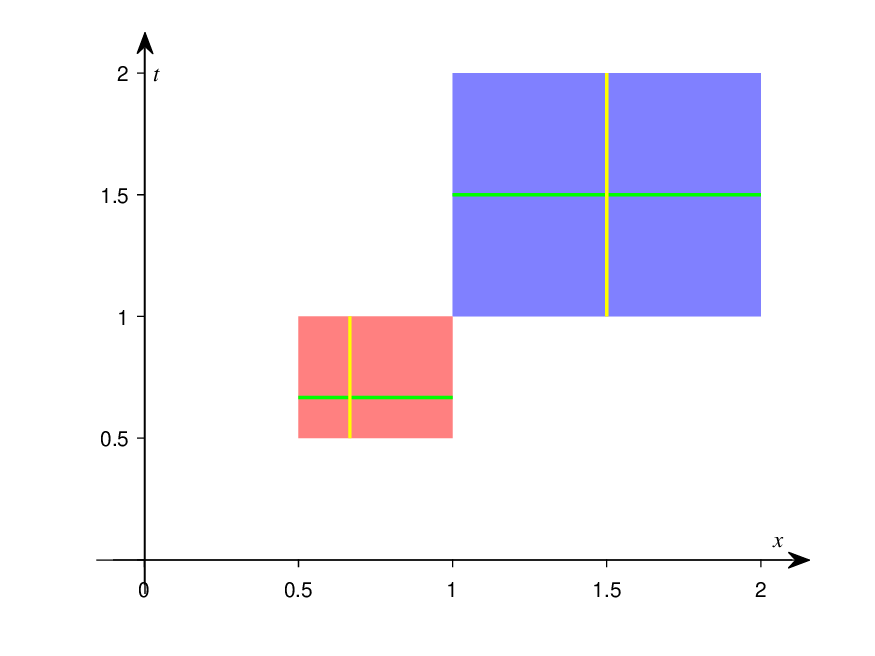}}
        \centerline{A}
	\end{minipage}
	\begin{minipage}{0.24\linewidth}
		\vspace{3pt}
		\centerline{\includegraphics[width=\textwidth]{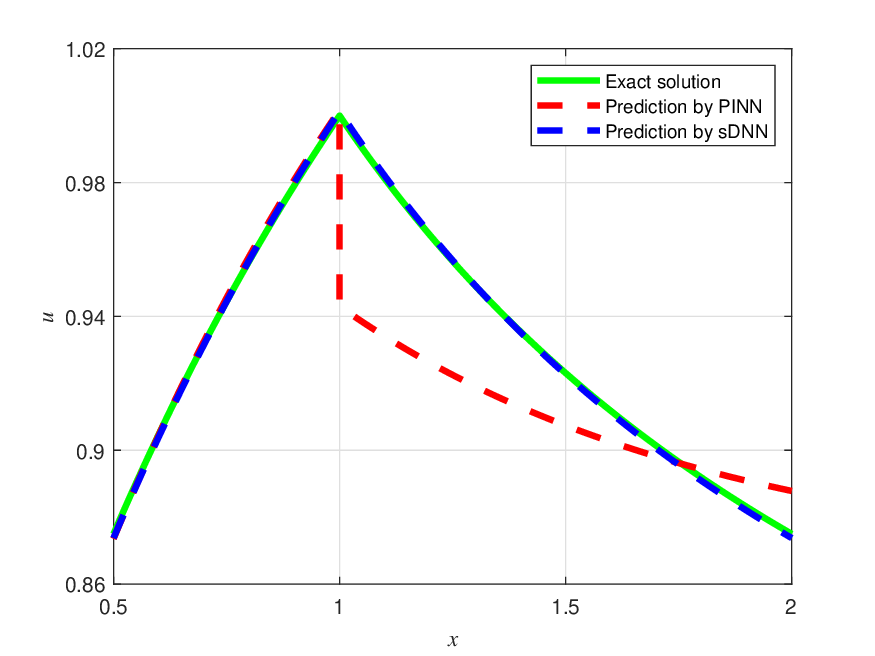}}
        \centerline{B}
	\end{minipage}
	\begin{minipage}{0.24\linewidth}
		\vspace{3pt}
		\centerline{\includegraphics[width=\textwidth]{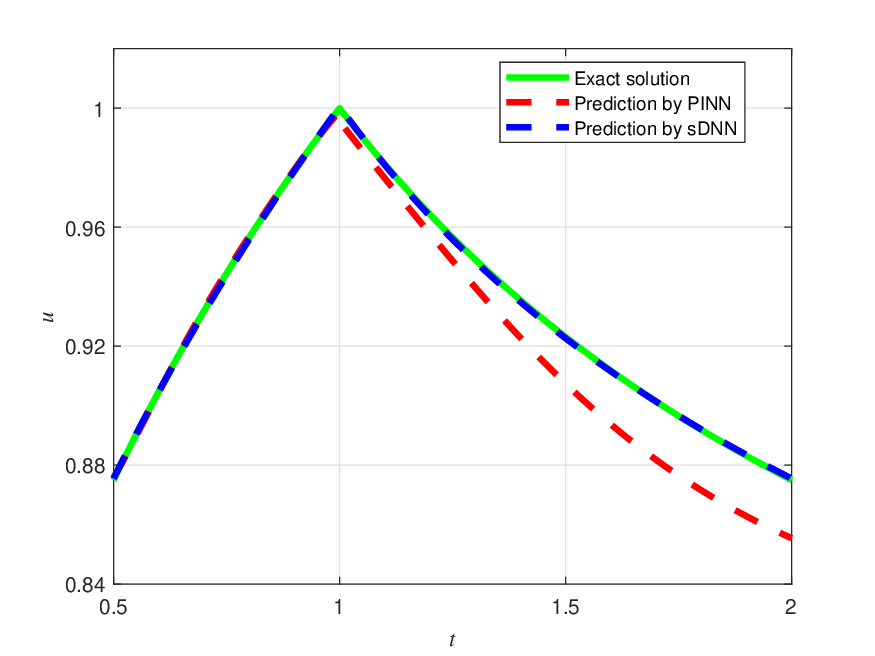}}
        \centerline{C}
	\end{minipage}
    \begin{minipage}{0.24\linewidth}
		\vspace{3pt}
		\centerline{\includegraphics[width=\textwidth]{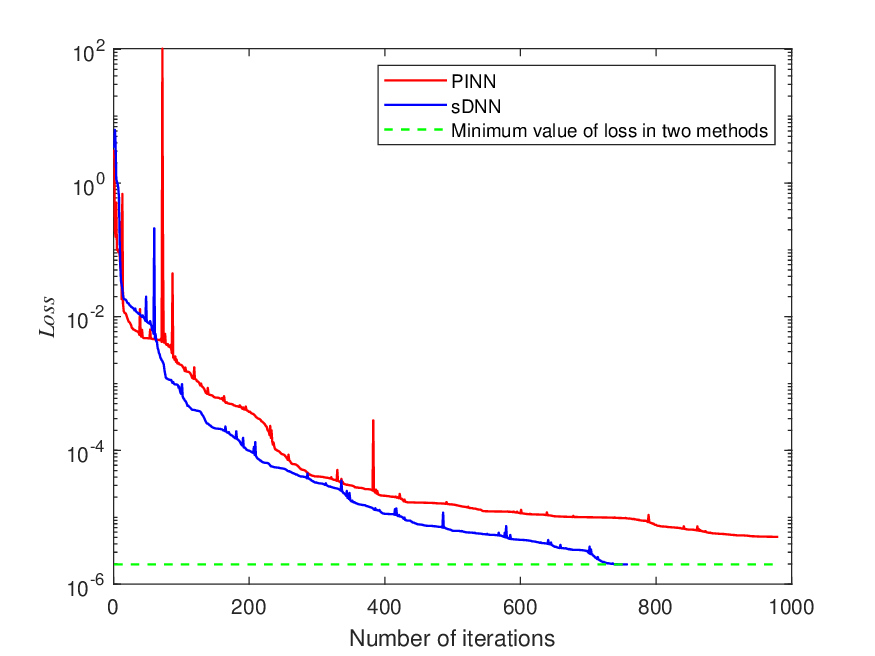}}
        \centerline{D}
	\end{minipage}
	\caption{(Color online) Korteweg-de Vries equation: Cross sections of predicted solutions by PINN and sDNN and exact solutions. {(A) Schematic diagrams of sampling and prediction domains where the green lines represent the two values of $t$ in cross section B and the yellow lines represent  the two values of $x$ in cross section C.}  (B) Cross sections for $t=2/3$ and $t=1.5$: when $x\in[0.5,1]$, then $t=2/3$; when $x\in[1,2]$, then $t=1.5$. (C) Cross sections for $x=2/3$ and $x=1.5$: when $t\in[0.5,1]$, then $x=2/3$; when $t\in[1,2]$, then $x=1.5$.  (D) Loss histories for the PINN and sDNN over number of iterations.}
\label{fig39}
\end{figure}

In addition, in Figure \ref{fig38}, we choose the case of $400$ collocation points to show the absolute errors of the learned solutions in and beyond of the sampling domain, where the sDNN and PINN have the closest $L_2$ relative error. Figure \ref{fig38}(A-B) shows that the maximum absolute errors of sDNN and PINN stay at the same order of magnitude, $1.66\times10^{-3}$ and $2.32\times10^{-3}$ respectively. However, in the prediction domain, PINN reaches $1.22\times10^{-1}$ of maximum absolute error while the sDNN remains at $1.66\times10^{-3}$, which shows that sDNN has a stronger solution extrapolation ability than the PINN.

Finally, along the two green line $t=2/3$ and $t=1.5$ in Figure \ref{fig39}(A), Figure \ref{fig39}(B) shows the two corresponding cross sections where in the sampling domain $t=2/3$ and $x \in [0.5, 1]$, both PINN and sDNN present good coincidence with the exact solution, but in the prediction domain $t=1.5$ and $x \in [1, 2]$, the sDNN still overlaps with the exact solution while the PINN largely deviates from the green line of exact solution. Figure \ref{fig39}(C) displays a similar situation in the cross sections at $x=2/3$ and $x=1.5$ which correspond to the yellow lines in  Figure \ref{fig39}(A). Moreover, Figure \ref{fig39}(D) demonstrates the strong learning ability of sDNN, where the loss value of PINN reaches $5.09\times10^{-6}$ after $980$ iterations, while the sDNN arrives at $1.98\times10^{-6}$ after only $758$ iterations.


\section{Conclusion}
We propose a sDNN for solving PDEs which exactly incorporates the finite order symmetry group of PDEs into the neural network architecture, where explicit structures of weight matrixes and bias vectors in each hidden layer are given whether the finite group has matrix representation or not. Consequently, the sDNN not only improve the accuracy in the sampling domain, but also exhibits strong solution extrapolation ability. Two groups of numerical experiments for each illustrated PDE validate that the sDNN outperforms PINN largely.

However, we have to admit that we are still in the very early stages of rigorously understanding the integration of physical laws of PDEs and neural networks, which motivate many future research directions. One problem is that, since the accuracies of both sampling domain and prediction domain are closely connected by the sDNN, then improving the accuracies in the sampling domain is a direct and efficient approach for highlighting the superiority of sDNN. Furthermore, since the neural networks that integrate the finite group is built upon the framework of PINN, {the variations of PINN, for example, the gradient PINN, the extended PINN and integrating merits of traditional numerical methods into PINN, can be used to further improve the performance of sDNN.} Therefore, certain novel approaches must be proposed to study the integration between the inherent properties of PDEs and the neural networks. Another obvious and thorny problem is how to insert the continuous symmetry of PDEs into the architecture of neural network. Unlike the finite group, the number of elements of continuous symmetry group is infinite and thus to a large extent the neural networks may have no succinct expression as the case of finite group. Such topics are already in progress and will be reported in our future work.
\section*{Acknowledgements}
The paper is supported by the Beijing Natural Science Foundation (No. 1222014), the National Natural Science Foundation of China (No. 11671014) and the Cross Research Project for Minzu University of China (No. 2021JCXK04).
\\\\\textbf{Declarations of interest: The authors have no conflicts to disclose.}


\begin{thebibliography}{99}
\bibitem{jr}J. Stoer and R. Bulirsch, Introduction to Numerical Analysis, New York, Springer-Verlag, 2013.
\bibitem{e-2018} E, W.N., Yu, B. The Deep Ritz Method: A deep learning-based numerical algorithm for solving variational problems, Commun. Math. Stat. 6 (2018) 1-12.
\bibitem{gm-2018}J. Sirignano, K. Spiliopoulos: DGM: A deep learning algorithm for solving partial differential equations. J. Comput. Phys. 375 (2018) 1339-1364.
\bibitem{2018a}M. Raissi, P. Perdikaris, G.E. Karniadakis, Physics-informed neural networks: A deep learning framework for solving forward and
inverse problems involving nonlinear partial differential equations, J. Comput. Phys. 378 (2019) 686-707.
\bibitem{at-2021} A. Zachary, T. Helen, AI accidents: An emerging threat, Center Secur. Emerg. Technol., Washington, DC, USA, Tech. Rep., 2021.
\bibitem{wang-2022} S. Wang, X. Yu, P. Perdikaris, When and why PINNs fail to train: A neural tangent kernel perspective, J. Comput. Phys. 449  (2022) 110768.
\bibitem{lmz-2021}L. Lu, X. Meng, Z. Mao, G.E. Karniadakis, DeepXDE: A deep learning library for solving differential equations, SIAM Rev. 63 (1) (2021) 208-228.
\bibitem{kim-2021} J. Kim, K. Lee, D. Lee, S.Y. Jhin, N. Park, DPM: a Novel Training Method for Physics-Informed Neural Networks in Extrapolation. In Proceedings of the AAAI
Conference on Artificial Intelligence. (2021)
\bibitem{ejg} E. Kharazmi, Z. Zhang, G.E. Karniadakis, hp-VPINNs: Variational physics-informed neural networks with domain decomposition, Comput. Methods Appl. Mech. Engrg.
 374 (2021) 113547.
\bibitem{J} J. Yu, L. Lu, X.H. Meng, G.E. Karniadakis, Gradient-enhanced physics-informed neural networks for forward and inverse PDE problems, Comput. Methods Appl. Mech. Engrg. 393 (2022) 114823.
\bibitem{jan}N. Lin, Y. Chen, A two-stage physics-informed neural network method based on conserved quantities and applications in localized wave solutions, J. Comput. Phys. 457 (2022) 111053.
\bibitem{zhang-2023a}Z.Y. Zhang, H. Zhang, L.S. Zhang, L.L. Guo, Enforcing continuous symmetries in physics-informed  neural network for solving forward and inverse problems of partial differential equations, J. Comput. Phys. 492 (2023) 112415.
\bibitem{tb-2022}T. Beucler, M. Pritchard, S. Rasp, J. Ott, P. Baldi, P. Gentine, Enforcing analytic constraints in neural networks emulating physical systems, Phys. Rev. Lett. 126 (2021) 098302.
\bibitem{zhu-2022}W. Zhu, W. Khademi, E.G. Charalampidis, P.G. Kevrekidis, Neural networks enforcing physical symmetries in nonlinear dynamical lattices: The case example of the Ablowitz-Ladik model, Physica D 434 (2022) 133264.
\bibitem{fang-2022}Z.W. Fang, A high-efficient hybrid physics-informed neural networks based on convolutional neural network, IEEE Trans. Neural Networks Learn. Syst. 33 (10) (2022) 5514-5526.
{\bibitem{pas-2023}P.A. Mistani, S. Pakravan, R. Ilango, F. Gibou, JAX-DIPS: neural bootstrapping of finite discretization methods and application to elliptic problems with discontinuities, J. Comput. Phys. 493 (2023) 112480.}
\bibitem{jag-2020b} A.D. Jagtap, G.E. Karniadakis, Extended physics-informed neural networks (xpinns): a generalized space-time domain decomposition based deep learning framework for nonlinear partial differential equations, Commun. Comput. Phys. 28 (5) (2020) 2002-2041.
    \bibitem{jag-2020a}A.D. Jagtap, E. Kharazmi, G.E. Karniadakis, Conservative physics-informed neural networks on discrete domains for conservation laws: applications to
forward and inverse problems, Comput. Methods Appl. Mech. Eng. 365 (2020) 113028.
    \bibitem{shu-2021} K. Shukla, A.D. Jagtap, G.E. Karniadakis, Parallel physics-informed neural networks via domain decomposition. J. Comput. Phys. 447 (2021) 110683.
\bibitem{ja-2020} A.D. Jagtap, K. Kawaguchi, G.E. Karniadakis, Locally adaptive activation functions with slope recovery for deep and physics-informed neural networks, Proc. R. Soc. A Math. Phys. Eng. Sci. 476 (2020) 20200334.
\bibitem{zhang-2023b} Z.Y. Zhang, S.J. Cai, H. Zhang, A symmetry group based supervised learning method for solving partial differential equations, Comput. Methods Appl. Mech. Engrg.  414 (2023) 116181.
\bibitem{wang-2021}S. Wang, Y. Teng, P. Perdikaris, Understanding and mitigating gradient pathologies in physics-informed neural networks. SIAM J. Sci. Comput. 43(5)(2021) A3055-A3081.
\bibitem{liu2023} S. Liu, C. Su, J. Yao, Z. Hao, H. Su, Y. Wu, J. Zhu, Preconditioning for Physics-Informed Neural Networks, (2024) arXiv:2402.00531.
\bibitem{pere-2023} S. Perez, S. Maddu, I. F. Sbalzarini, P. Poncet, Adaptive weighting of Bayesian physics informed neural
networks for multitask and multiscale forward and inverse problems, J. Comput. Phys. 491 (2023) 112342.
 \bibitem{rc-2022}R. van der Meer, C.W. Oosterlee, A. Borovykh, Optimally weighted loss functions for solving PDEs with Neural Networks, J. Comput. Appl. Math. 405 (2022) 113887.

\bibitem{km-2015}D.P. Kingma, J. Ba, Adam: A method for stochastic optimization, (2014) arXiv: 1412.6980.
\bibitem{ln-1989}D.C. Liu, J. Nocedal, On the limited memory BFGS method for large scale optimization, Math. Program. 45 (1) (1989) 503-528.
\bibitem{dk-2015}A.D. Keedwell, J$\acute{\text{o}}$ zsef D$\acute{\text{e}}$nes, Latin Squares and their Applications,205, North-Holland: B.V. Elsevier.
\bibitem{yar-2022}D. Yarotsky, Universal approximations of invariant maps by neural networks, Constr. Approx. 55 (2022) 407-474.
\bibitem{th-1995}T. Chen, H. Chen. Universal approximation to nonlinear operators by neural networks with arbitrary activation functions and its application to dynamical systems. IEEE Transactions on Neural Networks 6(4) (1995) 911-917.
\bibitem{stu-2008}B. Sturmfels. Algorithms in invariant theory. Springer Science \& Business Media, 2008.
\bibitem{xw-2010}X. Glorot,Y. Bengio, Understanding the difficulty of training deep feedforward neural networks, J. Mach. Learn. Res. 9 (2010) 249-256.
\bibitem{ms-1987}M. Stein, Large sample properties of simulations using Latin hypercube sampling, Technometrics 29 (1987) 143-151.
\bibitem{auta}A.G. Baydin, B.A. Pearlmutter, A.A. Radul, J.M. Siskind, Automatic differentiation in machine learning: a survey, J. Mach. Learn. Res. 18(153)(2018) 1-43.
\bibitem{ft-1997} E.F. Toro, Riemann Solvers and Numerical Methods for Fluid Dynamics: A Practical Introduction, Berlin: Springer-Verlag, 2009.
\bibitem{ak-2020}T. Aktosun, F. Demontis, C. van der Mee, Exact solutions to the sine-Gordon equation, J. Math. Phys. 51(2010) 123521.
\bibitem{aps-2000}A.P.S. Selvadurai, Poisson's equation, In: Partial Differential Equations in Mechanics 2. Berlin: Springer-Verlag, 2000.
\bibitem{aa-1981} W.F. Ames, Nonlinear Partial Differential Equations in Engineering (vol II), New York: Academic, 1972.
\bibitem{peh-2000}P.E. Hydron, How to onstrut the disrete symmetries of partial differential equations, Eur. J. Appl. Math. 11 (2000) 515-527.
\bibitem{Dor-2011} V. Dorodnitsyn, Application of Lie Groups to Difference Equations,  CRC Press, Boca Raton, 2011.
\bibitem{KdV}D. J. Korteweg, G. de Vries, On the change of form of long waves advancing in a rectangular canal, and on a new type of long stationary waves, Phil. Mag. 39 (1895) 422-443.
\bibitem{Pin-1999} A. Pinkus, Approximation theory of the MLP model in neural networks, Acta Numer. 8 (1999) 143-195.
\bibitem{Con-1996} G.M. Constantine and T.H. Savits, A multivariate Fa\`{a} di Bruno formula with applications, Trans. Amer. Math. Soc. 348 (1996) 503-520.
\end{thebibliography}
\end{document}